\documentclass[final,5p,twocolumn]{elsarticle} %5p,twocolumn, review

\usepackage{lineno}%,hyperref
\usepackage[hidelinks,colorlinks=true]{hyperref}
\usepackage{xspace}
\usepackage{booktabs}
\usepackage{tabularx}
\usepackage{multirow}
\usepackage{color, colortbl}
\usepackage{mathptmx} 
\usepackage{graphicx}
\usepackage{array}
\usepackage{amsmath}
\usepackage{setspace}
\usepackage{color}
\usepackage[normalem]{ulem}
\usepackage[caption=false,font=normalsize,labelfont=sf,textfont=sf]{subfig}
\usepackage{geometry}
\usepackage{enumitem}
\usepackage{longtable}
\usepackage{float}

\usepackage{etoolbox}

\journal{Robotics and Autonomous Systems}

\makeatletter
\def\ps@pprintTitle{%
	\let\@oddhead\@empty
	\let\@evenhead\@empty
	\def\@oddfoot{}%
	\let\@evenfoot\@oddfoot}
\DeclareRobustCommand\onedot{\futurelet\@let@token\@onedot}
\newcommand{\@onedot}{\ifx\@let@token.\else.\null\fi\xspace}

\newcommand{\ie}{\emph{i.e}\onedot}

\newcommand{\etal}{\emph{et al}\onedot}
\makeatother

\let\OLDthebibliography\thebibliography
\renewcommand\thebibliography[1]{
	\OLDthebibliography{#1}
	\setlength{\parskip}{0pt}
	\setlength{\itemsep}{0pt plus 0.3ex}
}

\begin{document}
%*************************************  FRONT MATTER     *****************************************
\setlength{\abovedisplayskip}{6pt}
\setlength{\belowdisplayskip}{8pt}
\begin{frontmatter}
	
\title{Keyframe-based monocular SLAM: design, survey, and future directions}
			
\author{Georges Younes\textsuperscript{1,2}}
\author{Daniel Asmar\textsuperscript{1}}
\author{Elie Shammas\textsuperscript{1}}
\author{John Zelek\textsuperscript{2}}
\address{\textsuperscript{1}Vision and Robotics Lab, American University of Beirut, Beirut, Lebanon}
\address{\textsuperscript{2}Systems Design Department, University of Waterloo, Waterloo, Ont. Canada}
\address{\textcopyright 2017. This manuscript version is made available under the CC-BY-NC-ND 4.0 license \url{http://creativecommons.org/licenses/by-nc-nd/4.0/}}
\address{Accepted Manuscript in Robotics and Autonomous Systems, v. 98, p. 67-88, 2017, \url{https://doi.org/10.1016/j.robot.2017.09.010} \vspace{-1cm} }
%© <year>. This manuscript version is made available under the CC-BY-NC-ND 4.0 license http://creativecommons.org/licenses/by-nc-nd/4.0/

		%\author{Vision and Robotics Lab,American University of Beirut}
		%\ead[url]{http://feaweb.aub.edu.lb/research/cvrl/}
%------------------------------------------------------------------------------------------	
\begin{abstract}
Extensive research in the field of monocular SLAM for the past fifteen years has yielded workable
systems that found their way into various applications in robotics and augmented reality. Although
filter-based monocular SLAM systems were common at some time, the more efficient keyframe-based
solutions  are becoming the de facto methodology for building a monocular SLAM system. The
objective of this paper is threefold: first, the paper serves as a guideline for people seeking to
design their own monocular SLAM according to specific environmental constraints. Second, it presents a
survey that covers the various keyframe-based monocular SLAM systems in the literature, detailing
the components of their implementation, and critically assessing the specific strategies made in
each proposed solution.  Third, the paper provides insight into the direction of future research in this
field, to address the major limitations still facing monocular SLAM; namely, in the issues of
illumination changes, initialization, highly dynamic motion, poorly textured scenes, repetitive
textures, map maintenance, and failure recovery.
\end{abstract}

%------------------------------------------------------------------------------------------	
	\begin{keyword}
		\texttt{Visual SLAM}\sep Monocular\sep Keyframe based.	
	\end{keyword}
%------------------------------------------------------------------------------------------	
	\end{frontmatter}

	%\linenumbers
    \newcommand{\correct}[2]{{\color{red}(\sout{#1}{ #2)}}}
    \newcommand{\comment}[1]{{\color{red}({ #1)}}}
    %\let\clearpage\relax    %%BUGGGGG
%	\twocolumn

%*************************************  INTRODUCTION *********************************************
\section{Introduction}
When we enter a room or a space, we assess our surroundings, build a map, and keep track of our
location with respect to all the objects in the space. Simultaneous Localization and Mapping (SLAM)
is the equivalent of this procedure for a machine.  A robot's location in the map is needed for
various tasks such as navigation, surveillance and manipulation, to name a few. SLAM can be performed
with various sensors and their combinations, such as cameras, LIDARS, range finders, GPS, IMU, etc.
With more information such as depth, location, and velocity are available, the easier the problem. If
only a single camera is present, the problem is more challenging since they are bearing only
sensors; however, the rewards are great because a single camera is passive, consumes low power, is
of low weight, needs a small physical space, is inexpensive, and ubiquitously found in hand-held
devices. Furthermore, cameras can operate across different types of environments, both indoor and
outdoor, in contrast to active sensors such as infrared based RGB-D sensors that are sensitive to
sunlight. For the aforementioned reasons, this paper handles SLAM solutions using a single camera
only and is referred to as monocular SLAM. Even though there are still many challenges facing
monocular SLAM, the research community is actively seeking and finding solutions to these problems. 

%The Simultaneous Localization and Mapping (SLAM) problem emerged from the basic need of an agent to
%localize itself in an unknown environment, and concurrently build a map of that environment.
%Recently, Visual SLAM solutions using a single camera\textemdash  referred to as monocular SLAM\textemdash  have been
%gaining considerable popularity;  with their low cost and small size, cameras are frequently
%used in applications where weight, and power consumption are deciding factors, such as in Unmanned
%Aerial Vehicle (UAV) applications. Furthermore, cameras can operate across different types of environments, both
%indoor and outdoor, in contrast to infrared based RGB-D sensors that are sensitive to
%sunlight.  Cameras are ubiquitously found in hand-held devices such as in phones and tablets; and with the
%emergence of augmented reality applications, a camera appears to be the natural sensor of choice to
%localize users, while projecting virtual scenes to them from the correct viewpoint.  	

A number of surveys for the general SLAM problem exist in the literature, but only a few of them
handle monocular SLAM in an exclusive manner.  The most recent survey on SLAM is the one by
\cite{cadena_2016_TRO}, which discusses the complete SLAM problem, but does not delve into the
specifics of keyframe-based monocular SLAM, as we do in this paper. In 2011,
\cite{scaramuzza_2011_ram} published a tutorial on Visual odometry, but did not detail the solutions put
forward by the research community; rather, it presented a generic design for earlier visual odometry
systems. In 2012, \cite{fuentes_2012_air} published a general survey on Visual SLAM, but also did
not detail the solutions put forward by different people in the community.  Also, subsequent to the
date of the aforementioned papers were published, almost thirteen new systems have been proposed,
with many of them introducing significant contributions to keyframe-based Monocular SLAM.  In 2015,
\cite{yousif_2015_iis} also published a survey on Visual SLAM with the main focus on filter-based
methods, visual odometry (VO), and RGB-D systems.  While filter-based Visual SLAM solutions were
common before 2010, most solutions thereafter designed their systems around a \textit{non-filter},
keyframe-based architecture.  The survey of Yousif \etal describes a generic Visual SLAM but lacks
focus on the details and problems of monocular keyframe-based systems. 

To our knowledge, this paper is unique in that it systematically discusses the different components
of keyframe-based monocular SLAM systems, while highlighting the nuances that exist in their
different implementations.  Our paper is a valuable tool for any scholar or practitioner in the
field of localization and mapping. With the many new proposed systems appearing every day, the
information is daunting for the novice and people are often perplexed as to which algorithm they
should use.  Furthermore, this paper produces structure for researchers to quickly pinpoint the
shortcomings of each of the proposed techniques and accordingly help them focus their effort on
alleviating these weaknesses.

The remainder of the paper is structured as follows.  Section \ref{sec:hist} presents the
architecture of a generic keyframe-based Monocular SLAM, and details the particulars of its  major
building blocks, mainly: data association, visual initialization, pose estimation,
topological/metric map generation, Bundle Adjustment (BA)/Pose Graph Optimization(PGO)/map maintenance, and global localization (failure recovery
and loop closure). Sections \ref{sec:design} and \ref{sec:ClosedSource} respectively survey all the
open-source and closed-source keyframe-based Monocular SLAM systems in the literature. Open-source
systems are treated in more depth, given the additional insight gained through access to their code.
In Section \ref{sec:discussion}, we delve into the traits of the seven building blocks and explain
how to best implement them for different environmental conditions. Finally, we provide in the
conclusion our insight into the current open problems in monocular SLAM and we propose possible
venues for the solution of each of these problems.

%*************************************  OVERVIEW **************************************************
\section{Keyframe-based Monocular SLAM Architecture}\label{sec:hist}
\begin{figure*}[!ht] %htb
	\centering
	\includegraphics[height=2in]{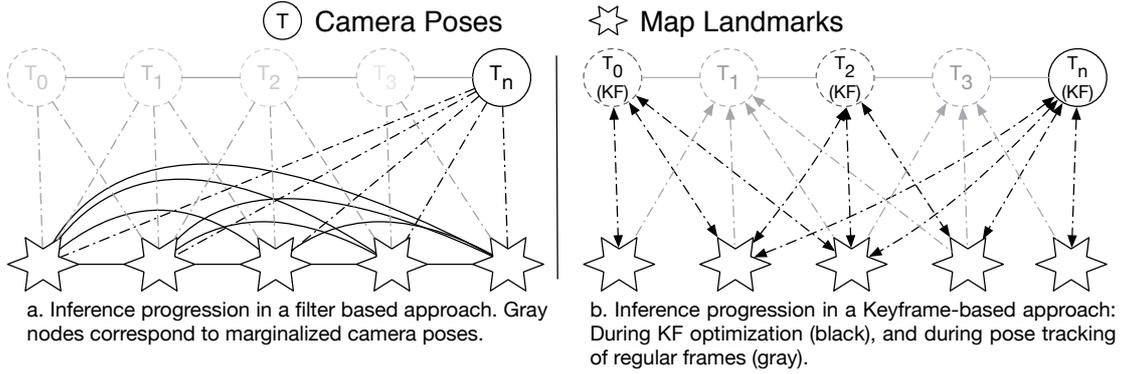}
	\caption{Inference in filter based vs. Keyframe-based VSLAM.}
	\label{fig:Inference}
\end{figure*}

Monocular SLAM solutions are either filter-based, such as using a Kalman filter; or keyframe-based,
relying on optimization to estimate both motion and structure. In filter-based systems, the localization and mapping are intertwined: the camera pose $T_n$, with the entire state of all landmarks in the
map, are tightly joined and need to be updated at every processed frame as shown in figure \ref{fig:Inference}-a. On the other hand, in keyframe-based systems, \textit{Localization} and \textit{Mapping} are separated into two steps: camera localization takes place on regular frames over a subset of the map (gray nodes in figure \ref{fig:Inference}-b), whereas keyframe-based optimization takes place on \textit{Keyframes}. 
 As a consequence of these differences, Strasdat \etal in 2010 showed that keyframe based methods
outperform filter-based ones; and it is therefore not surprising to note %(as the list in Table \ref{tab:allcontributions} in the appendix shows) 
that most new releases of monocular SLAM systems are keyframe-based. 

%------------------------------------------------------------------------------------------
%\begin{figure*}[!htb]
%\centering
%\subfloat[Filter based systems]{\includegraphics[width=2.5in,height=1.65in]{filter.pdf}%
%%\subfloat[Filter based systems]{\includegraphics[width=1.7in,height=1.25in]{filter.pdf}%
%\label{SubFig:Filter}}
%\hfil 
%\subfloat[Keyframe-based systems]{\includegraphics[width=2.5in,height=1.65in]{non_filter.pdf}%
%%\subfloat[Keyframe-based systems]{\includegraphics[width=1.7in,height=1.25in]{non_filter.pdf}%
%\label{SubFig:NonFilter}}
%\caption{Data inference in filter versus keyframe based monocular SLAM systems. %Figure inspired by\cite{strasdat_2010_ICRA}
%	}
%\label{Fig:Inference}
%\end{figure*}
%------------------------------------------------------------------------------------------  
For this reason and the fact that filter-based SLAM have been relatively well covered in the
literature in many surveys
(\cite{fuentes_2012_air}, \cite{saeedi_2016_jfr}, \cite{bailey_2006_mra}, \cite{yousif_2015_iis},etc.), the focus of this paper will be on the analysis and survey of only Keyframe-based monocular SLAM systems, hereafter referred to as KSLAM.
%------------------------------------------------------------------------------------------ 
\subsection{KSLAM architecture}\label{sec:Overview}
\begin{figure*}[!ht] %htb
	\centering
	\includegraphics[width=\textwidth]{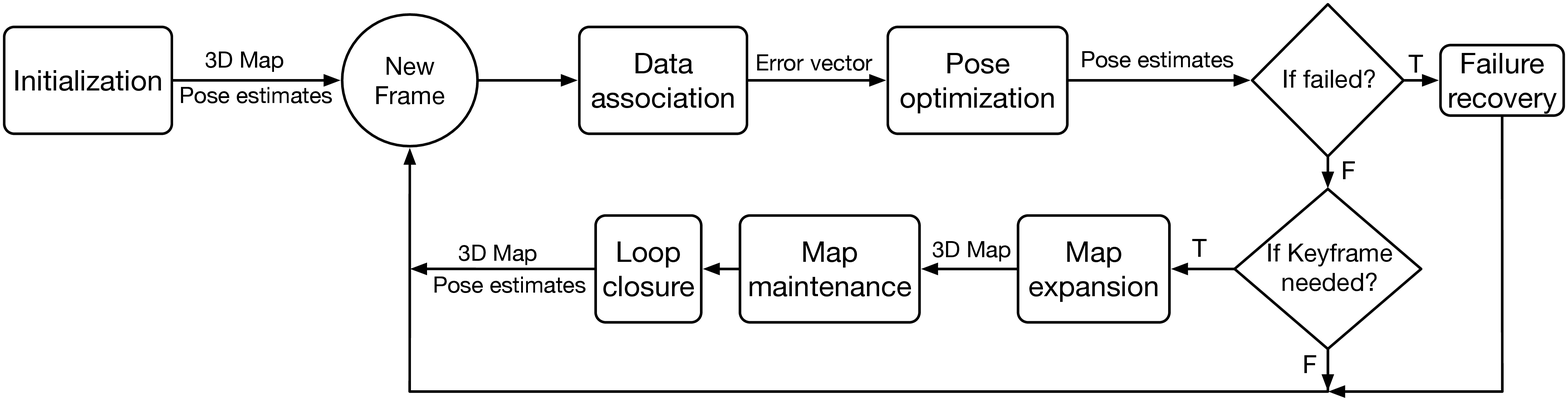}
	\caption{A generic KSLAM flowchart made of $7$ components: Visual initialization, data association, pose optimization, topological/metric map generation (map expansion), BA/PGO/map maintenance, failure recovery and loop closure. The basic types of information, returned/manipulated by each component, are overlaid  over their corresponding arrows, namely 3D map, pose estimates and error vector. }
	\label{fig:KSLAMGen}
\end{figure*}

Designing a KSLAM system requires the treatment of seven main components (Fig. \ref{fig:KSLAMGen});
namely (1) visual initialization , (2) data association, (3) pose estimation,
(4) topological/metric map generation, (5) BA/PGO/map maintenance, (6) failure recovery and (7) loop closure.
 Fig. \ref{fig:KSLAMGen} describes the general flow between these modules: at startup, KSLAM has no prior information about the camera pose nor the scene structure, the visual initialization module is responsible for establishing an initial 3D map and the first camera poses in the system. The visual initialization is the entry point of a KSLAM and runs only once at startup.
 When a new frame is available, data association uses the previous camera poses to guess a pose for the new frame; the predicted pose is used to establish associations with the 3D map. An error vector is then found as the difference between the \textit{true} measurements and their associated \textit{matches} generated using the guessed pose. 
 The error vector is iteratively minimized in the pose optimization module, using the guessed pose as a starting point. If the minimization diverges or the data association fails, failure recovery procedures are invoked. For regular frames, the pipeline ends here however, if the frame was chosen as a keyframe, it is used to triangulate new landmarks, thus expanding the 3D map.
 To ensure the coherency of the map, reduce errors and remove outliers, map maintenance continuously optimizes the map while another parallel process attempts to detect loop closures and accordingly minimize the errors accumulated over the traversed loop.
 
\begin{figure*}[p] %htb
	\centering
		\includegraphics[width=0.85\textwidth]{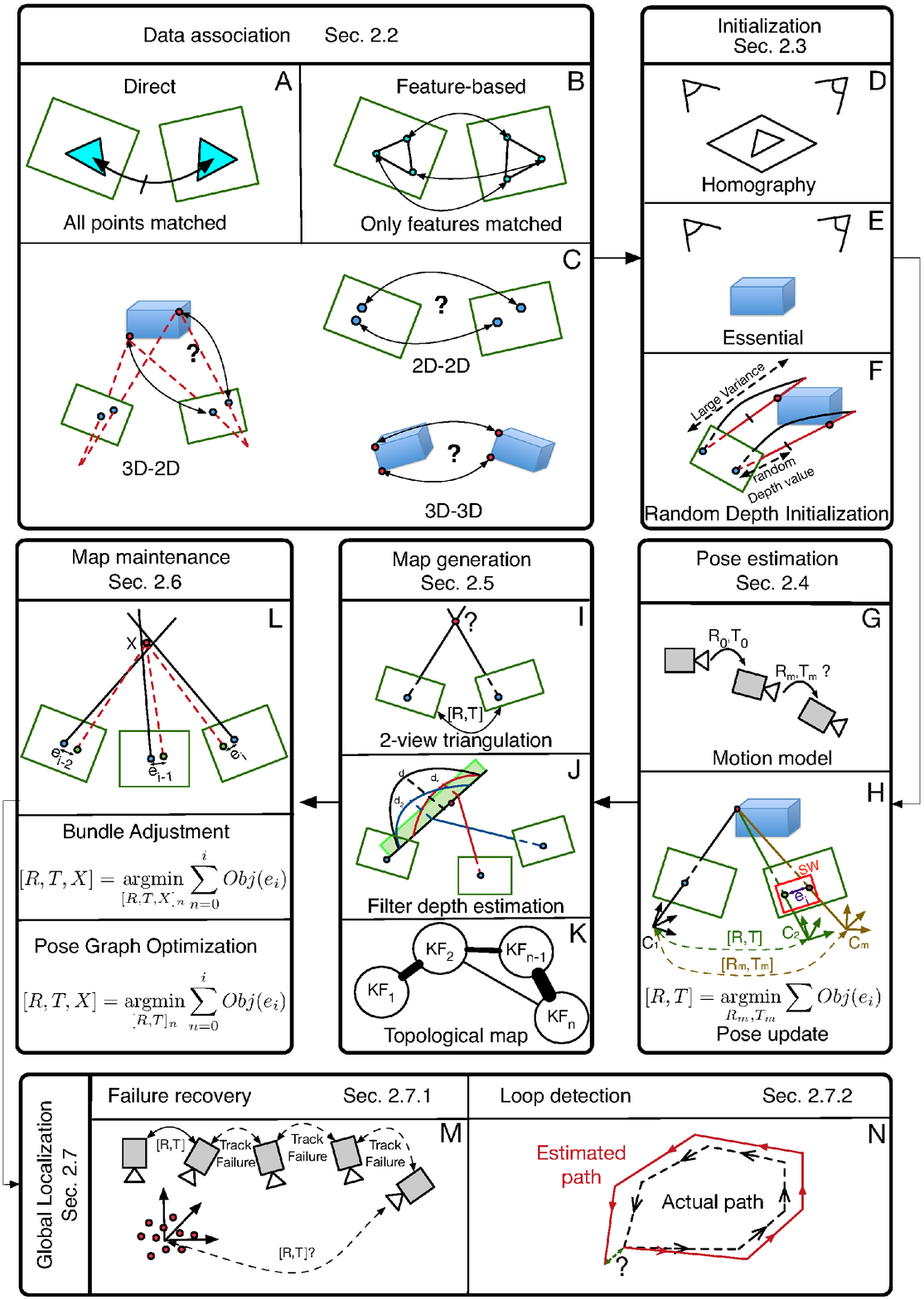}
    \caption{A graphic representation for the different building blocks of a generic KSLAM.}
	\label{fig:building}
\end{figure*}

Fig. \ref{fig:building} presents the different solutions for each of the KSLAM modules; we further elaborate on these solutions in the upcoming sections.
 
%------------------------------------------------------------------------------------------
%\newgeometry{left=1.5cm,right=1.5cm,bottom=2cm,top=1.5cm}
%\restoregeometry
%-----------------------------------------------------------------------------------------

\subsection{Data association}\label{sec:data}
VSLAM methods are categorized based on the type of information they process from the images,
as being direct, feature-based, or a hybrid of both.  

\subsubsection{Design choice}
\noindent {\textbf {a. Direct methods}}\\
Direct methods are categorized into either dense or semi-dense methods: dense methods exploit the information available at every pixel, whereas semi-dense methods exploit the information from every pixels at which the gradient of image brightness is significant. Fig. \ref{fig:building}-A shows an example of a direct approach: pixel values surrounding a location of interest (in this case a triangle) are aligned by a \textit{transformation} that minimizes the intensity values between the two locations of interest in both images. In practice a region of interest is defined as a square surrounding a pixel. 
%------------------------------------------------------------------------------------------
%\begin{figure*}[!hbt]
%	\centering
%    \subfloat[direct methods using all information of
%		the triangle to match to a query image.]{\includegraphics[width=2.5in]{DirectType.pdf}%
%        \label{SubFig:Direct}}
%	\hfil
%	\subfloat[indirect methods use the features of the
%		triangle to match to the features of a query image.]{\includegraphics[width=2.5in]{indirecttype.pdf}%
%	\label{SubFig:Indirect}}
%	\caption{Data types used by a Visual SLAM system.}
%	\label{fig:datatypes}
%\end{figure*}
%------------------------------------------------------------------------------------------
	%\begin{enumerate}[label=(\alph*)]
    
The basic underlying principle for all direct methods is known as the brightness consistency
constraint and is best described as:
%------------------------------------------------------------------------------------------
\begin{equation}
J(x,y,t)=I(x+u(x,y),y+v(x,y),t+1),
\label{BrightnessConstraint}
\end{equation}
%------------------------------------------------------------------------------------------
where $x$ and $y$ are pixel coordinates; $u$ and $v$ denote displacement functions of the pixel
$(x,y)$ between two images $I$ and $J$ of the same scene taken at time $t$ and $t+1$ respectively. The brightness consistency constraint is based on the assumption that a point from the world's surface, observed in image $I$, will have the same intensity when observed in a subsequent image $J$.
%Every pixel in the image provides one brightness constraint; however, it also adds two unknowns, $u$ and $v$, and hence the system becomes under-determined with $n$ equations and $2n$ unknowns, where $n$ is the number of pixels in the image. 
To render equation \eqref{BrightnessConstraint} solvable, \cite{lucas_1981_IJCAI} suggested,
in what they referred to as Forward Additive Image Alignment (FAIA), to replace all the individual
pixel displacements $u$ and $v$ by a single general motion model, in which the number of parameters
is dependent on the implied type of motion. FAIA iteratively minimizes the squared pixel-intensity
difference between the two images over the transformation parameters $p$:
%------------------------------------------------------------------------------------------
\begin{equation} 
	\underset{p}{\mathrm{argmin}}\sum_{x,y}[I(W(x,y,p))-J(x,y)]^2 , \label{FAIA} 
\end{equation}
%------------------------------------------------------------------------------------------
where $W(.,.,p)$ is a warping transform that encodes the relationship relating the two images and
$p$ corresponds to the parameters of the transform. Equation \eqref{FAIA} is non-linear and requires
an iterative non-linear optimization process, with a computational complexity of $O(n^2N+n^3)$ per
iteration, where $n$ is the number of parameters in $p$ and $N$ is the number of pixels in the image.
Since 1981, other variants of the FAIA were suggested such as FCIA (Forward Compositional Image
Alignment), ICIA (Inverse Compositional Image Alignment) and IAIA (Inverse Additive Image
Alignment) each with different computational complexities. A detailed comparison between these
variations can be found in \cite{baker_2004_IJCV}.\\
%as shown in Table \ref{tab:IaComp}
%\begin{table}[!htb]
%    	\renewcommand{\arraystretch}{1.3}
%    	\caption{Computational complexity of different gradient descent image alignment methods. Abbreviations used: number of warp parameters
%    		($n$), number of pixels in the image ($N$), and a positive integer ($k$)
%    		\label{tab:IaComp}}
%    	\centering
%    	\begin{tabular}{c c c }
%    		\hline
%    		Method&Pre-computation&Per-iteration\\
%    		\hline
%    		FAIA & - & $O(n^2N+n^3)$ \\
%    		\hline
%	    	FCIA & $O(nN)$ & $O(n^2N+n^3)$ \\
%    		\hline
%    		IAIA & $O(k^2N)$ & $O(nN + kN + k^3)$ \\
%    		\hline
%    		ICIA & $O(n^2N$) &  $O(nN + n^3)$ \\
%    		\hline
%    	\end{tabular}
%\end{table}
    
%%DIFFERENCE BETWEEN VO and VSLAM%%
%Further discussion regarding these methods is outside the scope of this work; however, it is
%noteworthy to mention that they are all capable of recovering the inter-frame camera motion and
%are referred to as solutions to the visual odometry task. While visual odometry estimates the
%camera motion between consecutive frames only, Visual SLAM further builds a map of the
%environment and uses it to optimize the camera's pose estimates. As a result, visual odometry
%methods tend to
%suffer from relatively large drifts.
%------------------------------------------------------------------------------------------

\noindent {\textbf {b. Feature-based methods}}\\ 
Feature-based methods were introduced to reduce the computational complexity of processing
each pixel; this is done by matching only salient image locations, referred to as features, or
keypoints. An example of feature-based matching is shown in Fig. \ref{fig:building}-B.
 A descriptor is associated to each feature, which is used to provide a
quantitative measure of similarity to other keypoints.
On one hand, features are expected to be distinctive, invariant to viewpoint and
illumination changes, as well as resilient to blur and noise; on the other hand, it is desirable for
feature extractors to be computationally efficient and fast.  Unfortunately, such objectives are
hard to achieve at the same time causing a trade-off between computational speed and feature quality.    
	
The computer vision community has developed, over decades of research, many different feature
extractors and descriptors, each exhibiting varying performances in terms of rotation and scale
invariance, as well as speed \citep{krig_2014_Apress}. The selection of an appropriate feature detector
depends on the platform's computational power and the type of environment. Feature detector examples include the
Hessian corner detector \citep{beaudet_1978_ICPR}, Harris detector \citep{harris_1988_FAVC},
Shi-Tomasi corners \citep{shi_1994_CVPR}, Laplacian of Gaussian detector
\citep{lindeberg_1998_IJCV}, MSER \citep{matas_2002_bmvc}, Difference of Gaussian
\citep{lowe_2004_IJCV} and the accelerated segment test family of detectors (FAST, AGAST, OAST)
\citep{mair_2010_ECCV}. 
	
%%%To minimize computational requirements, most indirect systems use FAST \citep{rosten_2006_ECCV} for their feature extractor, coupled with a feature descriptor for data association.  
Feature descriptors include, and are not limited to, BRIEF \citep{calonder_2012_PAMI}, BRISK
\citep{leutenegger_2011_ICCV}, SURF \citep{bay_2008_CVIU}, SIFT \citep{lowe_1999_ICCV}, HoG
\citep{dalal_2005_CVPR}, FREAK \citep{alahi_2012_CVPR}, ORB \citep{rublee_2011_ICCV}, and a low level local patch of pixels.  Further information regarding feature extractors and descriptors is outside the scope of this work, but the reader can refer to \cite{moreels_2007_IJCV},
\cite{hartmann_2013_ECMR}, \cite{rey_2014_CORR}, or \cite{hietanen_2016_NC} for comparisons.\\
%------------------------------------------------------------------------------------------
	
\noindent {\textbf {c. Hybrid methods}}\\ 
Different from the direct and feature-based methods, some systems such as SVO are considered hybrids,
which use a combination of both to refine the camera pose estimates, or to generate a dense/semi-dense map.
%------------------------------------------------------------------------------------------
Once a design is chosen, data association is defined as the process of establishing measurement correspondences across different images using either 2D-2D, 3D-2D, or 3D-3D
correspondences. The different types of data association are depicted in Fig. \ref{fig:building}-C.
%------------------------------------------------------------------------------------------

\subsubsection{Data association types}
\noindent {\textbf {2D-2D}}\\
In 2D-2D correspondence, the 2D feature's location in an image $I_2$ is sought, given its 2D position
in a previously acquired image $I_1$. Depending on the type of information available, 2D-2D
correspondences can be established in one of two ways: when a map is not available and neither the camera
transformation between the two frames nor the scene structure is available (\ie during system
initialization), 2D-2D data association is established through a search window surrounding the
feature's location from $I_1$ in $I_2$.   
When the transformation relating $I_1$ and $I_2$ is known (\ie the camera pose is
successfully estimated), 2D-2D data correspondences are established through epipolar geometry, where a
feature in $I_1$ is mapped to a line in $I_2$, and the two dimensional search window collapses to a
one dimensional search along a line. This latter case often occurs when the system attempts to triangulate
2D features into 3D landmarks during map generation. To limit the computational expenses, a bound is
imposed on the search region along the epipolar line.
    
In both methods, each feature has associated with it a descriptor, which can be used to  provide a
quantitative measure of similarity to other features.  The descriptor similarity measure varies with
the type of descriptors used; for example, for a local patch of pixels, it is typical to
use the sum of squared difference (SSD), or a Zero-Mean SSD score (ZMSSD) to increase robustness
against illumination changes, as is done in \citep{martin_1995_ICIAS}.  For higher order feature
descriptors \textemdash such as ORB, SIFT, or SURF\textemdash the L1-norm, the L2-norm, or Hamming distances may be used;
however, establishing matches using these measures is computationally intensive and may, if not
carefully applied, degrade real-time performance.  For such purpose, special implementations that
sort and perform feature matching in KD trees, or bags of words, are usually employed. Examples
include the works of \cite{muja_2009_CVTA}, and \cite{lopez_2012_TRO}.\\
%------------------------------------------------------------------------------------------
    
\noindent {\textbf {3D-2D}}\\ 
In 3D-2D data association, the camera pose and the 3D structure are known, and one seeks to estimate correspondences between the 3D landmarks and their 2D projection onto a newly acquired frame,
without the knowledge of the new camera pose ($T,R$). This type of data association is typically
used during the pose estimation phase of KSLAM. To solve this problem, previous camera poses are
exploited in order to yield a hypothesis on the new camera pose and accordingly, project the 3D
landmarks onto that frame. 3D-2D data association then proceeds similarly to 2D-2D feature matching,
by defining a search window surrounding the projected location of the 3D landmarks and searching
for matching feature descriptors.\\
%------------------------------------------------------------------------------------------
%\begin{figure}[!hbt]
%		\centering
%		\includegraphics[width=4.0in]{dataAssociation.pdf}
%		\caption{3D-2D Data association problem (n.a. stands for not available during the task).}
%		\label{Figure:DataAssociationl}
%\end{figure}
%------------------------------------------------------------------------------------------ 

\noindent {\textbf {3D-3D}}\\ 
3D-3D data association is typically employed to estimate and correct accumulated drift along loops: when a
loop closure is detected, descriptors of 3D landmarks, visible in both ends of the loop, are used to
establish matches among landmarks that are then exploited\textemdash as explained in
\citep{umeyama_1991_PAMI}\textemdash to yield a similarity transform between the frames at both ends of the loop. 
%--------------------------------------------------------------------------------------------------
\subsection{Visual Initialization}\label{sec:init}
Monocular cameras are bearing-only sensors, which cannot directly perceive depth;
nevertheless, a scaled depth can be estimated via temporal stereoscopy, after observing the same
scene through at least two different viewpoints.  After KSLAM is initialized, camera pose and 3D
structure build on each other, in a heuristic manner, to propagate the system in time, by expanding the map
to previously unobserved scenes, while keeping track of the camera pose in the map.
%to propagate their belief forward in time. 

The problem is more difficult during initialization, since neither pose, nor structure is known. %%%and KSLAM requires a special initialization phase, during which both a map of 3D landmarks, and the initial camera poses are generated.
%%%%%no below figure
%---------------------------------------------------------------------------------------------------
%\begin{figure}[!hbt]
%		\centering
%		\includegraphics[width=4in]{initialization.pdf}
%		\caption{Initialization required by any Visual SLAM system.}
%		\label{fig:initialization}
%\end{figure}
%----------------------------------------------------------------------------------------------------
In early monocular SLAM systems, such as in MonoSLAM \citep{davison_2007_PAMI}, initialization
required the camera to be placed at a known distance from a planar scene, composed of four corners
of a two dimensional square; the user initialized SLAM by keying in the distance separating
the camera from the square.  Thereafter, to lessen these constraints, researchers adopted the
methods developed by \cite{higgins_1981_LN} to simultaneously recover the camera pose and the 3D scene structure. Higgins's intuition was to algebraically eliminate the depth from the problem,
yielding both the Essential and the Homography matrices.  However, the elimination of depth has
significant ramifications on the recovered data: since the exact camera motion between the two views
cannot be recovered, the camera translation vector is recovered up to an unknown scale $\lambda$.
Since the translation vector between the two views defines the baseline used to triangulate 3D
landmarks, scale loss also propagates to the recovered 3D landmarks, yielding a scene that is also
scaled by $\lambda$. Fig. \ref{fig:building}-D and E describes the two-view initialization using a Homography (which assumes the observed scene to be planar) and an Essential matrix (which assumes the observed scene to be non-planar). % Furthermore, the decomposition of the Essential or Homography matrices yields
\begin{figure}[!htb]%!htb
	\centering
	\includegraphics[width=3.5in,trim={0cm 0cm 0cm 0cm},clip]{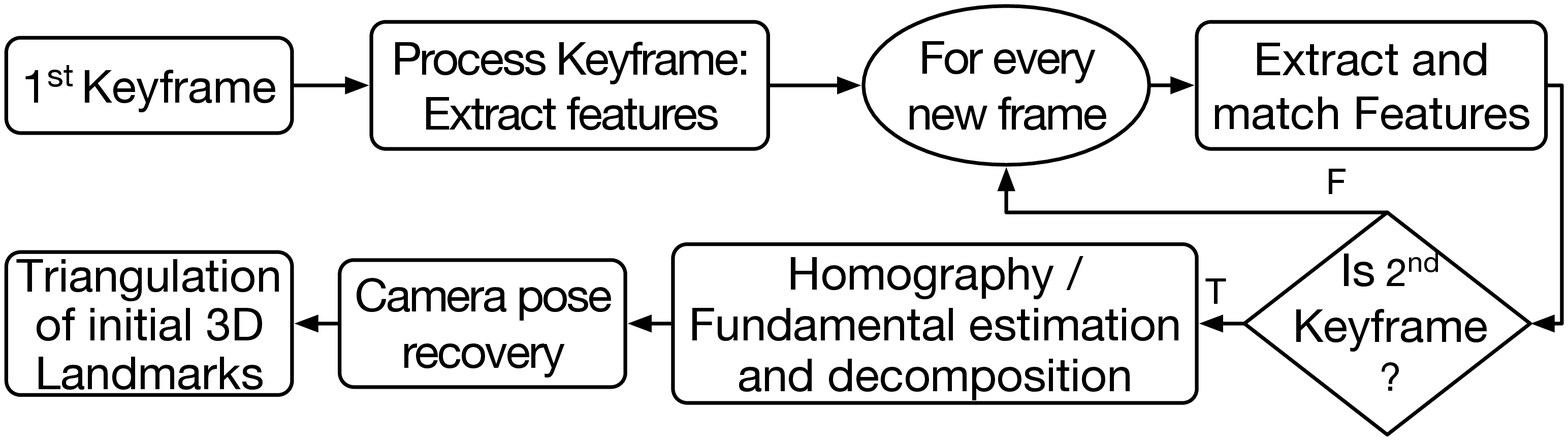}%3.5in for the 5p 
	\caption{Generic model-based initialization flowchart.}
		\label{fig:ModelBasedInit}
\end{figure}
%------------------------------------------------------------------------------------------

Figure \ref{fig:ModelBasedInit} shows the flowchart of a generic model-based initialization; the first frame processed by the KSLAM system is typically set as the first keyframe. Subsequent frames are processed by establishing 2D-2D data associations, which are monitored to decide whether the new frame is the second keyframe or not. The decision criteria is based on the 2D distances between the found matches in both images. The matches are then used to estimate a Homography (degenerate for non-planar scenes) or a Fundamental matrix (degenerate for planar scenes) using a robust model fitting method (RANSAC or MLESAC \citep{torr_2000_CVIU}). The estimated Homography or the Fundamental matrix are then decomposed as described in \cite{hartley_2003_Cambridge} into an initial scene structure and initial camera poses.
To mitigate degenerate cases, \textit{random depth} initialization (shown in Fig.\ref{fig:building}-F), as its name suggests, initializes a KSLAM by randomly assigning depth values with large variance to a single
initializing keyframe. The random depth is then iteratively updated over subsequent frames until the
depth variance converges. 
%------------------------------------------------------------------------------------------

\subsection{Pose estimation}\label{sec:pose} 
\subsubsection{Motion models}
\begin{figure*}[!ht]%!hbt
	\centering
	\includegraphics[width=\textwidth]{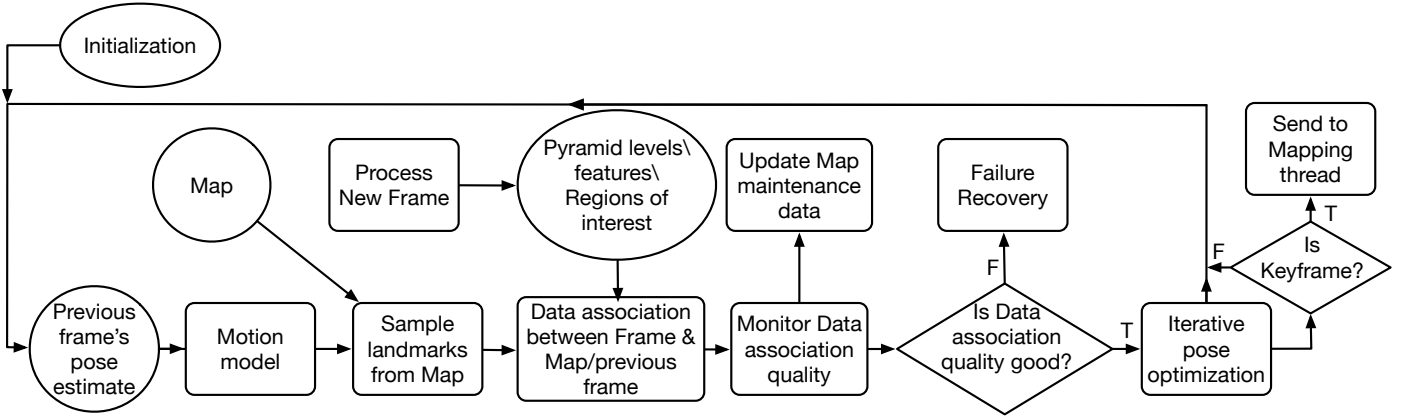}%6in
	\caption{Generic pose estimation flowchart.}
	\label{fig:posePipeline}
\end{figure*}

Figure \ref{fig:posePipeline} presents a generic flowchart for pose estimation.  Because data association is computationally expensive,
most monocular SLAM systems assume, for the pose of each new frame, a prior, which guides and limits
the amount of work required for data association. Estimating this prior (depicted in Fig. \ref{fig:building}-G) is generally the first task
in pose estimation: data association between the two frames is not known yet and one seeks to
estimate a prior on the pose of the second frame ($T,R$), given previously estimated poses.
	
Most systems employ a \emph{constant velocity} motion model that assumes a smooth camera motion and
use the pose changes across the two previously tracked frames to estimate the prior for the current
frame. Some systems assume no significant change in the camera pose between consecutive frames, and
hence they assign the prior for the pose of the current frame to be the same as the previously
tracked one.
%------------------------------------------------------------------------------------------		
%\begin{figure}[!h]
%		\centering
%		\includegraphics[width=3.5in]{pose.pdf}
%		\caption{Pose estimation required by any Visual SLAM system}
%		\label{fig:pose}
%\end{figure}
%------------------------------------------------------------------------------------------	

The pose of the prior frame is used to guide the data association procedure in several ways. It
helps determine a potentially visible set of features from the map in the current frame, thereby
reducing the computational expense of blindly projecting the entire map. Furthermore, it helps
establish an estimated feature location in the current frame, such that feature matching takes place
in small search regions, instead of across the entire image. Finally, it serves as a starting point
for the minimization procedure, which refines the camera pose.
%------------------------------------------------------------------------------------------
%------------------------------------------------------------------------------------------

\subsubsection{Pose optimization}
Direct and feature-based methods estimate the camera pose by minimizing a measure of error between
frames; direct methods measure the photometric error, modeled as the intensity difference between pixels; in contrast, feature-based methods measure
the re-projection error of landmarks from the map over the frame's prior pose.  The re-projection
error is formulated as the distance in pixels between a projected 3D landmark onto a frame, and its
corresponding 2-D position in the image.
A motion model is used to seed the new frame's pose at $C_m$ (Fig. \ref{fig:building}-H), and a list
of potentially visible 3D landmarks from the map are projected onto the new frame. Data association
takes place in a search window $S_w$ surrounding the location of the projected landmarks. KSLAM
then proceeds by minimizing an error vector (the geometric distance $d$ in the case of feature-based methods or the intensity residuals in the case of direct methods) over the parameters of the rigid body
transformation.  To gain robustness against outliers, the minimization takes place over an objective
function (Huber norm) that penalizes features with large errors. The camera pose optimization
problem is then defined as:
%------------------------------------------------------------------------------------------
\begin{equation}
	T_i=\underset{T_i}{\mathrm{argmin}}\sum_jObj(e_j),
	\label{eq:pose}
\end{equation}
    where $T_i$  is a minimally represented Lie group of either $S\xi(3)$ or $ sim(3)$ camera pose,
    $Obj(.)$ is an objective function and $e_j$ is the error defined through data association for
    every matched feature $j$ in the image. Finally, the system decides whether the new frame should
    be flagged as a keyframe or not. The decisive criteria can be categorized as either
    \textit{significant pose change} or \textit{significant scene appearance change}; a decision is
    usually made through a weighted combination of different criteria; examples of such criteria
    include: a significant change in the camera pose measurements (rotation and/or translation), the
    presence of a significant number of 2D features that are not observed in the map, a significant
    change in what the frame is observing (by monitoring the intensity histograms or optical flow),
    the elapsed time since the system flagged its latest keyframe, etc.
%------------------------------------------------------------------------------------------
%\begin{figure}[!hbt]
%		\centering
%		\includegraphics[width=4in]{Poseestimate.pdf}
%        \caption{Generic pose estimation procedure. $C_m$ is the new frame's pose estimated by the
%        motion model and $C_2$ is the actual camera pose.}
%		\label{fig:PoseEstimate}
%\end{figure}
%------------------------------------------------------------------------------------------

\subsection{Topological/metric map generation}\label{sec:generation} 
\begin{figure}[!htb]%!htb
		\centering
		\includegraphics[width=3.5in]{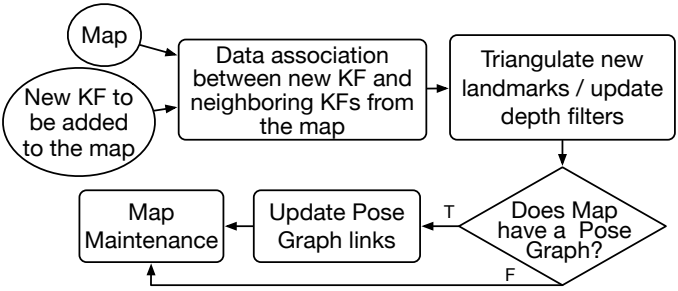}%3.5in
		\caption{Generic map generation flowchart.}
		\label{fig:mappingPipeline}
	\end{figure}

The map generation module is
responsible for generating a representation of the previously unexplored, newly observed
environment. Typically, the map generation module represents the world as a dense (for direct) or
sparse (for feature-based) cloud of points.  Figure \ref{fig:mappingPipeline} presents the flowchart of a
map generation module: different viewpoints of an unexplored scene are registered with their
corresponding camera poses through the pose tracking module. The map generation module then re-establishes data association between the new keyframe and a set of keyframes surrounding it, looking for matches. It then triangulates
2D points of interest into 3D landmarks as depicted in Fig.\ref{fig:building}-I and J; it also keeps track of their 3D coordinates, and expands
the map within what is referred to as a \emph{metric} representation of the scene. %However, as the camera explores large environments, metric maps suffer from the unbounded growth of their size, thereby leading to system failure.
	
Topological maps were introduced to alleviate the computational cost associated with processing a global metric representation, by forfeiting geometric information
in favor for connectivity information. In its most simplified form, a topological map consists of
nodes corresponding to locations, and edges corresponding to connections between the locations. In
the context of monocular SLAM, a topological map is an undirected graph of nodes that typically
represents keyframes linked together by edges, when shared data associations between the keyframes
exists, as depicted in Fig. \ref{fig:building}-K. For a survey on topological maps, the reader is referred to \cite{boal_2014_ROBOTICA}.
	%%%%%%%%%%%%%%%
	%------------------------------------------------------------------------------------------
	%------------------------------------------------------------------------------------------
In spite of the appeal of topological maps in scaling well with large scenes, metric information is
still required in order to maintain camera pose estimates. The conversion from a topological to a
metric map is not always trivial, and for this reason, recent monocular SLAM systems
\citep{mur-artal_2015_TRO,engel_2014_ECCV,lim_2011_CVPR,lim_2014_ICRA} employ hybrid maps, which are
locally metric and globally topological. The implementation of a hybrid map representation permits
the system to first reason about the world at a high level, which allows for efficient solutions to
loop closures and failure recovery using topological information;  and second, to increase
efficiency of the metric pose estimate, by limiting the scope of the map to a  local region
surrounding the camera \citep{fernandez_2015_Springer}. A hybrid map allows for local optimization
of the metric map, while maintaining scalability of the optimization over the global topological map
\citep{konolige_2010_BMVC}. 
	
	%In a topological map (See Fig. \ref{fig:PoseGraph}) every keyframe in the system is considered a
	%node and the thickness of the edges connecting the nodes is proportional to the number of common
	%landmarks observed in those nodes. 
	%-----------------------------------
	%\begin{figure}[!hbt]
	%	\centering
	%	\includegraphics[width=2.2in]{PoseGraph}
    %	\caption{Pose Graph diagram where the thickness of the edges connecting nodes is
    %	proportional to the number of landmarks in common between the nodes. }
	%	\label{fig:PoseGraph}
	%\end{figure}
%------------------------------------------------------------------------------------------
\subsubsection{Metric maps}
In a metric map, the structure of new 3D landmarks is recovered, given the pose transformation relating two keyframes observing the landmark, from epipolar geometry using the corresponding data associations between the keyframes \citep{hartley_1997_CVIU}.  %%%Since data association is prone to erroneous measurements, the map generation module is also responsible for the detection and handling of outliers, which can potentially destroy the map.

	%-dont comment out for longer version!!!!! -----------------------------------------------------------------------------------------
	%\begin{figure}[!hbt]
	%	\centering
	%	\includegraphics[width=3.5in]{generation}
	%	\caption{Map generation required by any Visual SLAM system.}
	%	\label{fig:generation}
	%\end{figure}
%------------------------------------------------------------------------------------------
Due to noise in data association and pose estimates of the tracked images, projecting rays from two
associated features will most probably not intersect in 3D space. To gain resilience against
outliers and to obtain better accuracy, the triangulation is typically performed over features
associated across more than two views.  Triangulation by optimization aims to estimate a landmark position $[x,y,z]$ from its
associated 2D features across $n$ views, by minimizing the sum of its re-projection errors in all keyframes observing it as described by:
%------------------------------------------------------------------------------------------
	 \begin{equation}
	 X=\underset{[x,y,z]}{\mathrm{argmin}}\sum_{n}e_n ,
	 \label{eq:triangulation}
	 \end{equation}
%------------------------------------------------------------------------------------------
%and shown in Fig. \ref{fig:2viewt},  
.  
%%%Uncomment this for longer version%%%%
%------------------------------------------------------------------------------------------
%\begin{figure}[!htb]
%		\centering
%		\includegraphics[width=0.45\textwidth]{2viewT.pdf}
%		\caption{Landmark triangulation by optimization.}
%		\label{fig:2viewt}
%\end{figure}
%------------------------------------------------------------------------------------------	
Filter based landmark triangulation %as shown in Fig. \ref{fig:FilterBasedTr}, 
recovers the 3D position of a landmark by first projecting into the 3D space a ray joining  the camera center of
the first keyframe  observing the 2D feature and its associated 2D coordinates. The projected ray is
then  populated with a filter having a uniform distribution ($D_1$) of landmark position
estimates, which are then updated as the landmark is observed across multiple views. The Bayesian
inference framework continues until the filter converges from a uniform distribution to a Gaussian
featuring a small variance ($D_3$) \citep{hochdorfer_2009_ICAR}. Filter-based triangulation results
in a delay before an observed landmark's depth has fully converged, and can be used for pose
tracking. %%%in contrast to triangulation by optimization methods, that can be used as soon as the landmark is triangulated from two views. 
To overcome this delay %%%and exploit all the information available from a feature that is not yet fully triangulated,  
\citep{civera_2008_TRO} suggested an inverse depth parametrization for newly observed features, with an associated variance that allows
for 2D features to contribute to the camera pose estimate, as soon as they are observed.

\subsubsection{Topological maps}
When a new keyframe is added into systems that employ hybrid maps, their topological map is updated
by incorporating the new keyframe as a node, and searching for data associations between the newly added node and surrounding ones; edges are then established to other nodes (keyframes) according to the number of found data associations: the thickness of the edges connecting the nodes is proportional to the number of common landmarks observed in those nodes.
%%%Uncomment this for longer version
%------------------------------------------------------------------------------------------
%\begin{figure}[!htb]
%		\centering
%		\includegraphics[width=0.45\textwidth]{FilterBasedTriang.pdf}
%		\caption{Landmark estimation using filter based methods.}
%		\label{fig:FilterBasedTr}
%\end{figure}
%------------------------------------------------------------------------------------------
\subsection{BA/PGO/map maintenance}\label{sec:maintenance}
\begin{figure}[htbp]%htb
		\centering
		\includegraphics[width=3.5in]{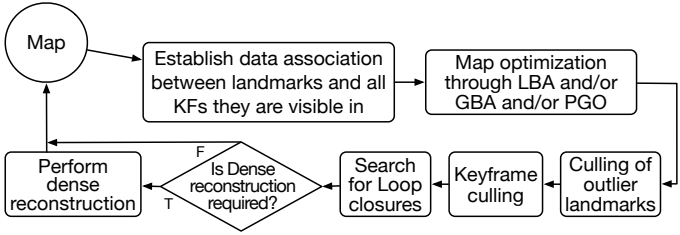}%3.5in
		\caption{Generic map maintenance flowchart.}
		\label{fig:maintpipeline}
	\end{figure}

Map maintenance takes care of optimizing the map through either bundle adjustment or pose graph
optimization \citep{kummerle_2011_ICRA}. Figure \ref{fig:maintpipeline} presents the steps required
for map maintenance of a generic monocular SLAM: during map exploration, new 3D landmarks are
triangulated based on the camera pose estimates; after some time, system drift manifests itself in
wrong camera pose measurements, due to accumulated errors in previous camera poses that were used to
expand the map. Map maintenance proceeds by establishing data association between the entire set of keyframes in the map or a subset of keyframes and performs a global bundle adjustment (GBA) or a local bundle adjustment (LBA) respectively. Outlier landmarks flagged from the optimization are then culled (removed from the map). To reduce the complexity of the optimization, redundant keyframes are also culled. 
Map maintenance is also responsible for detecting and optimizing loop closures as well as performing a dense map reconstruction for systems that allow for it.
%------------------------------------------------------------------------------------------
%------------------------------------------------------------------------------------------

Bundle adjustment is the problem of refining a visual reconstruction to produce a jointly
optimal 3D structure and viewing parameter estimates (camera pose and/or calibration).
What we mean by this is that the parameter estimates are found by minimizing some cost function
that quantifies the model fitting error, and that the solution is simultaneously optimal
with respect to both structure and camera variations
%%%%Bundle adjustment (BA) is the problem of refining a visual reconstruction to jointly produce an optimal structure and coherent camera pose estimates 
\citep{triggs_2000_va}. BA, depicted in Fig. \ref{fig:building}-L. is an optimization that minimizes the cost function defined by:
%\setlength{\belowdisplayskip}{4pt} \setlength{\belowdisplayshortskip}{4pt}
%\setlength{\abovedisplayskip}{4pt} \setlength{\abovedisplayshortskip}{4pt}
%------------------------------------------------------------------------------------------
\begin{equation}
        \underset{T,X}{\mathrm{argmin}}\sum_{i=1}^{N}\sum_{j\in S_i}Obj(e(T_i,X_j)),
        \label{eq:BA}
        \end{equation}
%------------------------------------------------------------------------------------------
\noindent where $T_i$ is a keyframe pose estimate and $N$ is the number of keyframes in the map or a subset of the map.
$X_j$ corresponds to the 3D pose of a landmark and $S_i$ represents the set of 3D landmarks observed
in Keyframe $i$. Finally, $e(T_i,X_j)$ is the re-projection error of a landmark $X_j$ on a keyframe
$T_i$, in which it is observed. 
	
Bundle adjustment is computationally involved and intractable if performed on all frames and all
poses. The breakthrough that enabled its application in PTAM is the notion of keyframes, where, %%%%in contrast to SfM methods that use all available frames, 
only select frames labeled as keyframes, are used in the map creation process.  Different algorithms apply different criteria for keyframe
labeling, as well as different strategies for BA. Some perform a Global Bundle Adjustment (GBA), over the entire map. Others  argue that a local BA only is sufficient to maintain a good quality map as such perform a Local Bundle Adjustment (LBA), over a local
number of keyframes (also known as windowed optimization);

To reduce the computational expenses of bundle
adjustment, \cite{strasdat_2011_ICCV} proposed to represent the monocular SLAM map by both a Euclidean
map for LBA, and a topological map for pose graph optimization that explicitly distributes the
accumulated drift along the entire map. PGO is best described by: 
%------------------------------------------------------------------------------------------
\begin{equation}
	    \underset{T}{\mathrm{argmin}}\sum_{i=1}^{N}\sum_{j\in S_i}Obj(e(T_i,X_j)).
	    \label{eq:PGO}
\end{equation}
where the optimization process take place over the keyframe poses only ($T_i$). 

\begin{figure}[!htb]
		\centering
		\includegraphics[width=0.45\textwidth]{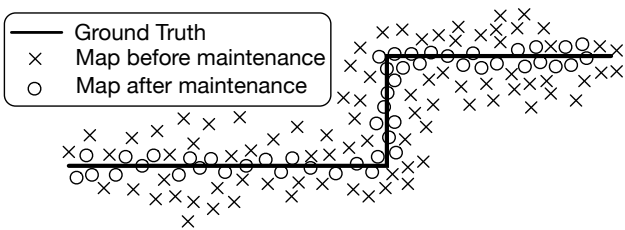}
		\caption{Map maintenance effects on the map.}
		\label{fig:maintenance}
\end{figure}

Figure \ref{fig:maintenance} shows the map maintenance effect, where the scene's
map is refined through outlier removal and error minimizations, in order to yield a more accurate scene
representation. 

%------------------------------------------------------------------------------------------
%Map maintenance is also responsible for detecting and removing outliers in the map due to noisy and faulty matched features. While the underlying assumption of most monocular SLAM algorithms is that the environment is static, some algorithms such as RD SLAM exploit map maintenance methods to accommodate slowly varying scenes (lighting and structural changes).  
%------------------------------------------------------------------------------------------
%------------------------------------------------------------------------------------------
\subsection{Global Localization}
Global localization is required when the camera loses track of its position and is required to
situate itself in a global map. Failure recovery and loop closure are both considered a form of
global localization. It is noteworthy to mention that loop closure and failure recovery revolve
around the same problem and solutions put forward to any of them could be used for the other. The
interested reader is referred to \cite{garcia_2015_ras} for a detailed survey on the topic. While
there is no generic strategy to handle the global localization module, its various implementations
are further discussed in Section \ref{sec:design}. For now, we will limit the extent of the discussion to the problems global localization attempts to adress. 
\subsubsection{Failure recovery}\label{sec:recovery}
Whether due to wrong user movement, such as abrupt changes in the camera pose resulting in motion
blur, or due to observing a featureless region, or for any other reason, monocular SLAM methods may
eventually fail. An example of monocular SLAM failure is shown in Fig. \ref{fig:building}-M, where due so sudden motion, the previously observed scene went out of view. Accordingly, a key module essential for the usability of any monocular SLAM system is
its ability to correctly recover from such failures. 
    %The failure problem shown in Figure \ref{fig:Recovery} is described as a lost camera pose that
    %needs to re-localize itself in the map it had previously created of its environment before
    %failure. For such purpose, different algorithms employ different methods. 
	%----------------------------------------------------
	%\begin{figure}[!hbt]
	%	\centering
	%	\includegraphics[width=3.5in]{FailRecovery}
    %	\caption{Failure recovery definition: The current camera pose is required without knowledge
    %	of the previous camera pose nor data associations are available (n.a.).}
	%	\label{fig:Recovery}
	%\end{figure}

%----------------------------------------------------

\subsubsection{Loop closure}\label{sec:closure}
Since keyframe-based monocular SLAM is an optimization problem, it is prone to drifts in camera pose estimates.
Returning to a certain pose after an exploration phase may not yield the same camera pose
measurement, as it was at the start of the run. (See Fig. \ref{fig:building}-N).
 Such camera pose drift can also manifest itself in a map scale drift, which will eventually lead the system to
erroneous measurements, and fatal failure. To address this issue, some algorithms detect loop
closures in an online monocular SLAM session, and optimize the loops track in an effort to correct the
drift and the error in the camera pose and in all relevant map data that were created during the
loop. The loop closure thread attempts to establish loops upon the insertion of a new keyframe, in
order to correct and minimize any accumulated drift by the system over time using either PGO or BA; the implementations of such optimizations has been made easier using libraries such as G2o \citep{kummerle_2011_ICRA} and Ceres \citep{agarwal_2013_ceres}. 

%------------------------------------------------------------------------------------------
%\begin{figure}[!htpb]
%		\centering
%		\includegraphics[width=2.5in]{LoopClosure.pdf}
%		\caption{Drift suffered by the Visual SLAM pose estimate after returning to its starting point.}
%		\label{fig:LoopClosure}
%\end{figure}
%------------------------------------------------------------------------------------------

%*************************************  DESIGN   **************************************************
\section{Design choices}\label{sec:design}
Now that we've established a generic framework for KSLAM, this section details the design choices, made by different KSLAM systems in the literature.
Table \ref{tab:contributions} lists all the KSLAM systems that, to our knowledge, exist to date; they are categorized as being either open-source or closed-source. Given the additional insight we
gained by having access to their code, we will delve deeper into the open-source systems than we will in those that are closed-source. While this section will be devoted to the details of open-source systems, including PTAM,
SVO, DT SLAM, LSD SLAM, ORB SLAM, DPPTAM and DSO; the closed source systems will be touched upon in Section \ref{sec:ClosedSource}.

%%%%%Comment out for longer version of this document%%%%%
%In 2007, short for Parallel Tracking and Mapping, PTAM \citep{klein_2007_ISMAR} was released, and
%since then many variations and modifications of it have been proposed, such as in
%algorithm to successfully separate tracking and mapping into two parallel computation threads that
%run simultaneously and share information whenever necessary.  This separation made the adaptation of
%off-line Structure from Motion (SfM) methods possible within PTAM, in real-time. Its
%ideas were revolutionary in the monocular SLAM community, and the notion of separation
%between tracking and mapping became the standard backbone of almost all monocular SLAM algorithms
%thereafter.
 %%%%%%%%%%%%%%%%%%%%%%%%
%------------------------------------------------------------------------------------------ 
%\newgeometry{left=1.5cm,bottom=2cm,top=1.5cm}
\definecolor{gray}{rgb}{0.75, 0.75, 0.75}
	\begin{singlespace}
\begin{table*}[!ht]
	\renewcommand{\arraystretch}{1.1}
	\centering
	\caption{Keyframe-based visual SLAM systems, with seven open-source, and sixteen
    closed-source systems }\label{tab:contributions}
	\centering
	\begin{tabular}{| c | p{12cm} | c | l | }
		\hline
		Year &Name  & Closed/ Open& Reference  \\
        \hline
		2006& Real-time Localization and 3D Reconstruction &closed&\cite{mouragnon_2006_CVPR} \\ 
		\hline
		%\rowcolor{gray}
		2007& Parallel Tracking and Mapping (PTAM)
        &open &\cite{klein_2007_ISMAR}\\
		\hline
		2008&An Efficient Direct Approach to Visual SLAM
        &closed   &\cite{silveira_2008_TRO}\\
		\hline
		%\rowcolor{gray}
		2010&Scale Drift-Aware Large Scale Monocular SLAM
        &closed &\cite{strasdat_2010_MIT}\\
		\hline
		2010&Live dense reconstruction with a single moving camera & closed
        &\cite{newcombe_2010_CVPR}\\
		\hline
		%\rowcolor{gray}
		2011&Dense Tracking and Mapping in Real-Time(DTAM)
        & closed  &\cite{newcombe_2011_ICCV}\\
		\hline
		2011&Omnidirectional dense large-scale mapping and navigation based on meaningful
        triangulation &closed  &\cite{pretto_2011_ICRA}\\
		\hline
		%\rowcolor{gray}
		2011&Continuous localization and mapping in a dynamic world (CD SLAM)
        &closed &\cite{pirker_2011_IROS}\\
		\hline
		2011&Online environment mapping & closed &\cite{lim_2011_CVPR}\\
		\hline
		%\rowcolor{gray}
		2011&Homography-based planar mapping and tracking for mobile phones
        &closed&\cite{pirchheim_2011_ISMAR}\\
		\hline
		2013& Robust monocular SLAM in Dynamic environments (RD SLAM)
        &closed &\cite{tan_2013_ISMAR}\\
		\hline
		%\rowcolor{gray}
		2013& Handling pure camera rotation in keyframe-based SLAM (Hybrid SLAM)  & closed
        &\cite{pirchheim_2013_ISMAR}\\
		\hline
		2014&Efficient keyframe-based real-time camera tracking&closed&\cite{dong_2014_IJCV}\\
		\hline
		%	\rowcolor{gray}
		2014& Semi-direct Visual Odometry (SVO)
        &open   &\cite{forster_2014_ICRA}\\
		\hline
		2014& Large Scale Direct monocular SLAM (LSD SLAM)
        &open   &\cite{engel_2014_ECCV}\\
		\hline
		%\rowcolor{gray}
		2014& Deferred Triangulation SLAM (DT SLAM)
        &open & \cite{herrera_2014_IC3DV}\\
		\hline
		2014&Real-Time 6-DOF Monocular Visual SLAM in a Large Scale Environment
        &closed&\cite{lim_2014_ICRA}\\
		\hline
		%	\rowcolor{gray}
		2015&Robust large scale monocular Visual SLAM
        &closed &\cite{bourmaud_2015_CVPR}\\
		\hline
		2015& ORB SLAM  &open &\cite{mur-artal_2015_TRO}\\ 
		\hline
		%	\rowcolor{gray}
		2015&Dense Piecewise Parallel Tracking and Mapping (DPPTAM)
        &open   &\cite{concha_2015_IROS}\\
		\hline
		2016&Multi-level mapping: Real-time dense monocular SLAM &closed   &\cite{greene_2016_ICRA}\\
		\hline
		%	\rowcolor{gray}
		2016&Robust Keyframe-based Monocular SLAM for Augmented Reality &closed
        &\cite{liu_2016_ISMAR}\\
		\hline
		2016&Direct Sparse Odometry (DSO) &open  &\cite{engel_2016_ARXIV}\\
		\hline
\end{tabular}
\end{table*}
	\end{singlespace}
\subsection{ Data association}
\label{Dir/InD}
\subsubsection{Data association design choices}
Table \ref{T1} summarizes the design choices for the data association used by open source KSLAM systems; 
\begin{figure}[!htb] %htb
	\centering
	\includegraphics[width=3in]{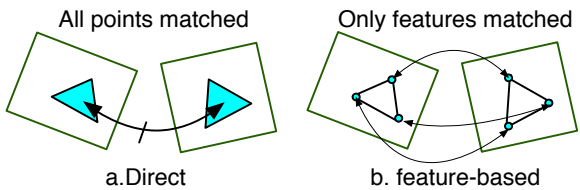}
	\caption{Data association design choices: Direct vs. feature based methods}
	\label{fig:MatchPara}
\end{figure}
%From the list of open-source feature-based methods (systems that adopt the data association design of Fig. \ref{fig:MatchPara}b) surveyed in this paper, 
\textbf{PTAM} and \textbf{DT
SLAM} use FAST features \citep{rosten_2006_ECCV} associated with a local patch of pixels as
descriptors; \textbf{ORB SLAM} uses ORB features with associated ORB descriptors
\citep{rublee_2011_ICCV}. These are considered filter-based, as shown in Fig. \ref{fig:MatchPara}b; in contrast, direct methods, that adopts the data association design of Fig. \ref{fig:MatchPara}a, such as \textbf{LSD SLAM} and
\textbf{DPPTAM}, extract and make use of \textit{all} pixels that have a photometric gradient. \textbf{DSO} argues that using all the pixel information with a photometric gradient
introduces redundancy in the system, and requires a regularization step; therefore, it suggests to
subsample the pixels by dividing the image into blocks, keeping a fixed number of pixels with the
highest gradient in each block. This ensures that first, the sampled pixels are well distributed across the
image, and second, the sampled pixels have sufficiently high image gradients with respect to their
immediate  surroundings.  The sampled pixels are referred to as candidate points. Different than other systems, \textbf{SVO} employs a hybrid approach in which it sequentially alternates between direct and feature-based methods.
%------------------------------------------------------------------------------------------
\newcolumntype{L}{>{\centering\arraybackslash}m{0.7cm}}
	\begin{table*}[hbtp]%!hbt
		\renewcommand{\arraystretch}{1.3}
		\caption{Method used by different monocular SLAM systems. Abbreviations used: indirect
			(i), direct (d), and hybrid (h)
			\label{T1}}
		\centering
		\begin{tabular}{c c c c c c c c}
			\hline
			System&PTAM &SVO&DT SLAM& LSD SLAM& ORB SLAM & DPPTAM & DSO\\
			\hline
			Method & I & H & I & D & I & D & D\\
			\hline
		\end{tabular}
\end{table*}
%------------------------------------------------------------------------------------------

\subsubsection{Data association types:}
Table \ref{tab:association} summarizes the feature extractors and their corresponding descriptors employed by various
open-source KSLAM systems.
%------------------------------------------------------------------------------------------  
\begin{table*}[hbtp]%!hbt
		\small\sf\centering
		\caption{Feature extractors and descriptors. Abbreviations:local patch of
			pixels (L.P.P.), intensity gradient (I.G.)
			\label{tab:association}}
		%\FloatBarrier
		\begin{tabular}{l c c c c c c c}%l m{1.2cm} m{1.2cm} m{1.4cm} m{1.6cm} m{1.8cm} m{1.2cm} m{1.2cm}
			\toprule
			System&{PTAM} & {SVO} &DT SLAM&LSD SLAM& ORB SLAM & DPPTAM&DSO\\
			\midrule
			Feature type & FAST & FAST &  FAST & I.G. & FAST& I.G.  &I.G.\\
			\midrule
			Feature descriptor& L.P.P. & L.P.P. & L.P.P.& L.P.P. & ORB & L.P.P. &L.P.P.\\
			\bottomrule
		\end{tabular}
\end{table*}
%------------------------------------------------------------------------------------------  

\begin{figure}[!bt] %htb
	\centering
	\includegraphics[width=0.4\textwidth]{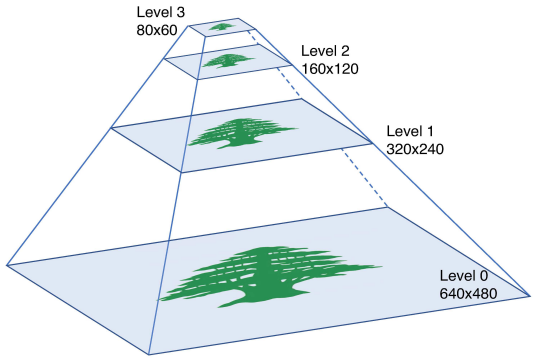}
	\caption{A 4-level pyramid representation of an image. Each level is generated by blurring the level before it, and sub-sampling it by a factor of 2.}
	\label{fig:Pyr}
\end{figure}

\textbf{PTAM} %%%After a successful initialization of PTAM, a
generates a $4$ level pyramid representation of every incoming frame, as shown in Fig. \ref{fig:Pyr}, %%%(\emph{e.g.}, level 1: 640x480, level 2: 320x240), 
and uses it to enhance the features robustness to scale changes, and to increase the
convergence radius of the pose estimation module.
 FAST features are extracted at each level with a Shi-Tomasi score \citep{shi_1994_CVPR} for each feature is estimated as a measure of the feature's saliency. Features with a relatively smaller score are removed before non-maximum suppression takes place.
%%%Ideally, a large number of features across all pyramid levels is recommended; however, To keep the system computationally tractable, and to reduce the effect of non-salient features from degrading system performance, features having a Shi-Tomasi score below a threshold are removed before non-maximum suppression takes place. Fine tuning of the threshold parameters for every pyramid level, in the designated environment of operation, would generally yield improved tracking quality
%%%%Longer version%%%%	
%Each pyramid level has a different threshold for both the Shi-Tomasi score and non-maximum
%suppression, thereby giving control over the strength and the number of features to be tracked
%across the pyramid levels. Fine tuning of these parameters in the designated environment of
%operation would generally yield improved tracking quality.
%%%%
Once 2D features are extracted, 3D landmarks are projected onto the new frame, using a pose estimate prior (from motion model). 3D-2D data association is then employed.
%%%, and feature correspondences are established within a search window surrounding the projected landmark's location. 
The descriptor used for data association is extracted from the 2D image
from which the 3D landmark was first observed. To take into account viewpoint changes, the
landmark's local patch of pixels descriptor is warped through an affine projection, which simulates
how it would appear in the current frame. This however constitutes a limitation in PTAM, since, for
large changes in camera viewpoints, the warping transform fails to accurately reflect the correct
distortion, thereby causing data association failure.
%Systems such as \cite{tan_2013_ismar} disregard features that have a significant view point
%difference between the currently tracked frame and their originating keyframe.
%\citep{tan_2013_ismar}; other systems such as 

When processing a new frame $T_i$, \textbf{DT SLAM} estimates a  2D similarity transform through image registration with the previous frame $T_{i-1}$, and transforms, using the estimated 2D similarity, features extracted from $T_i$ into $T_{i-1}$. 3D landmarks are then projected onto $T_{i-1}$ and data association takes place, similar to how it is done in PTAM. DT SLAM also keeps track of 2D landmarks, which are features that were previously observed but
were not triangulated into 3D landmarks due to the lack of parallax between the different frames
observing them (\ie when the camera  undergoes a pure rotation motion). For every 2D landmark, the Euclidean distance between its epipolar line and the transformed feature is estimated; if it falls below a threshold, the feature is considered as a potential match to the 2D landmark. Data association through Zero Mean Sum of Squared Distance (ZMSSD) is then attempted to validate the matches.

\textbf{SVO} generates a five level pyramid representation of the incoming frame; data
association is first established through iterative direct image alignment, starting from
the highest pyramid level up until the third level. Preliminary data association from this step is used as a prior to a FAST feature matching procedure, similar to PTAM's warping technique, with a Zero-Mean SSD score.

\textbf{ORB SLAM} extracts FAST corners throughout eight pyramid levels. To ensure a
homogeneous distribution along the entire image, each pyramid level is divided into cells and the
parameters of the FAST detector are tuned online to ensure a minimum of five corners are extracted per cell. A 256-bit ORB descriptor is then computed for each extracted feature. 
%%%The higher order feature descriptor ORB is used to establish correspondences between features.
ORB SLAM discretizes and stores the descriptors into bags of words, known as visual vocabulary \citep{lopez_2012_TRO},
which are used to speed up image and feature matching by constraining those features that belong to the same node in the vocabulary tree.  
To deal with viewpoint changes, ORB SLAM proposes to keep track of all the keyframes in which a landmark is observed and choose the descriptor from the keyframe that has the smallest viewpoint difference with the current frame.

\textbf{DSO} Candidate points, sampled across the image, are  represented by eight pixels spread
around the target point. DSO claims that using this number of pixels in a specific pattern was
empirically found to return a good trade-off between three objectives: computational time, sufficient information for tracking to take place, and resilience to motion blur. Each of the selected pixels around the candidate point contributes to the energy functional, which it seeks to
minimize during tracking. Within this formulation, data association is still inherent from the
direct image alignment scheme; however, using only the candidate points and their selected
surrounding pixels, as opposed to using all pixels with gradients in an image. The added value this
approach has over regular keypoint detectors is its adaptive ability to sample candidate points in
low textured regions of the image.

%------------------------------------------------------------------------------------------
\subsection{Visual initialization: choices made}
\label{init_choi}
\begin{figure*}[!hbt] %htb
	\centering
	\includegraphics[scale=1]{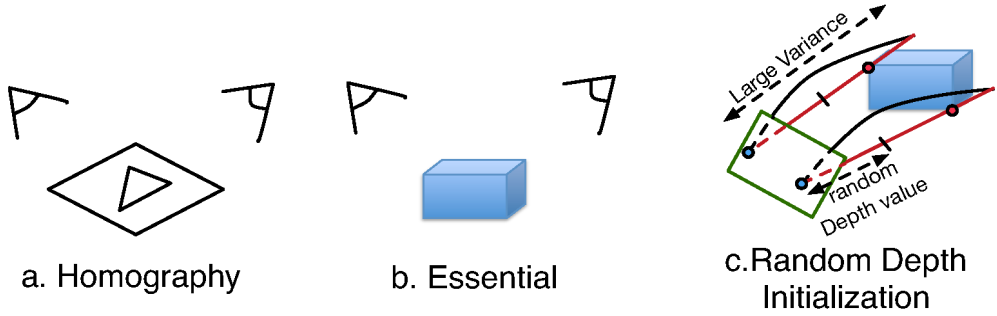}
	\caption{Different initialization methods used in KSLAM. (a) Homographies assume the scene to be planar; (b) The Essential matrix assumes the scene to be non-planar. (c) Random depth initialization, assign a random depth value to the first keyframe and \textit{hope} they converge to a correct configuration in subsequent frames.}
	\label{fig:Init2}
\end{figure*}

%%%%%Longer version
%Given the set of possible degeneracies of the aforementioned initializations, some systems suggest
%initializing SLAM by assigning to all landmarks in the first frame, a random uniform depth
%measurement. Both the camera poses and the depth measurements are subsequently refined as the
%camera moves around the scene. However, such approaches require processing a significant number of
%frames before the system converges to a correct camera-scene configuration; thereby resulting in a period of unreliable tracking. 
The following section discusses the choices made for the initialization of KSLAM in light of Fig. \ref{fig:Init2}, which graphically depicts the different initialization options. Table \ref{tab:initial} summarizes the initialization methods employed by different open source KSLAM  systems, along with their association assumption of the observed scene at startup.

%------------------------------------------------------------------------------------------
\newcolumntype{L}{>{\centering\arraybackslash}m{0.6cm}}
\begin{table*}[!hbt]
		\renewcommand{\arraystretch}{1.3}
		\caption{Initialization. Abbreviations used: Homography
			decomposition (h.d.),
			Essential decomposition (e.d.), Random depth initialization (r.d.), Planar (p), Non-planar (n.p.),
			No assumption (n.a.)\label{tab:initial}}
		\centering
		\begin{tabular}{c c c c c c c c} %m{4cm}
			\hline
			System&PTAM& SVO &DT SLAM&LSD SLAM& ORB SLAM & DPPTAM&DSO\\
			\hline
			Initialization & h.d. & h.d. &  e.d. & r.d. & h.d.+e.d. & r.d.& r.d.\\
			\hline
			Initial scene assumption& p & p & n.p. & n.a. & n.a.& n.a.&n.a.\\
			\hline
		\end{tabular}
\end{table*}
%------------------------------------------------------------------------------------------

\textbf{PTAM}'s \citep{klein_2007_ISMAR} initial release suggested using the five-point
algorithm \citep{nister_2004_PAMI} to estimate and decompose a Fundamental matrix into an $SE(3)$
transformation relating both initializing keyframes; the transformation is then used to triangulate an assumed non-planar initial scene. PTAM's initialization was later changed to the usage of a Homography
\citep{faugeras_1988_IJPRAI}, where the scene is assumed to be composed of 2D planes. PTAM's
initialization requires the user's input to capture the first two keyframes in the map;
furthermore, it requires the user to perform, in between the first and the second keyframe, a slow, smooth and relatively significant translational motion parallel to the observed scene. 
%%%FAST Features extracted from the first keyframe are tracked in a 2D-2D data association scheme in each incoming frame, until the user selects the second keyframe.  
As the 2D-2D matching procedure takes place via ZMSSD without warping the features, establishing correct matches
is susceptible to both motion blur, and significant appearance changes of features as a result of
camera rotations; hence, the strict requirements on the user's motion during the initialization. 
%%%To ensure minimum false matches, the features are searched for twice: once from the current frame to the previous frame, and a second time in the opposite direction. If the matches in both directions are not coherent, the feature is discarded.   Once the second keyframe is successfully incorporated into the map, the Homography relating both keyframes is estimated; then, the corresponding camera poses are determined.
The generated initial map is scaled such as the estimated translation between the first two
keyframes corresponds to 0.1 units, before structure-only BA takes place. 
%%%The mean of the 3D landmarks is selected to serve as the world coordinate frame, while the positive z-direction is chosen such that the camera poses reside along its positive side.	
	
\textbf{SVO}, \citep{forster_2014_ICRA} adopted a Homography for initialization with the same procedure as PTAM;
%%%however, SVO requires no user input and it selects the first acquired image as the first keyframe;
SVO extracts FAST features and tracks them using KLT (Kanade-Lucas-Tomasi feature tracker)
\citep{tomasi_1991_IJCV} across incoming frames.  To avoid the need for
a second input by the user, SVO monitors the median of the baseline distance of the features, tracked between
the first keyframe and the current frame; and whenever this value reaches a certain threshold,
sufficient parallax is assumed, and the Homography can be estimated.
%%%The camera poses are then determined and the landmarks are triangulated and used to estimate an initial scene depth. Bundle adjustment is then applied for the two frames and all their associated landmarks, before the second keyframe is adopted and passed to the map management thread.

\textbf{DT SLAM} does not have an explicit initialization phase; rather, it is integrated
within its tracking module as an Essential matrix estimation method.
		
\cite{engel_2014_ECCV} suggested in \textbf{LSD SLAM}, and later in \textbf{DSO}, a randomly
initialized scene's depth from the first viewpoint, %%%that is later refined through measurements across subsequent frames. 
Both systems use an initialization method that does not require two view
geometry; %Instead of tracking features across two frames, as the other systems do, 
it takes place on a single frame: pixels of interest (\emph{i.e.}, image
locations that have high intensity gradients) in the first keyframe are given a random depth value with an associated large variance. This results in an initially erroneous 3D map. The pose estimation methods are then invoked to estimate the pose of newly incoming frames using the erroneous map, which in return results in erroneous pose estimates. However, as the system process more frames of the same scene, the originally erroneous depth map converges to a stable solution. The initialization is considered complete when the depth variance of the initial scene converges to a minimum.
	
\textbf{DPPTAM}, \citep{concha_2015_IROS} borrows from LSD SLAM's initialization procedure,
and therefore also suffers from the problem of random depth initialization, where several keyframes
must be added to the system before a stable configuration is reached.
	
\textbf{ORB SLAM} deals with the limitations arising from all the above methods by computing, in parallel, both a Fundamental matrix and a
Homography \citep{mur-artal_2015_TRO} ; in order to select the appropriate model, each model is penalized
according to its symmetric transfer error \citep{hartley_2003_Cambridge}. %%%Then, decomposition takes place, and both the scene structure and the camera poses are recovered, before a bundle adjustment step optimizes the map.
If the chosen model yields poor tracking quality, and too few feature
correspondences in the upcoming frame, the initialization is discarded, and the system restarts
with a different pair of frames. 
% It is noteworthy to mention that the relationship between image coordinates and corresponding
    % 3D point coordinates in all the listed initialization methods, aside that of monoSLAM, can
    % only be determined up to an unknown scale $\lambda$.
  
%------------------------------------------------------------------------------------------
\subsection{Pose estimation: choices made} 
  \begin{figure}[!hbt] %htb
  	\centering
  	\includegraphics[width=0.4\textwidth, trim={0cm 0cm 0cm 0cm},clip]{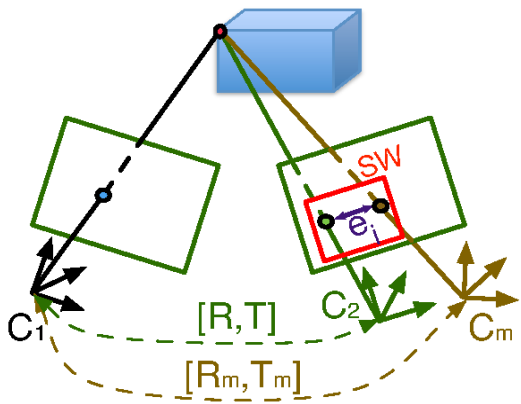}
  	\caption{A generic depiction of the pose estimation module: given a previous pose and a map, the frame of the new pose is first \textit{guessed} using a motion model $C_m$. The motion model bounds the data association into a search window $SW$. An error vector $e_i$ (either geometric or photometric) is then computed and minimized over the initial guess $C_m$ until it converges to the actual pose $C_2$. }
  	\label{fig:PoseEstim2}
  \end{figure}
This section discusses the details of the pose estimation module, as proposed by all open source KSLAM systems, in light of the generic pose estimation process depicted in Fig. \ref{fig:PoseEstim2}.
Table. \ref{tab:pose} summarizes the pose estimation methods used by different monocular SLAM systems.

\textbf{PTAM} defines the camera pose $C$ as an $SE (3)$ transformation that can be minimally represented by six parameters. The mapping from the
full $SE (3)$ transform to its minimal representation $S\xi (3)$ and vice versa can be done through
logarithmic and exponential mapping in Lie algebra \citep{hall_2015_Springer}. The minimally represented $S\xi (3)$ transform
is of great importance as it reduces the number of parameters to optimize from twelve to six,
leading to significant speedups in the optimization process.
	
In PTAM, pose estimation starts by estimating a prior to the frame's pose using a decaying constant velocity motion model. The prior is then refined using a Small Blurry Image (SBI)\textendash the smallest image resolution in the pyramid representation of the frame\textemdash by applying an \emph{Efficient Second Order minimization}
\citep{Benhimane_2007_IJRR}. %%%The velocity of the prior is defined as the change between the current estimate of the pose and the previous camera pose. 
If the velocity is high, PTAM anticipates a fast motion is taking place, and hence, the presence of motion blur and thus restricts tracking to take place only at the highest pyramid levels (most resilient to motion blur) in what is referred to as a coarse tracking stage. Otherwise, the coarse tracking stage is followed by a fine tracking stage. However, when the camera is stationary, the coarse stage may lead to jittering of the camera's pose, and is therefore turned off. 
\newcolumntype{N}{>{\centering\arraybackslash}m{1.8cm}}%1.8cm and m -> p
\newcolumntype{V}{>{\centering\arraybackslash}m{1.8cm}}%2.3cm m->p
%------------------------------------------------------------------------------------------
\begin{table*}[!hbt]
		\small\sf\centering
		%\FloatBarrier
	\caption{Pose estimation. Abbreviations are as follows:
			constant velocity motion model (c.v.m.m), same as previous pose (s.a.p.p.), similarity
			transform with previous frame (s.t.p.f.), optimization
			through minimization of geometric error(o.m.g.e.),  optimization through minimization of photometric error
			(o.m.p.e.), Essential matrix decomposition (E.m.d.), pure rotation estimation from 2 points
			(p.r.e.), significant pose change (s.p.c.), significant scene appearance change (s.s.a.c)
			\label{tab:pose}}
	\begin{tabular}{m{1.8cm} N N V N N N N}%change m to 1.8cm
			\toprule
			System&PTAM & SVO &DT SLAM&LSD SLAM& ORB SLAM &DPPTAM& DSO\\
			\midrule
			Motion prior & c.v.m.m. +ESM & s.a.p.p.  & s.t.p.f. & s.a.p.p.&c.v.m.m. or place recogn.&
			c.v.m.m. or s.a.p.p.&multiple c.v.m.m.\\
			\midrule
			Tracking& o.m.g.e.& o.m.p.e. & e.m.d. or o.m.g.e. or p.r.e.& o.m.p.e. &o.m.g.e.&o.m.p.e.&o.m.p.e. and o.m.g.e.\\
			\midrule
			keyframe add criterion& s.p.c. & s.p.c.  & s.s.a.c. & s.p.c. &s.s.a.c. &s.p.c.&s.s.a.c. or s.p.c.\\
			\bottomrule
	\end{tabular}
\end{table*}
%------------------------------------------------------------------------------------------
The initial camera pose prior is then refined by minimizing a tukey-biweight
\citep{maronna_2006_Wiley} objective function of the re-projection error that down-weights
observations with large residuals. %%%If fine tracking is to take place, features from the lowest pyramid levels are selected, and a similar procedure to the above is repeated.
To determine the tracking quality, PTAM monitors the ratio of
successfully matched features in the frame, against the total number of attempted feature matches. %%%If the tracking quality is questionable, the tracking thread operates normally but no keyframes are accepted by the system.  If the tracker's performance is deemed bad for three consecutive frames, the tracker is considered lost and failure recovery is initiated.%%%%%%%%%% 

\textbf{SVO} assumes the pose of the new frame to be the same as the previous one;%%%SVO uses a sparse, model-based image alignment, in a pyramidal scheme, in order to estimate an initial camera pose estimate. It starts by assuming the camera pose at time $t$ to be the same as at $t-1$
 it then searches for the transformation that minimizes the photometric error of the image pixels with associated depth measurements in the current frame, with respect to their location in the previous one.
The minimization takes places through thirty Gauss Newton iterations of the inverse compositional image alignment method. 
	
SVO does not employ explicit feature matching for every incoming frame; rather, it is performed implicitly as a byproduct of the image alignment step. Once image alignment takes place, landmarks
that are expected to be visible in the current frame, are projected onto the image. To decrease the computational complexity and to maintain only the strongest features, the frame is divided into
a grid, and only the strongest feature per grid cell is used. The 2D location of the projected landmark is fine-tuned by minimizing the photometric error between its associated patch from its location in the current frame, and a warp of the landmark generated from
the nearest keyframe observing it. This minimization violates the epipolar constraint for
the entire frame, and further processing in the tracking module is required: motion-only bundle
adjustment takes place, followed by a structure only bundle adjustment that refines the 3D
location of the landmarks, based on the refined camera pose. Finally, a joint (pose and structure)
local bundle adjustment fine-tunes the reported camera pose estimate. During this last stage, the
tracking quality is continuously monitored and, if the number of observations in a frame, or the number of features between consecutive frames drop,
tracking quality is deemed insufficient, and failure recovery methods are initiated.
    %The camera pose estimation is also responsible for flagging certain frames as keyframes. A
    %frame is selected as a keyframe in SVO if its distance (pose) to the nearest keyframe is larger
    %than a threshold proportional to the mean of the observed scenes depth in the current frame. 

\textbf{DT SLAM} maintains a camera pose based on three tracking modes: full pose
estimation, Essential matrix estimation, and pure rotation estimation. When a sufficient number of
3D matches exists, a full pose can be estimated; otherwise, if a sufficient number of 2D matches
that exhibit small translations is established, an Essential matrix is estimated; and finally, if a
pure rotation is exhibited, two points are used to estimate the absolute orientation of the matches
\citep{kneip_2012_eccv}. Pose estimation aims, in an iterative manner, to
minimize the error vector of both 3D-2D re-projections, and 2D-2D matches. When tracking failure
occurs, the system initializes a new map and continues to collect data for tracking in a different
map; however, the map making thread continues to look for possible matches between the keyframes of
the new map and the old one, and once a match is established, both maps are fused together, thereby
allowing the system to handle multiple sub-maps, each at a different scale. 
	
The tracking thread in \textbf{LSD SLAM} is responsible for estimating the pose of the current frame
with respect to the currently active keyframe in the map, using the previous frame pose as a
prior. The required pose is represented by an $SE (3)$ transformation, and is found by an iteratively
re-weighted Gauss-Newton optimization that minimizes the variance normalized photometric residual error, as described in \cite{engel_2013_ICCV}. A keyframe is considered active if it is the most recent keyframe accommodated in the map.
To minimize outlier effects, measurements with large residuals are down-weighted from one iteration to the next.
	
Pose estimation in \textbf{ORB SLAM} is established through a constant velocity motion
model prior, followed by a pose refinement using optimization.  As the motion model is expected to be easily violated through abrupt motions, ORB SLAM detects such failures by tracking the number of matched features; if it falls below a certain threshold, map points are projected onto the current
frame, and a wide-range feature search takes place around the projected locations. 
%%%If tracking fails,  ORB SLAM invokes its failure recovery method to establish an initial frame pose via global re-localization. 
	
In an effort to make ORB SLAM operate in large environments, a subset of the global map, known as
the local map, is defined by all landmarks corresponding to the set of all keyframes that share
edges with the current frame, as well as all neighbors of this set of keyframes from the pose graph. The selected landmarks are filtered out to keep only the
features that are most likely to be matched in the current frame. Furthermore, if the distance from
the camera's center to the landmark is beyond the range of the valid features, the landmark
is also discarded. The remaining set of landmarks is then searched for and matched in the current
frame, before a final camera pose refinement step .
    %Thetracking thread is also responsible for labeling the current frame a keyframe or not. In
    %contrast to some of the other systems that tends to add keyframes when the camera's pose has
    %moved a significant distance, ensuring a minimum positional change (rotational and
    %translational), ORB SLAM aims to ensure a minimum visual change while spawning as many
    %keyframes as possible to increase the robustness of the tracking algorithm and maintain
    %real-time operation.  Overall, the following must be met so that the current frame is
    %considered a keyframe:
    %------------------------------------------------------------------------------------------
    %\begin{enumerate} \item 20 frames have passed since the last successful global re-localization
    %procedure. 
	%%	\item 20 frames have passed since the insertion of the previous keyframe or the local mapping thread is idle.
    %	\item The current frame tracks successfully more than 50 features.  \item The current frame
    %	track less than $90\%$ of K-ref. 
    %\end{enumerate}
	  
Similar to LSD SLAM, \textbf{DPPTAM} optimizes the photometric error of high gradient pixel locations between two images, using the ICIA formulation over the $SE(3)$ transform relating them.
The minimization is started using a constant velocity motion model, unless the photometric error
increases after applying it.  If the latter is true, the motion model is disregarded, and the pose of
the last tracked frame is used. Similar to PTAM, the optimization takes place in the tangent space
$S\xi(3)$ that minimally parameterizes the rigid body transform by six parameters.

\textbf{DSO} tracking and mapping threads are intertwined; in DSO, all frames are simultaneously
tracked, and used in the map update process; however, each frame contributes differently, and is
treated according to whether it's considered a keyframe or not. DSO uses two parallel threads: a
front-end thread, and a mapping thread. This section will further elaborate on the front-end thread, whereas the
mapping thread will be detailed in Section \ref{sec:subMapGen}.  

DSO front-end initializes the system at startup using random-depth
initialization; it computes the intensity gradients, and tracks the current frame with respect to the currently active
keyframe. Different than other systems, DSO does not use a single frame pose prior; rather, it
attempts a direct image alignment by looping over multiple pose guesses, in a pyramidal
implementation, and removes guesses that yield higher
residuals between iterations. The final pose estimate that yields the
smallest residual error is then assigned to the current frame. The list of initial pose guesses
includes:
\begin{itemize}
	\setlength\itemsep{-3pt}
    \item a constant velocity motion model (CVMM),
    \item a motion model that assumes twice the motion of the CVMM guess,
    \item half of the CVMM guess,
    \item no motion at all (use the pose of the active keyframe) and 
    \item randomly selected small rotations surrounding the CVMM guess.
\end{itemize}
Finally, the front-end checks if the frame should be a keyframe based on one of the following three
conditions: $1^{st}$ when the field of view between the current frame and the last observed keyframe has changed
significantly; $2^{nd}$ when the camera undergoes a significant amount of translation; and $3^{rd}$ if the relative brightness
factor between the two frames changes significantly. A weighted mixture between these conditions 
is used, and compared to a threshold to decide whether a frame becomes a keyframe or not. On
average, around five to ten keyframes are added per second. If desired, one can manually select the
rate of acceptance of keyframe, and disable any of the above selection criteria.

\newcolumntype{L}{>{\centering\arraybackslash}m{1.9cm}}%0.6 cm twocol
%------------------------------------------------------------------------------------------
\subsection{Topological/Metric Map generation: choices made}
\label{sec:subMapGen}
\begin{figure*}[!hbt] %htb
    	\centering
    	\includegraphics[width=1\textwidth]{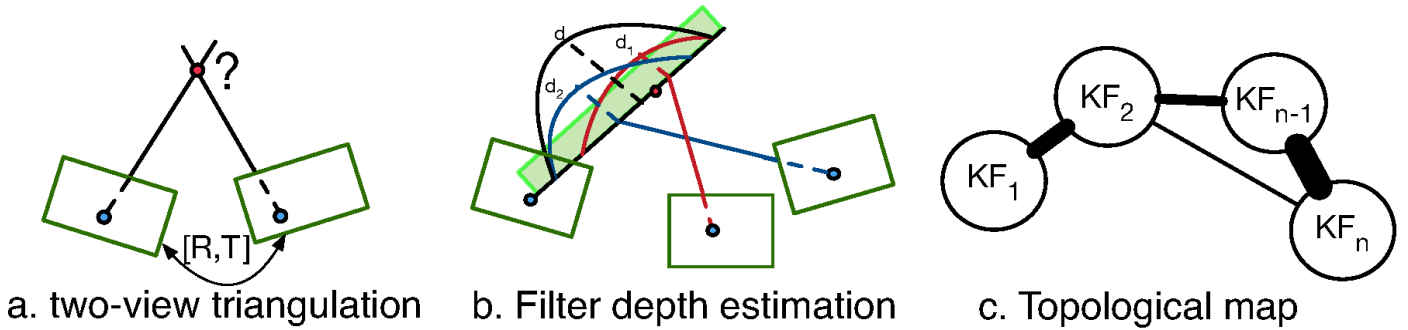}
    	\caption{A generic depiction of the different solutions employed to generate/expand a representation of the observed scene: a) two-view triangulation represents a world surface point by a 3D vector $(x,y,z)^t$. b) Filter depth estimation assumes a world surface point to exist along a semi-infinite ray going from the center of the first camera in which it was observed to infinity, passing by its corresponding 2 dimensional pixel projection.The depth is then found through subsequent filter updates from small baseline observations. c) A Topological map is an undirected graph, that represent keyframes as nodes connected with an edge to other keyframes, with which they share a significant amount of data association.}
    	\label{fig:Triang2}
    \end{figure*}

This section discusses the details of the map generation module, as proposed by all open source KSLAM systems, in light of the generic map expansion process depicted in Fig.\ref{fig:Triang2}.
Table \ref{tab:generation} summarizes map generation methods employed by different monocular SLAM
systems. %%%they can be divided into two main categories: triangulation by optimization (PTAM and ORB SLAM) and filter-based depth estimation (SVO, LSD SLAM, DPPTAM). filter-based depth depth estimation is typically employed in direct systems: it reduces the number of parameters to estimate for a landmark from three (x,y,z coordinates) to one (depth), which allows systems to estimate the depth of the thousands of features per frame in real-time. 
%------------------------------------------------------------------------------------------
\begin{table*}[!hbt]
		\renewcommand{\arraystretch}{1.3}
		\caption{Map generation. Abbreviations: 2 view
			triangulation (2.v.t.), particle filter with inverse depth parametrization (p.f.), 2D landmarks
			triangulated to 3D landmarks (2D.l.t.), depth map propagation from previous frame (p.f.p.f.),
			depth map refined through small baseline observations (s.b.o.), multiple hypotheses photometric error minimization (m.h.p.m.)
			\label{tab:generation}}
		\centering
		\begin{tabular}{c c c c c c c c} %twocolumn change the L to l
			\hline
			System&PTAM&SVO&DT SLAM&LSD SLAM& ORB SLAM & DPPTAM&DSO\\
			\hline
			Map generation & 2.v.t.& p.f. &  2D.l.t. & p.f.p.f and s.b.o.  & 2.v.t.& m.h.p.m&s.b.o. \\
			\hline
			Map type &metric&metric&metric&hybrid & hybrid&metric&metric\\
			\hline
		\end{tabular}
\end{table*}
%------------------------------------------------------------------------------------------

When a new keyframe is added in \textbf{PTAM}, all bundle adjustment operations are halted,
and the new keyframe inherits the pose from the coarse tracking stage. The potentially visible set
of landmarks estimated by the tracker are then re-projected onto the new keyframe, and feature
matches are established. Correctly matched landmarks are marked as seen again; this is done to keep
track of the quality of the landmarks and to allow for the map refinement step to remove corrupt
data. 
	
New landmarks are generated by establishing and triangulating feature matches between the newly
added keyframe and its nearest keyframe (in terms of position) from the map. Landmarks that are
already existent in the map are projected onto both keyframes, and feature matches from the current keyframe are searched for along their corresponding epipolar lines in the second keyframe, at regions that do not contain projected landmarks.  The average depth of the projected landmarks is used to constrain the epipolar search, from a line to a segment.
	
 %%%The map generation thread in SVO runs parallel to the tracking thread, and is responsible for creating and updating the map. 
\textbf{SVO} parametrizes 3D landmarks using an inverse depth parameterization model \citep{civera_2008_TRO}.
Upon insertion of a new keyframe, features possessing the highest Shi-Tomasi scores are chosen to
initialize a number of depth filters.  These features are referred to as seeds, and are initialized
along a line propagating from the camera center to the 2D location of the seed in the originating
keyframe. The only parameter that remains to be solved for is then the depth of the landmark, which
is initialized to the mean of the scene's depth, as observed from the keyframe of origin. 
	
During the times when no new keyframe is being processed, the map management thread monitors and
updates map seeds by subsequent observations, in a fashion similar to \cite{vogiatzis_2011_IJCV}.
The seed is searched for in new frames along an epipolar search line, which is limited by the
uncertainty of the seed, and the mean depth distribution observed in the current frame.  As the
filter converges, its uncertainty decreases, and the epipolar search range decreases. If seeds fail
to match frequently, if they diverge to infinity,  or if a long time has passed since their
initialization, they are removed from the map. 	
%%%The filter converges when the distribution of the depth estimate of a seed transitions from the initially assumed uniform distribution into a Gaussian one. The seed in then added into the map, with the mean of the Gaussian distribution as its depth. 
This process however limits SVO to operate in environments of relatively uniform depth
distributions. Since the initialization of landmarks in SVO relies on many observations in order for
the features to be triangulated, the map contains few,if any, outliers, and hence no outlier
deletion method is required. However, this comes at the expense of a delayed time before the
features are initialized as landmarks and added to the map.
    
Inspired by the map generation module of SVO, \textbf{REMODE} was released by
\citep{pizzoli_2014_ICRA} as a standalone map generation module that takes input measurements from
SVO (Sparse map and pose estimates) and generates a per-pixel dense map of the observed scene, based
on the probabilistic Bayesian scheme suggested in \citep{vogiatzis_2011_IJCV}.

\textbf{DT SLAM} aims to add keyframes when enough visual change has occurred; the three
criteria for keyframe addition are (1) for the frame to contain a sufficient number of new 2D
features that can be created from areas not covered by the map, or (2) a minimum number of 2D
features can be triangulated into 3D landmarks, or (3) a given number of already existing 3D
landmarks have been observed from a significantly different angle.  The map in DT SLAM contains both
2D  and 3D landmarks, where the triangulation of 2D features into 3D landmarks is done through two
view triangulation by optimization, and is deferred until enough parallax between the keyframes is
observed\textemdash hence the name of the algorithm.

\textbf{LSD SLAM}'s map generation module  is mainly responsible for the selection and accommodation of new keyframes into the map. 
Its functions can be divided into two main categories, depending on whether the current frame is a
keyframe or not; if it is, depth map creation takes place by keyframe accommodation as described below; if
not, depth map refinement is done on regular frames. 
%%%Since LSD SLAM is a direct method, it operates on the underlying assumption of small displacements between frames. To maintain tracking quality, LSD SLAM requires frequent addition of keyframes into the map, as well as relatively high frame rate cameras.
	
When the system is accommodating a new keyframe, the estimated depth map from the previous keyframe
is projected onto it, and serves as its initial guess.  Spatial regularization then takes place, by
replacing each projected depth value with the average of its surrounding values, and the variance is chosen as the minimal variance value of the neighboring measurements. 
% The projected depth map is scaled to have a mean inverse depth value of 1. The scaling ratio is then incorporated into the $Sim (3)$ transformation that describes the new keyframe pose in the map. This is done to allow for scale optimization and correction during subsequent steps.
The $Sim (3)$ of a newly added keyframe is then estimated and refined in a direct, scale-drift aware
image alignment scheme with respect to other keyframes in the map, over the seven degree of freedom
$Sim (3)$ transform. %%% in contrast to the 6 DOF $SE (3)$ transform used for the regular tracking procedure.
    % Furthermore, as the photometric error alone does not constrain the scale, the depth error
    % integration into the minimization of the $Sim (3)$ transform is mandatory.  
Due to the non-convexity of the direct image alignment method on $Sim (3)$, an accurate
initialization to the minimization procedure is required; for such a purpose, ESM (Efficient Second
Order minimization) \citep{Benhimane_2007_IJRR} and a coarse to fine pyramidal scheme with very low resolutions proved to increase the convergence radius of the task. 
    %The newly accommodated keyframe is flagged as the currently active keyframe and subsequent
    %frames are tracked according to it. 
%If the map generation module deems the current frame as not being a keyframe, depth map refinement
%takes place by establishing stereo matches for each pixel in a suitable reference frame. The
%reference frame for each pixel is determined by the oldest frame it was observed in, where the
%disparity search range and the observation angle do not exceed a threshold. A 1-D search along the
%epipolar line for each pixel is performed with an SSD metric.
    %to establish pixel matches; the search range is limited by the prior depth value at a given
    %pixel if available; otherwise, the entire range is scanned. 
%To minimize computational cost and reduce the effect of outliers on the map, not all established
%stereo matches are used to update the depth map; instead, a subset of pixels is selected for which
%the accuracy of a disparity search is sufficiently large. The accuracy is determined by three
%criteria: the photometric disparity error, the geometric disparity error, and the pixel to inverse
%depth ratio. Further details regarding these criteria can be found in \cite{engel_2013_ICCV}. Finally, depth map regularization  and outlier handling, similar to the
%keyframe processing step, take place.

\begin{figure}[!hbt] %htb
	\centering
	\includegraphics[width=0.45\textwidth]{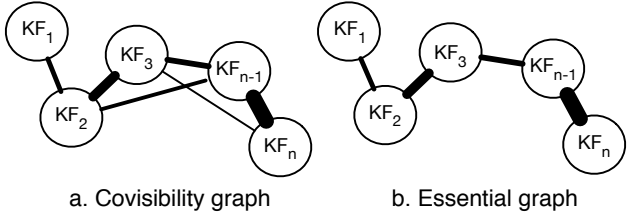}
	\caption{The difference between a. co-visibility graph and b. essential graph. An essential graph is obtained by thresholding the co-visibility graph to keep the strongest two edges per keyframe.}
	\label{fig:Topo2}
\end{figure}

\textbf{ORB SLAM}'s local mapping thread is responsible for keyframe insertion, map point
triangulation, map point culling, keyframe culling, and local bundle adjustment. ORB SLAM
incorporates a hybrid, one metric and two topological maps. The two topological maps, referred to as
co-visibility and essential graphs, are built using the same nodes (keyframes) however, with different edges (connections) between them. The co-visibility graph allows for as many connections as available between nodes; in contrast to the essential graph that allows every node to have at most two
edges, by only keeping the strongest two edges. The difference between them is contrasted in Fig. \ref{fig:Topo2}. 
The mapping thread is responsible for updating the co-visibility and essential graphs with the appropriate edges, as well as computing
the bag of words representing the newly added keyframes in the map.  The metric map is propagated by triangulating new landmarks from ORB features, which appear in at least two nodes connected to the new keyframe in the
co-visibility graph.  To prevent outliers, triangulated landmarks are tested for positive depth,
re-projection error, and scale consistency in all keyframes they are observed in, before finally
incorporating them into the map.

Landmark triangulation in \textbf{DPPTAM} takes place over several overlapping observations
of the scene using inverse depth parametrization; the map maker aims to minimize the photometric
error between a high gradient pixel patch in the last added keyframe, and the corresponding patch of
pixels, found by projecting the feature from the keyframe onto the current frame. The minimization
is repeated ten times for all high-gradient pixels, when the frame exhibits enough translation; the
threshold for translation is increased from one iteration to the next, to ensure sufficient baseline
distance between the frames. The end result is ten hypotheses for the depth of each high-gradient
pixel. To deduce the final depth estimate from the hypotheses, three consecutive tests are
performed, including gradient direction test, temporal consistency, and spatial consistency.

All frames in \textbf{DSO} are used in the map building process; while keyframes are used to expand the map
and perform windowed optimization, regular (non-keyframe) frames are
used to update the depth of the already existing candidate points. DSO maintains two thousand candidate points per keyframe. The estimated pose of subsequent regular
frames, the location of the candidate points in the active keyframe and their variance, are all used
to establish an epipolar search segment in the regular frame. The image location along the epipolar
segment, which minimizes the photometric error, is used to update the depth and the variance of the
candidate point, using a filter-based triangulation, similar to LSD SLAM.  DSO adopts the inverse
depth paradigm as a parameterization for the 3D world which reduces the
parameters to optimize to one variable; thereby reducing computational cost. This estimated depth is
used as a prior for a subsequently activated candidate point in a windowed optimization.
In its active window of optimization, DSO maintain seven \textit{active keyframes}, along with two thousand \textit{active points}, equally
distributed across the active keyframes.  As new keyframes and candidate points are accommodated by
the system, older ones are marginalized: where the number of active keyframes exceeds $7$, the
system chooses a keyframe from the active window and marginalizes it. The choice of the keyframe is
done by maximizing a heuristically designed distance score, which ensures the
remaining active keyframes to be well distributed across the space between the first and last
keyframes in the active window, and closer to the most recently added keyframe.
Also if ninety five percent of a frame's points are marginalized, the frame is dropped out of the system.

%failure nor can detect loop closures. 
%------------------------------------------------------------------------------------------	
\subsection{BA/PGO/Map maintenance: choices made}
\begin{figure*}[!hbt] %htb
	\centering
	\includegraphics[width=0.9\textwidth]{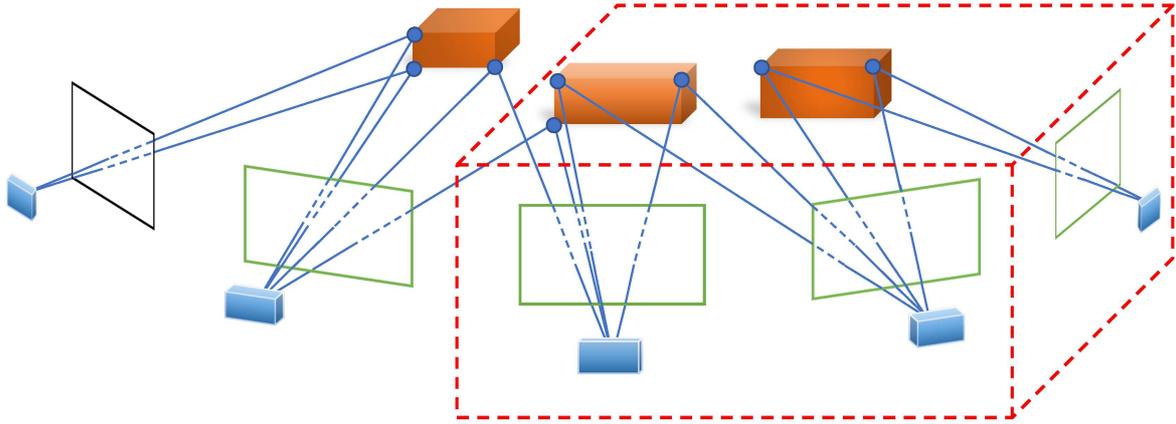}
	\caption{Different optimization strategies: GBA jointly optimizes the entire map whereas LBA jointly optimizes a subset of the map (encapsulated in the red cube). PGO optimizes the poses of a subset of frames from the map (green frames). Structure only bundle adjustment optimizes the 3D position of the landmarks (blue circles) while holding their associated frames poses fixed. }
	\label{fig:Opti2}
\end{figure*}

As detailed in section \ref{sec:maintenance}, there are many variations of the optimization process that can be applied to KSLAM, namely LBA (local bundle adjustment), GBA(global bundle adjustment), PGO (pose graph optimization), and structure only bundle adjustment. These different variations are shown in Fig. \ref{fig:Opti2}.  

Table \ref{tab:maintenance} summarizes the map maintenance procedure adopted by different KSLAM systems. 
%-----------------------------------------------------------------------------------------
\newcolumntype{N}{>{\centering\arraybackslash}p{1.4cm}}%1.8cm for twocolumns
\newcolumntype{X}[1]{>{\raggedright\let\newline\\\arraybackslash\hspace{0pt}}m{#1}}

\begin{table*}[!hbt]
		\renewcommand{\arraystretch}{1.3}
		\caption{Map maintenance. Abbreviations used: Local Bundle
			Adjustment (LBA), Global Bundle Adjustment (GBA), Pose Graph Optimization (PGO), Structure Only Bundle Adjustment (SOBA),
			\label{tab:maintenance}}
		\centering
		\begin{tabular}{m{1.6cm} m{2cm} m{2.05cm} m{1.95cm} m{2cm} m{2cm} m{1.6cm} m{2cm}}
			\hline
			System&PTAM&SVO&DT SLAM&LSD SLAM& ORB SLAM & DPPTAM&DSO\\
			\hline
			Optimization & LBA \& GBA & LBA \& SOBA & LBA \& GBA & PGO& PGO \& LBA& Dense mapping&LBA\\
			\hline
			Scene type &static \& small & static \& uniform depth & static \& small  &
			static \& small or large & static \& small or large& static \& indoor planar&static \& small or large\\
			\hline
		\end{tabular}
\end{table*}

%------------------------------------------------------------------------------------------
%%%The map making thread in PTAM runs parallel to the tracking thread and does not operate on a frame by frame basis; instead, it only processes keyframes. 
When the map making thread is not processing new keyframes, \textbf{PTAM} performs various optimizations and maintenance to the map, such
as an LBA for local map convergence and a GBA for the map's global convergence. The computational cost in PTAM scales with the map and becomes intractable as the number of keyframes
gets large; for this reason, PTAM is designed to operate in small workspaces.  Finally, the optimization thread applies data refinement by first searching and updating landmark observations in
all keyframes, and then by removing all landmarks that failed, many times, to successfully match
features.
	
For runtime efficiency reasons, \textbf{SVO}'s map management maintains only a fixed number of
keyframes in the map and removes distant ones when new keyframes are added. This is performed so
that the algorithm maintains real-time performance after prolonged periods of operation over large distances.
	
Aside from the map generation module, \textbf{DT SLAM} employs a third thread that continuously optimizes the entire map in the background through a sparse GBA. 

\textbf{LSD SLAM} runs a third parallel thread that continuously optimizes the map in the
background by a generic implementation of a pose graph optimization using the g2o-framework
\citep{kummerle_2011_ICRA}. This however leads to an inferior accuracy when compared to other
methods. Outliers are detected by monitoring the probability of the projected depth hypothesis at
each pixel of being an outlier or not. To make the outliers detection step possible, LSD SLAM keeps
records of each successfully matched pixel during the tracking thread, and accordingly increases or
decreases the probability of it being an outlier. %%%During the spatial regularization step in the map generation module, if the probability that all contributing neighbors are outliers is above a given threshold, the depth hypothesis is removed from the system.
 %%%To maintain a good quality map and counter the effect of frequently adding features from keyframes, 

\textbf{ORB SLAM} employs rigorous landmark culling to ensure few outliers in the map. A landmark must be correctly matched to twenty five percent of the frames in which it is
predicted to be visible. It also must be visible from at least three keyframes after more than one
keyframe has been accommodated into the map, since it was spawned. Otherwise, the landmark is
removed. To maintain lifelong operation and to counter the side effects of the presence of a high number of keyframes in the map, a rigorous keyframe
culling procedure takes place as well. Keyframes that have ninety percent of their associated
landmarks observed in three other keyframes are deemed redundant, and removed.  The local mapping
thread also performs a local bundle adjustment over all keyframes connected to the latest
accommodated keyframe in the co-visibility graph, and all other keyframes that observe any landmark
present in the current keyframe.

\textbf{DPPTAM} produces real-time dense maps by employing a dense mapping thread that exploits planar
properties of man-made indoor environments. Keyframes are first segmented into a set of 2D
superpixels, and all 3D landmarks from the map are projected onto the keyframe, and assigned to
different superpixels according to the distance of their projections to the appropriate superpixel
in the keyframe. 3D points belonging to contours of the superpixels are used to fit 3D planes to
each superpixel. To determine if the superpixel's plane is to be added into the map, three test are
performed, including the normalized residual test, the degenerate case detection, and the temporal consistency test. Finally, a full dense map is reconstructed, by estimating depth at every pixel, using depth priors of the 3D planes associated with the superpixels. 

Analogous to LBA, \textbf{DSO} performs a  windowed optimization over the combined photometric (intensity) and
geometric residual of all active points between the set of active keyframes, using six
Gauss-Newton iterations. If the resulting residual of the most recently added keyframe after the
optimization is large, the newly added keyframe is dropped.  Map maintenance in DSO is also
responsible for outlier detection and handling in an early stage of the system. The point is
disregarded and dropped out if one of of the four following conditions is satisfied: 
\begin{enumerate}[topsep=0pt,itemsep=-1ex,partopsep=1ex,parsep=1ex]
    \item a point cannot be distinctively identified from its immediate surroundings,
\item a point  falls out of bound of the search region,
\item a point match is found farther than two pixels from its original location or
\item the point has a relatively high photometric residual with respect to the median residual of the entire frame.
\end{enumerate}

%------------------------------------------------------------------------------------------
\subsection{Failure recovery: choices made}
Table \ref{tab:recovery} summarizes the failure recovery mechanisms used by different monocular SLAM
system. 
%------------------------------------------------------------------------------------------
	\begin{table*}[!hbt]
		\renewcommand{\arraystretch}{1.3}
		\caption{Failure recovery. Abbreviations used: photometric
			error minimization of SBIs (p.e.m.), image alignment with last correctly tracked keyframe
			(i.a.l.), image alignment with random
			keyframe (i.a.r.), bag of words place recognition (b.w.), image alignment with arbitrary small rotations around the last tracked pose(i.a.a.r.)\label{tab:recovery}}
		\centering
		\begin{tabular}{l c c c c c c c}
			\hline
			System&PTAM&SVO&DT SLAM&LSD SLAM& ORB SLAM & DPPTAM& DSO\\
			\hline
			Failure recovery & p.e.m. &i.a.l. &  none & i.a.r. & b.w.&i.a.l&i.a.a.r.\\
			\hline
		\end{tabular}
	\end{table*}
%------------------------------------------------------------------------------------------
    
Upon detecting failure, \textbf{PTAM}'s tracker initiates a recovery procedure, where the SBI of each incoming frame is compared to the database of SBIs for all keyframes.  If the intensity
difference between the incoming frame and its nearest looking keyframe is below a certain threshold,
the current frame's pose is assumed to be equivalent to that of the corresponding keyframe. ESM
tracking takes place to estimate the rotational change
between the keyframe and the current frame. If converged, %%% a PVS of landmarks from the map is projected onto the estimated pose and 
the tracker attempts to match the landmarks to the features in the frame. If a sufficient number of features are correctly matched, the tracker resumes normally; otherwise, a new frame is acquired and the tracker remains lost. 

\textbf{SVO}'s %%% When tracking quality is deemed insufficient, SVO initiates its failure recovery method. 
first procedure in the recovery process is to apply image alignment between the incoming
frame and the closest keyframe to the last known correctly tracked frame. If more than thirty
features are correctly matched during this image alignment step, then the re-localizer considers
itself converged and continues tracking regularly; otherwise, it attempts to re-localize using new
incoming frames. 

\textbf{LSD SLAM}'s recovery procedure first chooses, at random, from the pose graph, a keyframe that
has more than two neighboring keyframes connected to it. It then attempts to align
the currently lost frame to it. If the outlier-to-inlier ratio is large, the keyframe is discarded, and replaced by another keyframe at random; otherwise, all neighboring keyframes connected to it in the pose graph are then tested. If the number of neighbors with a large inlier-to-outlier ratio is larger than the number of neighbors with a large outlier-to-inlier ratio, or if there are more than five neighbors with a large inlier-to-outlier ratio, the neighboring keyframe with the largest ratio is set as the active keyframe, and regular tracking resumes.

Upon running, \textbf{ORB SLAM}'s re-localizer transforms the current frame into a bag of words and queries
the database of keyframes for all possible keyframes that might be used to re-localize from. The
place recognition module implemented in ORB SLAM, used for both loop detection and failure recovery,
relies on bags of words, as frames observing the same scene share a big number of common visual
vocabulary. In contrast to other bag of words methods that return the best queried hypothesis from
the database of keyframes, the place recognition module of ORB SLAM returns all possible hypotheses
that have a probability of being a match larger than seventy five percent of the best match. The
combined added value of the ORB features, along with the bag of words implementation of the place
recognition module, manifest themselves in a real-time, high recall, and relatively high tolerance to
viewpoint changes during re-localization and loop detection. All hypotheses are then tested through
a RANSAC implementation of the PnP algorithm \citep{lepetit_2009_IJCV}, which determines the camera
pose from a set of 3D to 2D correspondences. The camera pose with the most inliers is then
used to establish more matches to features associated with the candidate keyframe, before an
optimization over the camera's pose takes place.

\textbf{DSO} does not have  a global based failure recovery method. When the minimization of
DSO's pose tracking diverges, the last successfully tracked camera pose is used to generate multiple
arbitrary random rotations around it. The generated poses are used in an attempt to localize at the
coarsest pyramid level with the most recent active keyframe; if the photometric minimization
succeeds, regular tracking resumes, otherwise, tracking fails. 
%------------------------------------------------------------------------------------------
\subsection{Loop closure: choices made}
Table \ref{tab:closure} summarizes how loop closure is addressed by each KSLAM system.
%------------------------------------------------------------------------------------------
\begin{table*}[!hbt]
		\renewcommand{\arraystretch}{1.2}
		\caption{Loop closure. Abbreviations used: Bag of Words place
			recognition (B.W.p.r), sim(3) optimization (s.o.)\label{tab:closure}}
		\centering
		\begin{tabular}{c c c c c c c c}
			\hline
			System&PTAM&SVO&DT SLAM&LSD SLAM& ORB SLAM & DPPTAM&DSO\\
			\hline
			Loop closure & none & none & none  & FabMap +s.o. & B.W.p.r. +s.o.&none&none\\
			\hline
		\end{tabular}
\end{table*}

When tracking failure occurs in \textbf{DT SLAM}, it starts a new sub map and
instantly start tracking it, while it invokes, in the background, a loop closure thread that attempts
to establish data associations across different sub-maps. DT SLAM's loop closure module in this
context is a modified version of PTAM's failure recovery module, but employed across sub maps. When
a sufficient number of data associations are successfully established between two keyframes, their corresponding
sub-maps are merged together through a similarity transform optimization.

Whenever a keyframe is processed by \textbf{LSD SLAM}, loop closures are searched for
within its ten nearest keyframes as well as through the appearance based model of FABMAP
\citep{glover_2012_ICRA} to establish both ends of a loop. Once a loop edge is detected, a pose
graph optimization minimizes the similarity error established at the loop's edge, by distributing
the error over the poses of the loop's keyframes.
	
Loop detection in \textbf{ORB SLAM} takes place via its global place recognition module,
that returns all hypotheses of keyframes, from the database that might correspond to the opposing
loop end. 
%%%To ensure enough distance change has taken place, they compute the similarity transform between all connected keyframes to the current keyframe in the thresholded co-visibility graph. On the other hand,  if the similarity score is less than a threshold, the hypothesis of a loop is removed. If a sufficient number of inliers support the refined similarity transform, the queried keyframe is considered to be the other end of the loop and loop fusion takes place.
%%%The loop fusion first merges duplicate map points in both keyframes and inserts a new edge in the co-visibility graph that closes the loop by correcting the $Sim (3)$ pose of the current keyframe, using the similarity transform between the loop ends.  
%%%Using the corrected pose, 
All landmarks associated with the queried keyframe and its neighbors are projected to, and searched for, in all keyframes associated with the
current keyframe in the co-visibility graph. The initial set of inliers, as well as the found
matches, are used to update the co-visibility and Essential graphs, thereby establishing many edges
between the two ends of the loop. Finally, a pose graph optimization over the essential graph takes
place, similar to that of LSD SLAM, which minimizes and distributes the loop closing error along the
loop nodes.

Marginalized keyframes and points in \textbf{DSO} are permanently  dropped out of the system and never used
again; therefore, DSO does not employ any global based localization methods to recover from failure, or detect loop closures.

%*************************************  CLOSED SOURCE   *******************************************
\section{Closed source systems}\label{sec:ClosedSource}
We have discussed so far methods presented in open-source monocular SLAM systems; however, many
closed source methods also exist in the literature that we could not fully dissect due to the limited details presented in their papers. 
 This section aims to provide a quick chronological survey
of these systems, which put forward many interesting ideas for the reader. 
%Table \ref{tab:closed_source} lists in chronological order each of these systems.
To avoid repetition, we will not outline the complete details of each system; rather, we will focus on
what we feel has additive value to the reader beyond the information provided in Section \ref{sec:design}. For the remainder of this section, each paragraph is a concise summary of the main contributions in the closed-source systems.

In 2006, \textbf{\cite{mouragnon_2006_CVPR}} were the first to introduce the concept of keyframes in
monocular SLAM and employed a local Bundle Adjustment in real-time over a subset of keyframes in the
map. To ensure sufficient baseline, the system is initialized by automatic insertion of three
keyframes. 
%the first captured image is taken as Keyframe \#1, the second and third keyframes are chosen when
%the number of tracked 2D-2D features drops below two preset thresholds. 
%%%However, the system does not use the three views to perform the initialization; instead, it solves for the initialization using the 5-point algorithm of \cite{nister_2004_PAMI}, implemented only between the first and third keyframes, and
The second keyframe is only used to ensure that there has been sufficient baseline between the first and the third.\\
%%%therefore, the initialization is still susceptible to the common limitations of other initialization methods.

\textbf{\cite{silveira_2008_TRO}} proposed a real-time direct solution by assuming large patches surrounding regions of high intensity gradients as planar; and then performing image
alignment by minimizing the photometric error of these patches across incoming frames, in a
single optimization step that incorporates Cheirality, geometric, and photometric constraints.
The system employs a photogeometric generative model to gain resilience against lighting
changes, and monitors the errors in the minimization process to flag outliers.\\
% that ensures geometrical validity of the reconstructed scene as well as system's resilience to
% lighting variations through their proposed photogeometric generative model. 
%To detect outliers, the system monitors for the patches of pixels the photometric error from
%the optimization as well as the geometric error induced by significant warping and accordingly
%remove them from the map.

In 2009, \textbf{\cite{dong_2009_IJCV}} suggested, in Keyframe-Based Real-Time Camera Tracking, to perform
monocular SLAM in two steps. The first step is  an offline procedure during which, SIFT features are extracted and triangulated using SfM methods into 3D landmarks. The generated scene is then sub-sampled into keyframes according to a
selection criterion that minimizes a redundancy term and a completeness term. To enhance the performance of subsequent steps, the keyframes descriptors are stored in a vocabulary tree.

During the online session, the system estimates a subset of keyframes from the map, using
the vocabulary tree, with keyframes that share the most features with the current frame. It then employs a parallelized 2D-3D data association to establish an error vector between the
landmarks observed in the keyframes and the current frame, which is used to estimate the current camera pose.

The system was then updated in 2014 in \textbf{\cite{dong_2014_IJCV}} to include a GPU accelerated SIFT
feature extraction and matching,  with a proposed two-pass feature matching procedure.  In the first
pass,  SIFT features from the current frame are matched to the selected keyframes,
using the offline-generated vocabulary tree. This is followed by an outlier rejection
implementation through RANSAC;  if the remaining number of inliers is below a
threshold, the second pass is performed. During the second pass,  the epipolar geometry, found
from the first pass, is used to constrain
local SIFT feature matching between the keyframes and the current frame. The process repeats
until a sufficient number of inliers is found. \\
	
\textbf{\cite{strasdat_2010_MIT}} introduced similarity transforms into monocular SLAM, allowing
for scale drift estimation and correction, once the system detects loop closure.  Feature
tracking is performed by a mixture of top-bottom and bottom-up approaches, using a dense
variational optical flow, and a search over a window surrounding the projected landmarks.
Landmarks are triangulated by updating information filters, and loop detection is performed using
a bag of words discretization of SURF features \citep{bay_2008_CVIU}. 
%2D-2D SURF features are matched between the current frame and the loop frame based on their
%descriptors. The depth of the matched 2D features in the loop frame is then interpolated using
%the depth distribution of nearby regular FAST features and hence 2D-3D SURF matches may be used
%to estimate a 6 d.o.f. loop constraint using the p4p method \cite{lepetit_2009_ijcv}. However,
%the above formulation does not constrain the scale so to account for scale drift, the depth of
%the SURF
%features is then interpolated only this time using the depth of the regular features from the
%current frame so that the scale may be found as the median scale difference over all the
%correspondences. 
The loop is finally closed by applying a pose graph optimization over the similarity transforms
relating the keyframes.\\

\textbf{\cite{newcombe_2010_CVPR}} suggested a hybrid monocular SLAM system that relied on
feature-based SLAM (PTAM) to fit a dense surface estimate of the environment that is refined
using direct methods.  A surface-based model is then computed and polygonized to best fit the
triangulated landmarks from the feature-based front end. A parallel process chooses a batch of
frames that have a potentially overlapping surface visibility, in order to estimate a dense
refinement over the base mesh, using a GPU accelerated implementation of variational optical flow.

In an update to this work, Newcombe published, in 2011, Dense Tracking and Mapping in Real-Time (DTAM)
\textbf{\cite{newcombe_2011_ICCV}} that removed the need for PTAM as a front-end to the system, and generalized the dense reconstruction to fully solve the monocular SLAM pipeline, by performing, online, a dense reconstruction, given camera pose estimates that are found through whole image alignment. \\

Similar to the work of Newcombe \citep{newcombe_2011_ICCV}, \textbf{\cite{pretto_2011_ICRA}} modeled the
environment as a 3D piecewise smooth surface, and used a sparse feature based front-end as a base for a Delaunay triangulation to fit a mesh that is used to interpolate a dense reconstruction of the environment.\\

\textbf{\cite{pirker_2011_IROS}} released CD SLAM in 2011, with the objectives to handle short- and long-term environmental changes and to handle mixed indoor/outdoor environments. 
%To achieve its goals, CD SLAM needs a fast and robust data association in large dynamic scenes
%and to have the map size proportional to the explored space as well as a robust loop closure
%detection and correction mechanism to account for drifts accumulated over long-term
%operation.
%
To limit the map size and gain robustness against significant rotational changes, CD SLAM suggests the use of a modified Histogram of Oriented Cameras descriptor (HOC)
\citep{pirker_2010_BMVC}, with a GPU accelerated descriptor update, and a probabilistic weighting scheme to handle outliers.  Furthermore, it suggests the use of large-scale nested loop closures with scale drift correction, and provide a geometric adaptation to update the feature descriptors after loop closure.  
%The camera pose in CD SLAM is formulated as a similarity transform and every map point is
%associated to a SIFT and a HOC descriptor. 
Keyframes are organized in an undirected, unweighted pose graph.
%To prevent redundant addition of landmarks in the map, in an effort to control the map size and
%gain robustness against rotational changes , HOC descriptors store information for the same
%landmark from different view points.  
Re-localization is performed using a non-linear least square minimization, initialized with the pose of the best matching candidate keyframe from the map, found through FABMAP. \\

\textbf{RD SLAM \citep{tan_2013_ISMAR}} was released with the aim to handle occlusions and slowly varying, dynamic scenes.  RD SLAM employs a heavily parallelized GPU accelerated SIFT and stores them in a KD-Tree \citep{bentley_1975_ACM} that further accelerates feature matching based on the nearest neighbor of the queried feature in the tree. While the KD-tree is meant to accelerate SIFT feature matching, updating it with new features is computationally intensive. RD SLAM suggests the usage of another tree, alternating between both, activating one tree at a time for matching, and another tree is passively waiting to be updated when a sufficient number of new SIFT features is observed. RD SLAM disregards feature matches that exhibit a viewpoint angle difference larger than thirty degrees.  To cope with dynamic objects and slowly varying scenes, RD SLAM suggests a prior-based adaptive RANSAC scheme that samples, based on the outlier ratio of features in previous frames, the features in the current frame from which to estimate the camera pose. Furthermore, it performs landmark and keyframe culling, using
histograms of colors to detect and update changed image locations, while sparing temporarily occluded landmarks.\\
%To gain immunity against slowly varying scene structures and gradual lighting changes, RD SLAM
%employs an online 3D points and keyframes update and culling mechanism at each incoming frame
%to remove 3D landmarks from the map that might have changed using histograms of colors with a
%mechanism to spare the points that the system suspects are temporarily occluded.

\textbf{\cite{pirchheim_2013_ISMAR}} dealt with the problem of pure rotations in the camera motion by building local panorama maps, whenever the system explores a new scene with  pure rotational
motion.  The system extracts phonySIFT descriptors as described in \cite{wagner_2010_TOVCC} and establishes feature correspondences using an accelerated matching method through hierarchical
k-means.
%The system is initialized using a model based detector as described in
%\cite{wagner_2010_tovcc}.
When insufficient 3D landmarks are observed during pose estimation, the system transitions into
a rotation-only estimation mode and starts building a panorama map until the camera observes part
of the finite map.\\

Also in 2013, \textbf{\cite{pradeep_2013_ISMAR}} published MonoFusion: Real-time 3D Reconstruction of
Small Scenes with a Single Web Camera.
%%% which is capable of generating a dense reconstruction of small scenes. MonoFusion employs an Essential matrix decomposition as an initialization module, and a camera pose tracking and map generation method similar to PTAM. 
MonoFusion estimates a dense depth map for every tracked frame $I$ by selecting, from a pool of its surrounding keyframes, the keyframe $I'$ that shares with it the most number
of feature matches. MonoFusion then proceeds by assigning for every pixel a set of random depth
values, and chooses the depth measurement that yield the least reprojection error between the two
frames. %%%MonoFusion then assumes that pixels belong to relatively larger regions with the same depth and argues that its random depth assignments is likely to yield low re-projection error for certain pixels; therefore, it suggests to propagate the depth measurements for those pixels to their surrounding neighborhood in a two pass scheme: first, from the upper-left corner, then from the lower-right corner of the image. 
Post processing steps are then employed to remove depth measurements with high re-projection error and isolated regions of similar depths (as they are most likely to be outliers). Finally a volumetric volume of the reconstructed scene is obtained by fusing the depth maps of every frame into a signed distance field (SDF), as described in \cite{curless_1996_CGIT}.\\

In the work of \textbf{\cite{lim_2014_ICRA}}, the sought after objective is to handle tracking, mapping,
and loop closure, all using the same binary feature, through a hybrid map representation.  Whenever a loop is detected, the map is converted to its metric form,
where a local bundle adjustment take place before returning the map back to its topological
form.\\

\textbf{\cite{bourmaud_2015_CVPR}} published an offline monocular SLAM system, which employs a
divide-and-conquer strategy, by segmenting the map into submaps. 
%A unique descriptor for every submap is computed using Bags of Words discretization of all SURF
%descriptors in the corresponding submap.
A similarity transform is estimated between each submap and its ten nearest neighbors.  A global
similarity transform, relating every submap to a single global reference frame, is computed by a
pose graph optimization, where the reference frames are stored in  a graph of submaps. The
above procedure is susceptible to outliers in the loop detection module, and hence the need for
an efficient outlier handling mechanism. For such purpose, temporally
consecutive similarity measurements are always considered as inliers. The outlier rejection
module proceeds then by integrating the similarities over the shortest loop it can find, and
monitors the closure error to accept the loop. To cope with a very large number of submaps, a
loopy belief propagation algorithm cuts the main graph into subgraphs, before applying a non-linear
optimization. %\\

In 2016, \textbf{\cite{greene_2016_ICRA}} published Multi-Level Mapping: Real-time Dense Monocular SLAM,
which is built on top of LSD SLAM, and allows it to generate and track a dense map, in
real-time; in contrast to LSD SLAM's default semi-dense map. The key contribution that allowed
such real-time operation is the introduction of Quad-trees keyframe representations: a pyramid
level representation of the keyframe is generated, and stored in trees. The depth and variance
measurements, estimated in a similar fashion to LSD SLAM, at the higher pyramid levels of the keyframes, allow for a denser representation of low-texture regions, once projected back onto the full image
resolution. Holes, corresponding to pixels that failed to converge during their depth estimates
are then filled using its surrounding neighboring pixels at their appropriate pyramid levels
before being mapped back onto the full resolution. Spatial regularization is then employed to
remove outlier measurements, and smooth the noise in the estimated depth map. Finally, the
estimated dense map undergoes a pose graph optimization over $sim(3)$ that ensures its alignment
with surrounding keyframes, and allows its incorporation within the global map.\\

In 2016, \textbf{RKSLAM} (Robust keyframe based monocular SLAM for augmented reality) \citep{liu_2016_ISMAR}
was published with the goal to robustify monocular slam algorithms for the scenarios encountered
during augmented reality sessions, specifically when the camera undergoes fast motions performed by
untrained personnel. RKSLAM argues that for erratic motions, motion models as employed by other
systems leads to drastic failure, and suggests to employ three different Homographies to solve for
the SLAM task:
%------------------------------------------------------------------------------------------
\begin{enumerate}[topsep=0pt,itemsep=-1ex,partopsep=1ex,parsep=1ex] 
    \item a global Homography that relates the current frame to other keyframes in the map, 
    \item a set of local Homographies extracted between the current frame and another keyframe through RANSAC and finally 
    \item planar specific Homographies relating specific planar surfaces observed in a keyframe and the current frame.  
\end{enumerate} 
%------------------------------------------------------------------------------------------
The Homographies are used to estimate a set of 3D points possibly available in the current field of
view, and accordingly warp them taking into account the viewpoint changes from which they were
initially observed, and finally project them on the current image.
To allow the system to accommodate much needed features as the user performs fast maneuvers, the local mapping thread was
moved from the backend of the system to its frontend and operates simultaneously with the tracking thread, at frame-rate. The local mapping system is responsible for maintaining and generating weakly supported 3D landmarks that have not yet been fully optimized by the global mapping thread that runs in the background. 

%The weakly supported 3D landmarks are the result of a feature triangulation
%observed in a keyframe and its corresponding match with another frame (that is not a keyframe)
%within a window surrounding the current frame. RKSLAM argues that, for small baselines, the depth
%estimates may not be accurate; and hence, they replace the depth estimated for these 3D landmarks
%with the mean depth of the entire keyframe in which it was first observed in, followed by an
%optimization that minimizes the re-projection of the 3D landmark onto both frames. To counter
%failure from severe motion blur, RKSLAM falls back on IMU measurements to estimate the camera pose
%and attempt to maintain tracking. If IMU measurements are not available, RKSLAM simulates virtual
%measurements by assuming a null linear acceleration, and a rotational velocity, estimated from small
%blurry image alignment.  Failure recovery and keyframe selection employed in RKSLAM are similar to
%the ones suggested by \citep{klein_2007_ISMAR}. Finally, it suggests the same failure recovery
%routine as a loop closure detection mechanism.

Most of the aforementioned closed source systems are not as popular as their open source
counterparts; however, they influenced their development. They also provide a lot of constructive ideas that can be built upon to enhance
the open source systems. For example ideas put forward by RDSLAM can be adapted to robustify KSLAM
systems in dynamic and occluded environments. Other ideas such as the adoption of a photogenerative
model in monocular SLAM by \cite{silveira_2008_TRO} was recently employed in the open source work of \cite{engel_2016_ARXIV}. 
%In conclusion, we have highlighted and summarized the main contributions of these systems; however, to avoid repetition, we will limit our discussion to this level and refer the interested reader for the appropriate references within this section for further details.    
	
%*************************************  CONCLUSIONS   *********************************************
\section{Discussion}\label{sec:discussion}

During the preparation of this survey, we considered including a benchmark, in which the different
monocular SLAM systems in the literature are compared. However, each different flavor of monocular
SLAM is favored by different operational conditions. For example, SVO \citep{forster_2014_ICRA}
prefers high frame rate inputs from downward looking cameras, DPPTAM \citep{concha_2015_IROS} can
only operate in indoor environments where most of the observed scene is composed of planar surfaces.
DT SLAM \citep{herrera_2014_IC3DV} requires the scene to be repeatedly observed.
Accordingly, there does not exist any public dataset in the literature that would allow us to
perform an unbiased experimental comparison across all systems. Instead, in
this section, we discuss and evaluate the ramifications of the decisions made in each component of the different monocular SLAM systems, providing a theoretical insight into the limitations of the various modules designs.

\subsection{Traits of Data association}
\subsubsection{Traits of direct methods}
Direct methods exploit all information available in
the image and are therefore more robust than feature-based methods in regions with poor texture and blur.
Nevertheless, direct methods are susceptible to scene illumination changes, due to the violation of the underlying brightness consistency assumption (eq. \eqref{BrightnessConstraint}) . In an effort to gain resilience
against this mode of failure, the recently released DSO models the image formation process, and
attempts to incorporate the scene irradiance in the energy functional, at the expense of
adding a calibrated image formation model which is used to correct the images at a pre-processing step. The
model is estimated through an additional offline calibration process described in
\citep{engel2_2016_arxiv}.
	
Furthermore, during the non-linear optimization process, eq.\eqref{FAIA}, %and its equivalent in other image alignment variants, 
is linearized through a first order Taylor expansion. %on $I(W(x,y,p))$; 
While the linearizion is valid when the parameters of the warping transform tends to zero, higher order terms becomes dominant and the linearizion becomes invalid for large transforms.  Therefore, a second disadvantage of direct methods is the assumption of small motions between the images
(typically not more than 1 pixel). To relax this constraint, direct monocular SLAM systems employ a
pyramidal implementation, where the image alignment process takes place sequentially from the
highest pyramid level to the lowest, using the results of every level as a prior to the next level.
They also suggest the usage of high fame rate cameras to alleviate this issue; some systems employ
an efficient second order minimization (ESM \cite{Benhimane_2007_IJRR}) to estimate a rotation prior
that helps increase the convergence radius.  Despite these efforts, the tolerated baseline for data
association in direct methods is considerably smaller than the tolerated baseline in feature-based
methods.
	
Another disadvantage of direct methods is that the calculation of the photometric error at every
pixel is computationally intensive; therefore, real-time monocular SLAM applications of direct methods,
until recently, were not considered feasible. However, with the recent advancements in parallelized
processing and with the introduction of semi-dense inverse depth filtering, it became possible to
integrate direct methods into KSLAM solutions
\citep{forster_2014_ICRA,engel_2015_IROS,concha_2015_IROS}.

\subsubsection{Traits of feature-based methods}    
Feature-based methods are relatively robust to lighting changes and can tolerate wider baselines;
however, the extraction processes that make them resilient to these factors are generally
computationally expensive. For real-time operation constraints, most systems employ a trade-of
between a feature type to use in one hand, and
the robustness and resilience to environment factors on the other. To mitigate this constraint,
other systems, such as the work of \cite{tan_2013_ISMAR}, resort to parallelized GPU implementations
for feature detection and extraction. 
    
\begin{figure}[!hbt] %htb
	\centering
	\includegraphics[width=0.5\textwidth]{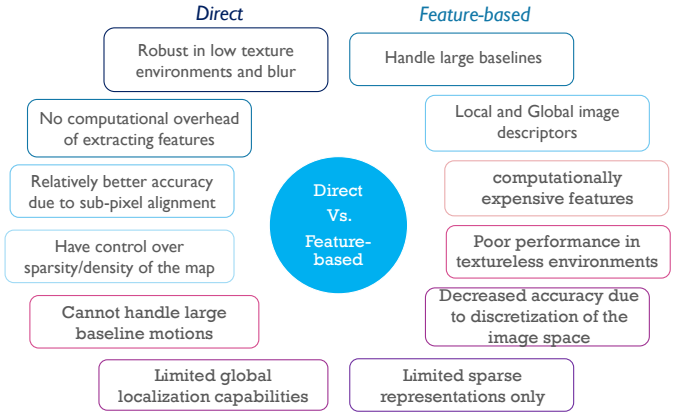}
	\caption{ Comparison between the different traits of direct vs. feature-based KSLAM systems.}
	\label{fig:FeatvsDir}
\end{figure}

Another disadvantage of feature-based methods is that even the top performing feature descriptors are
limited in the amount of scene change (lighting and viewpoint) they can handle before failure.  Feature
matching is also prone to failure in similar-self-repeating texture environments, where a feature in
$I_1$ can be ambiguously matched to multiple other features in $I_2$.  Outliers in the data
association module can heavily degrade the system performance by inducing errors in both the camera
poses and the generated map until the point of failure. Feature-based methods also suffer from lack of features in textureless regions, causing feature-based KSLAM to fail in texture-deprived environments. A summary of the comparison between direct and feature-based methods is shown in Fig. \ref{fig:FeatvsDir}.
%%%%%%%%%%%%% data association 2%%%%%%%%%%%%%%%%%%%%%%%%%%%%%%%%%%
\subsubsection{Traits of the different data association types}
In general establishing data associations remains one of the biggest challenges in KSLAM.  Systems that limit
the search range along the epipolar line using the observed depth information, implicitly assume a
relatively smooth depth distribution. Violating this assumption (\ie when the scene includes
significance variance in the observed depth) causes the 2D features corresponding to potential
future 3D landmarks to fall outside the boundaries of the epipolar segment, and the system ends up
neglecting them.  Other limitations for data association arise from large erratic accelerations in
the camera's motion, also causing features to fall outside the scope of the search window. Such a
scenario is common when the camera is operated by an untrained user. Under the same type of motions,
image pollution with motion blur also negatively impacts the performance of data association methods
to the point of failure.

Erroneous data association is also a very common problem that can cause false positives in self
repeating environments.  Most current implementations of data association address this problem
through a bottom-up approach, where low level information from image pixels or from features, is
used to establish correspondences.  To mitigate some of these issues, a number of systems have
attempted to use geometric features of a higher significance, such as lines
\citep{bista_2016_RAL,zhang_2011_IMVIP,dube_2016_ARXIV,klein_2008_ECCV,micusik_2015_CVPR,vakhitov_2016_ECCV,verhagen_2014_WACV,
yammine_2014_CSVT},  superpixels or planar
features\citep{concha_2014_ICRA,concha_2015_IROS,martinez_2010_BMVA}, or priors on 3D shapes in the
scene \citep{galvez_2015_ARXIV}.  Recent advances in machine learning are promising alternatives to
remedy some of the data association issues by automatically learning to extract and match features
using the methods suggested and not limited to
\cite{serra_2015_ICCV,verdie_2015_CVPR,han_2015_CVPR,zagoruyko_2015_CVPR,yi_2016_CVPR}.
%yi_2016_arxiv balntas_2016_arxiv...(a ton of other papers) or by scene understanding and labeling
%\citep{yang_2016_iros,kundu_2014_eccv,fioraio_2013_cvpr}.

\subsection{Traits of initialization} 

Aside from the random depth initialization of LSD SLAM and DSO, all the suggested methods described
above suffer from degeneracies under certain conditions, such as under low-parallax
movements of the camera, or when the scene's structure assumption\textemdash Fundamental matrix's
assumption for general non-planar scenes or the Homography's assumption of planar scenes\textemdash is violated.
    
PTAM's initialization procedure is brittle and remains tricky to perform, especially for
inexperienced users. Furthermore, it is subject to degeneracies when the planarity of the initial
scene's assumption is violated, or when the user's motion is inappropriate; therby crashing the system,
without means of detecting such degeneracies. 

As is the case in PTAM, the initialization of SVO requires the same type of motion and is prone to
sudden movements, as well as to non-planar scenes. Furthermore, monitoring the median of the
baseline distance 
between features is not a good approach to automate the initial keyframe pair selection, as it is
prone to failure against degenerate cases, with no means of detecting them.
    
The model based initialization of ORB SLAM attempts to automatically initialize the system by
monitoring the baseline and the scene across a window of images. If the observed scene is relatively
far, while the camera slowly translates in the scene, the system is not capable of detecting such
scenarios, and fails to initialize.
    
While a random depth initialization from a single image does not suffer from the degeneracies of two
view geometry methods, the depth estimation requires the processing of subsequent frames to converge, resulting in an intermediate tracking phase where the generated map
is not reliable. It requires slow translational motions, and the convergence of the initialization is not guaranteed.

\subsection{Traits of pose estimation} 

Systems relying on constant motion
models, such as PTAM and ORB SLAM are prone to tracking failure when abrupt changes in the direction
of the camera's motion occurs. While they both employ a recovery from such failures, PTAM's
tracking performance is exposed to false positive
pose recovery; as opposed to ORB SLAM that first attempts to increase the search window before
invoking its failure recovery module.
    
Another limitation of feature-based pose estimation is the detection and handling of occlusions. As the
camera translates in the scene, some landmarks in the background are prone to occlusions from
objects in the foreground. When the system projects the 3D map points onto the current frame, it
fails to match the occluded features, and counts them toward the camera tracking quality assessment.
In extreme cases, the tracking quality of the system might be deemed bad and
tracking failure recovery procedures are invoked even though camera pose tracking did not fail.
Furthermore, occluded points are flagged as outliers and passed to the map maintenance module to be
removed, depriving the map from valid useful landmarks that were erroneously flagged due to
occlusions in the scene.   
    
Other systems, that use the previously tracked pose as a prior for the new frame's pose, are also prone to the same limitations of constant
velocity models. Furthermore,  they require small displacements between frames, limiting
their operation to relatively expensive high frame rate cameras (typically $>70 fps$) such that the
displacement limitation is not exceeded. Another limitation of these methods is inherent from their
use of direct data association. Their tracking module is susceptible to variations in the lighting
conditions. To gain some resilience to lighting changes in direct methods, \cite{engel_2016_ARXIV}
suggest an off-line photometric calibration process to parametrize and incorporate lighting
variations within the camera pose optimization process. 
    
A common limitation that plagues most tracking modules is the presence of dynamic objects in the
observed environment. As most KSLAM systems assume a static scene, the tracking modules of most
systems suffer from tracking failures: a significantly large dynamic object in the scene could trick
the system into thinking that the camera itself is moving, while it did not move relative to the
environment. Small, slowly moving objects can introduce noisy outlier landmarks in the map and
require subsequent processing and handling to be removed. Small, fast moving objects on the other
hand, don't affect the tracking module as much. 2D features corresponding to fast moving small
objects tend to violate the epipolar geometry of the pose estimation problem, and are easily flagged
and removed from the camera pose optimization thread; however, they can occlude other landmarks. To
address the effects of occlusions and dynamic objects in the scene, \cite{tan_2013_ISMAR} suggest,
for slowly varying scenes, to sample based on previous camera pose locations in the image that are
not reliable, and discard them during the tracking phase.

\subsection{Traits of map generation}

A major limitation in the triangulation-by-optimization method is the requirement of a significant baseline separating two viewpoints observing the same feature. Hence, it is prone to failure when the camera's motion is made of pure rotations. To counter such modes of failure, DT SLAM introduced 2D landmarks that can be used to expand the map
during pure rotations, before they are triangulated into 3D landmarks. However, the observed scene
during the rotation motion is expected to be re-observed with more baseline, for the landmarks to
transition from 2D to 3D. Unfortunately, in many applications this is not the case; for example, a
camera mounted on a car making a turn cannot re-observe the scene, and eventually tracking failure
occurs. DT SLAM addresses such cases by generating a new sub map and attempts to establish
connections to previously created sub-maps by invoking a thread to look for similar keyframes across
sub-maps, and establish data associations between them. In the mean time, it resumes tracking in
the new world coordinate frame of the new sub-map. This however renders the pose estimates obsolete;
at every tracking failure, the tracking is reset to the new coordinate frame, yielding useless pose
estimates until the sub-maps are joined together, which may never occur. 
     
In filter based triangulation methods, outliers are easily flagged as landmarks whose distribution
remain approximately uniform after a  number of observations have been incorporated in the
framework. This reduces the need for a subsequent processing step to detect and handle outliers.
Also, landmarks at infinity feature parallax values that are too small for triangulation purposes;
but yet, can be used to enhance
the camera's rotation estimates, and kept in the map, and are transitioned from
infinity to the metric map, when enough parallax between the views observing them is recorded.
However, these benefits come at the expense of increased complexity in implementing a probabilistic
framework, which keeps track and updates the uncertainty in the depth distribution of every pixel with
a gradient in the system. 
    
Furthermore, while the dense and semi-dense maps can capture a much more meaningful representation
of a scene than a sparse set of 3D landmarks, the added value is diminished by the challenges of
handling immense amounts of data in 3D. Hence there is a need for additional, higher level semantic
information to reason about the observed scene, and to enhance the system's overall performance.
While monocular SLAM systems have been shown to improve the results of semantic labeling
\citep{sudeep_2015_ARXIV}, the feedback from the latter to the former remains a challenging problem.
Previous work on the matter include but are not limited to \citep{kundu_2014_ECCV,galvez_2015_ARXIV,fioraio_2013_CVPR,savarese_2012_CVPR,yang_2016_IROS}.

\subsection{Traits of BA/PGO/map maintenance}

Pose Graph Optimization (PGO) returns inferior results to those produced by GBA, while PGO optimizes
only for the keyframe poses\textemdash and accordingly adjusts the 3D structure of
landmarks\textemdash GBA and LBA jointly optimize for both keyframe poses and 3D structure. The
stated advantage comes at the cost of computational time, with PGO exhibiting significant speed up
compared to the other methods.  PGO is often employed during loop closure as the computational cost
of running a full BA is often intractable on large-scale loops; however, pose graph optimization may
not yield optimal result if the errors accumulated over the loop are distributed along the entire
map, leading to locally induced inaccuracies in regions that were not originally wrong.

\subsection{Traits of global localization}

For successful re-localization or loop detection, global localization methods employed by PTAM, SVO and DT SLAM require
the camera's pose to be near the recorded keyframe's pose, and would otherwise fail when there
is a large displacement between the two. Furthermore,  they are highly sensitive to any change in the
lighting conditions of the scene, and may yield many false positives when the observed environment is
composed of self repeating textures.  Other methods that rely on bags of words representation of
high dimensional features are susceptible to failure when the
training set of the bag of words classifier is not representative of the working environment in
which the system is operating.

%*************************************  CONCLUSIONS   *********************************************
\section{Conclusions}
\label{sec:conclusion}
During the course of our review, we have outlined the essential building blocks of a generic monocular SLAM
system; including data association, visual initialization, pose estimation,
topological/metric map generation, BA/PGO/map maintenance, and global localization.  We have also discussed the details of the latest open-source state of the art systems
in monocular SLAM including PTAM, SVO, DT SLAM, LSD SLAM, ORB SLAM, DPPTAM, and DSO. Finally, we
compiled and summarized what added information closed-source keyframe-based monocular SLAM systems
have to offer. 
Although extensive research has been dedicated to this field, it is our opinion that each of the
building blocks discussed above could benefit from many improvements of which we list the following:
\begin{itemize}
	  \setlength\itemsep{0em}
\item robust data association against illumination changes, dynamic scenes,  and occluded environments, 
\item a robust initialization method that can operate without an initial scene assumption,
\item an accurate camera pose estimate that is not affected by sudden movements, blur, noise, large
    depth variations, nor moving objects, 
\item a map making module capable of generating an efficient dense scene representation in regions
    of little texture, while incorporating a higher level of perception,
\item a map maintenance method that improves the map, with resilience against dynamic, changing environments, and finally, 
\item a failure recovery procedure capable of reviving the system from significantly large changes in camera viewpoints.  
\end{itemize}
These are all desired properties that remain challenging topics in the field of keyframe-based monocular Visual SLAM.
Furthermore, with the recent advancements in machine learning, researchers are moving towards integrating semantic data within the context of VSLAM. While the incorporation of semantic data into VSLAM is undoubtedly the next step in the right direction, we argue that such integration requires a hybrid fusion approach that tightly integrates metric, topological and semantic representations in a symbiotic relationship, a research area relatively uncharted.

%*************************************  APPENDIX   ************************************************
%\include{Appendix}
%*************************************  ACKNOWLEDGEMENTS   ****************************************	
\section*{Acknowledgements} 
This work was funded by the ENPI (European Neighborhood Partnership Instrument) grant \#
I-A/1.2/113, the Lebanese National Research Council (LNCSR), and the Canadian National Science
Research Council (NSERC).    
%******************  REFERENCES   *****************************************************************
\section*{References}
\begin{singlespace}
{
	\footnotesize{\bibliography{younes.bib}}

\begin{thebibliography}{111}
\expandafter\ifx\csname natexlab\endcsname\relax\def\natexlab#1{#1}\fi
\providecommand{\url}[1]{\texttt{#1}}
\providecommand{\href}[2]{#2}
\providecommand{\path}[1]{#1}
\providecommand{\DOIprefix}{doi:}
\providecommand{\ArXivprefix}{arXiv:}
\providecommand{\URLprefix}{URL: }
\providecommand{\Pubmedprefix}{pmid:}
\providecommand{\doi}[1]{\href{http://dx.doi.org/#1}{\path{#1}}}
\providecommand{\Pubmed}[1]{\href{pmid:#1}{\path{#1}}}
\providecommand{\bibinfo}[2]{#2}
\ifx\xfnm\relax \def\xfnm[#1]{\unskip,\space#1}\fi
%Type = Article
\bibitem[{Cadena et~al.(2016)Cadena, Carlone, Carrillo, Latif, Scaramuzza,
  Neira, Reid, and Leonard}]{cadena_2016_TRO}
\bibinfo{author}{C.~Cadena}, \bibinfo{author}{L.~Carlone},
  \bibinfo{author}{H.~Carrillo}, \bibinfo{author}{Y.~Latif},
  \bibinfo{author}{D.~Scaramuzza}, \bibinfo{author}{J.~Neira},
  \bibinfo{author}{I.~Reid}, \bibinfo{author}{J.~Leonard},
\newblock \bibinfo{title}{Past, present, and future of simultaneous
  localization and mapping: Toward the robust-perception age},
\newblock \bibinfo{journal}{IEEE Transactions on Robotics} \bibinfo{volume}{32}
  (\bibinfo{year}{2016}) \bibinfo{pages}{1309--1332}.
%Type = Article
\bibitem[{Scaramuzza and Fraundorfer(2011)}]{scaramuzza_2011_ram}
\bibinfo{author}{D.~Scaramuzza}, \bibinfo{author}{F.~Fraundorfer},
\newblock \bibinfo{title}{Visual odometry [tutorial]},
\newblock \bibinfo{journal}{IEEE Robotics Automation Magazine}
  \bibinfo{volume}{18} (\bibinfo{year}{2011}) \bibinfo{pages}{80--92}.
%Type = Article
\bibitem[{Fuentes-Pacheco et~al.(2012)Fuentes-Pacheco, Ruiz-Ascencio, and
  Rend{\'o}n-Mancha}]{fuentes_2012_air}
\bibinfo{author}{J.~Fuentes-Pacheco}, \bibinfo{author}{J.~Ruiz-Ascencio},
  \bibinfo{author}{J.~M. Rend{\'o}n-Mancha},
\newblock \bibinfo{title}{Visual simultaneous localization and mapping: a
  survey},
\newblock \bibinfo{journal}{Artificial Intelligence Review}
  \bibinfo{volume}{43} (\bibinfo{year}{2012}) \bibinfo{pages}{55--81}.
%Type = Article
\bibitem[{Yousif et~al.(2015)Yousif, Bab-Hadiashar, and
  Hoseinnezhad}]{yousif_2015_iis}
\bibinfo{author}{K.~Yousif}, \bibinfo{author}{A.~Bab-Hadiashar},
  \bibinfo{author}{R.~Hoseinnezhad},
\newblock \bibinfo{title}{An overview to visual odometry and visual slam:
  Applications to mobile robotics},
\newblock \bibinfo{journal}{Intelligent Industrial Systems} \bibinfo{volume}{1}
  (\bibinfo{year}{2015}) \bibinfo{pages}{289--311}.
%Type = Article
\bibitem[{Saeedi et~al.(2016)Saeedi, Trentini, Seto, and Li}]{saeedi_2016_jfr}
\bibinfo{author}{S.~Saeedi}, \bibinfo{author}{M.~Trentini},
  \bibinfo{author}{M.~Seto}, \bibinfo{author}{H.~Li},
\newblock \bibinfo{title}{Multiple-robot simultaneous localization and mapping:
  A review},
\newblock \bibinfo{journal}{Journal of Field Robotics} \bibinfo{volume}{33}
  (\bibinfo{year}{2016}) \bibinfo{pages}{3--46}.
%Type = Article
\bibitem[{Bailey and Durrant-Whyte(2006)}]{bailey_2006_mra}
\bibinfo{author}{T.~Bailey}, \bibinfo{author}{H.~Durrant-Whyte},
\newblock \bibinfo{title}{Simultaneous localization and mapping (slam): part
  ii},
\newblock \bibinfo{journal}{IEEE Robotics Automation Magazine}
  \bibinfo{volume}{13} (\bibinfo{year}{2006}) \bibinfo{pages}{108--117}.
%Type = Inproceedings
\bibitem[{Lucas and Kanade(1981)}]{lucas_1981_IJCAI}
\bibinfo{author}{B.~D. Lucas}, \bibinfo{author}{T.~Kanade},
\newblock \bibinfo{title}{{An Iterative Image Registration Technique with an
  Application to Stereo Vision}},
\newblock in: \bibinfo{booktitle}{International Joint Conference on Artificial
  Intelligence - Volume 2}, IJCAI'81, \bibinfo{publisher}{Morgan Kaufmann
  Publishers Inc.}, \bibinfo{address}{San Francisco, CA, USA},
  \bibinfo{year}{1981}, pp. \bibinfo{pages}{674--679}.
%Type = Article
\bibitem[{Baker and Matthews(2004)}]{baker_2004_IJCV}
\bibinfo{author}{S.~Baker}, \bibinfo{author}{I.~Matthews},
\newblock \bibinfo{title}{{Lucas-Kanade 20 Years On: A Unifying Framework}},
\newblock \bibinfo{journal}{International Journal of Computer Vision}
  \bibinfo{volume}{56} (\bibinfo{year}{2004}) \bibinfo{pages}{221--255}.
%Type = Inbook
\bibitem[{Krig(2014)}]{krig_2014_Apress}
\bibinfo{author}{S.~Krig}, \bibinfo{title}{Interest Point Detector and Feature
  Descriptor Survey}, \bibinfo{publisher}{Apress}, \bibinfo{address}{Berkeley,
  CA}, \bibinfo{year}{2014}, pp. \bibinfo{pages}{217--282}. \URLprefix
  \url{http://dx.doi.org/10.1007/978-1-4302-5930-5_6}.
  \DOIprefix\doi{10.1007/978-1-4302-5930-5_6}.
%Type = Inproceedings
\bibitem[{Beaudet(1978)}]{beaudet_1978_ICPR}
\bibinfo{author}{P.~R. Beaudet},
\newblock \bibinfo{title}{{Rotationally invariant image operators}},
\newblock in: \bibinfo{booktitle}{International Conference on Pattern
  Recognition}, \bibinfo{year}{1978}.
%Type = Inproceedings
\bibitem[{Harris and Stephens(1988)}]{harris_1988_FAVC}
\bibinfo{author}{C.~Harris}, \bibinfo{author}{M.~Stephens},
\newblock \bibinfo{title}{A combined corner and edge detector},
\newblock in: \bibinfo{booktitle}{In Proc. of Fourth Alvey Vision Conference},
  \bibinfo{year}{1988}, pp. \bibinfo{pages}{147--151}.
%Type = Inproceedings
\bibitem[{Shi and Tomasi(1994)}]{shi_1994_CVPR}
\bibinfo{author}{J.~Shi}, \bibinfo{author}{C.~Tomasi},
\newblock \bibinfo{title}{{Good features to track}},
\newblock in: \bibinfo{booktitle}{Computer Vision and Pattern Recognition,
  1994. Proceedings CVPR '94., 1994 IEEE Computer Society Conference on},
  \bibinfo{year}{1994}, pp. \bibinfo{pages}{593--600}.
%Type = Article
\bibitem[{Lindeberg(1998)}]{lindeberg_1998_IJCV}
\bibinfo{author}{T.~Lindeberg},
\newblock \bibinfo{title}{Feature detection with automatic scale selection},
\newblock \bibinfo{journal}{Int. J. Comput. Vision} \bibinfo{volume}{30}
  (\bibinfo{year}{1998}) \bibinfo{pages}{79--116}.
%Type = Inproceedings
\bibitem[{Matas et~al.(2002)Matas, Chum, Urban, and Pajdla}]{matas_2002_bmvc}
\bibinfo{author}{J.~Matas}, \bibinfo{author}{O.~Chum},
  \bibinfo{author}{M.~Urban}, \bibinfo{author}{T.~Pajdla},
\newblock \bibinfo{title}{Robust wide baseline stereo from maximally stable
  extremal regions},
\newblock in: \bibinfo{booktitle}{Proc. BMVC}, \bibinfo{year}{2002}, pp.
  \bibinfo{pages}{36.1--36.10}. \bibinfo{note}{Doi:10.5244/C.16.36}.
%Type = Article
\bibitem[{Lowe(2004)}]{lowe_2004_IJCV}
\bibinfo{author}{D.~G. Lowe},
\newblock \bibinfo{title}{Distinctive image features from scale-invariant
  keypoints},
\newblock \bibinfo{journal}{Int. J. Comput. Vision} \bibinfo{volume}{60}
  (\bibinfo{year}{2004}) \bibinfo{pages}{91--110}.
%Type = Inproceedings
\bibitem[{Mair et~al.(2010)Mair, Hager, Burschka, Suppa, and
  Hirzinger}]{mair_2010_ECCV}
\bibinfo{author}{E.~Mair}, \bibinfo{author}{G.~D. Hager},
  \bibinfo{author}{D.~Burschka}, \bibinfo{author}{M.~Suppa},
  \bibinfo{author}{G.~Hirzinger},
\newblock \bibinfo{title}{{Adaptive and Generic Corner Detection Based on the
  Accelerated Segment Test}},
\newblock in: \bibinfo{booktitle}{Proceedings of the European Conference on
  Computer Vision (ECCV'10)}, \bibinfo{year}{2010}.
  \DOIprefix\doi{10.1007/978-3-642-15552-9{\_}14}.
%Type = Article
\bibitem[{Calonder et~al.(2012)Calonder, Lepetit, Ozuysal, Trzcinski, Strecha,
  and Fua}]{calonder_2012_PAMI}
\bibinfo{author}{M.~Calonder}, \bibinfo{author}{V.~Lepetit},
  \bibinfo{author}{M.~Ozuysal}, \bibinfo{author}{T.~Trzcinski},
  \bibinfo{author}{C.~Strecha}, \bibinfo{author}{P.~Fua},
\newblock \bibinfo{title}{Brief: Computing a local binary descriptor very
  fast},
\newblock \bibinfo{journal}{IEEE Transactions on Pattern Analysis and Machine
  Intelligence} \bibinfo{volume}{34} (\bibinfo{year}{2012})
  \bibinfo{pages}{1281--1298}.
%Type = Inproceedings
\bibitem[{Leutenegger et~al.(2011)Leutenegger, Chli, and
  Siegwart}]{leutenegger_2011_ICCV}
\bibinfo{author}{S.~Leutenegger}, \bibinfo{author}{M.~Chli},
  \bibinfo{author}{R.~Y. Siegwart},
\newblock \bibinfo{title}{Brisk: Binary robust invariant scalable keypoints},
\newblock in: \bibinfo{booktitle}{Computer Vision (ICCV), 2011 IEEE
  International Conference on}, \bibinfo{year}{2011}, pp.
  \bibinfo{pages}{2548--2555}. \DOIprefix\doi{10.1109/ICCV.2011.6126542}.
%Type = Article
\bibitem[{Bay et~al.(2008)Bay, Ess, Tuytelaars, and Van~Gool}]{bay_2008_CVIU}
\bibinfo{author}{H.~Bay}, \bibinfo{author}{A.~Ess},
  \bibinfo{author}{T.~Tuytelaars}, \bibinfo{author}{L.~Van~Gool},
\newblock \bibinfo{title}{Speeded-up robust features (surf)},
\newblock \bibinfo{journal}{Comput. Vis. Image Underst.} \bibinfo{volume}{110}
  (\bibinfo{year}{2008}) \bibinfo{pages}{346--359}.
%Type = Inproceedings
\bibitem[{Lowe(1999)}]{lowe_1999_ICCV}
\bibinfo{author}{D.~Lowe},
\newblock \bibinfo{title}{{Object recognition from local scale-invariant
  features}},
\newblock in: \bibinfo{booktitle}{International Conference on Computer Vision
  (ICCV), the seventh IEEE}, volume~\bibinfo{volume}{2},
  \bibinfo{publisher}{IEEE}, \bibinfo{year}{1999}, pp.
  \bibinfo{pages}{1150--1157 vol.2}.
%Type = Inproceedings
\bibitem[{Dalal and Triggs(2005)}]{dalal_2005_CVPR}
\bibinfo{author}{N.~Dalal}, \bibinfo{author}{B.~Triggs},
\newblock \bibinfo{title}{Histograms of oriented gradients for human
  detection},
\newblock in: \bibinfo{booktitle}{Computer Vision and Pattern Recognition,
  2005. CVPR 2005. IEEE Computer Society Conference on},
  volume~\bibinfo{volume}{1}, \bibinfo{year}{2005}, pp.
  \bibinfo{pages}{886--893 vol. 1}. \DOIprefix\doi{10.1109/CVPR.2005.177}.
%Type = Inproceedings
\bibitem[{Alahi et~al.(2012)Alahi, Ortiz, and Vandergheynst}]{alahi_2012_CVPR}
\bibinfo{author}{A.~Alahi}, \bibinfo{author}{R.~Ortiz},
  \bibinfo{author}{P.~Vandergheynst},
\newblock \bibinfo{title}{Freak: Fast retina keypoint},
\newblock in: \bibinfo{booktitle}{Computer Vision and Pattern Recognition
  (CVPR), 2012 IEEE Conference on}, \bibinfo{year}{2012}, pp.
  \bibinfo{pages}{510--517}. \DOIprefix\doi{10.1109/CVPR.2012.6247715}.
%Type = Inproceedings
\bibitem[{Rublee et~al.(2011)Rublee, Rabaud, Konolige, and
  Bradski}]{rublee_2011_ICCV}
\bibinfo{author}{E.~Rublee}, \bibinfo{author}{V.~Rabaud},
  \bibinfo{author}{K.~Konolige}, \bibinfo{author}{G.~Bradski},
\newblock \bibinfo{title}{{ORB: An efficient alternative to SIFT or SURF}},
\newblock in: \bibinfo{booktitle}{International Conference on Computer Vision
  (ICCV)}, \bibinfo{year}{2011}, pp. \bibinfo{pages}{2564--2571}.
%Type = Article
\bibitem[{Moreels and Perona(2007)}]{moreels_2007_IJCV}
\bibinfo{author}{P.~Moreels}, \bibinfo{author}{P.~Perona},
\newblock \bibinfo{title}{Evaluation of features detectors and descriptors
  based on 3d objects},
\newblock \bibinfo{journal}{Int. J. Comput. Vision} \bibinfo{volume}{73}
  (\bibinfo{year}{2007}) \bibinfo{pages}{263--284}.
%Type = Inproceedings
\bibitem[{Hartmann et~al.(2013)Hartmann, Klussendorff, and
  Maehle}]{hartmann_2013_ECMR}
\bibinfo{author}{J.~Hartmann}, \bibinfo{author}{J.~H. Klussendorff},
  \bibinfo{author}{E.~Maehle},
\newblock \bibinfo{title}{{A comparison of feature descriptors for visual
  SLAM}},
\newblock in: \bibinfo{booktitle}{Mobile Robots (ECMR), 2013 European
  Conference on}, \bibinfo{year}{2013}, pp. \bibinfo{pages}{56--61}.
  \DOIprefix\doi{10.1109/ECMR.2013.6698820}.
%Type = Article
\bibitem[{Rey{-}Otero et~al.(2014)Rey{-}Otero, Delbracio, and
  Morel}]{rey_2014_CORR}
\bibinfo{author}{I.~Rey{-}Otero}, \bibinfo{author}{M.~Delbracio},
  \bibinfo{author}{J.~Morel},
\newblock \bibinfo{title}{Comparing feature detectors: {A} bias in the
  repeatability criteria, and how to correct it},
\newblock \bibinfo{journal}{CoRR} \bibinfo{volume}{abs/1409.2465}
  (\bibinfo{year}{2014}).
%Type = Article
\bibitem[{Hietanen et~al.(2016)Hietanen, Lankinen, K{\"a}m{\"a}r{\"a}inen,
  Buch, and Kr{\"u}ger}]{hietanen_2016_NC}
\bibinfo{author}{A.~Hietanen}, \bibinfo{author}{J.~Lankinen},
  \bibinfo{author}{J.-K. K{\"a}m{\"a}r{\"a}inen}, \bibinfo{author}{A.~G. Buch},
  \bibinfo{author}{N.~Kr{\"u}ger},
\newblock \bibinfo{title}{A comparison of feature detectors and descriptors for
  object class matching},
\newblock \bibinfo{journal}{Neurocomputing}  (\bibinfo{year}{2016}).
%Type = Inproceedings
\bibitem[{J{\'e}r{\^o}me~Martin(1995)}]{martin_1995_ICIAS}
\bibinfo{author}{J.~L.~C. J{\'e}r{\^o}me~Martin},
\newblock \bibinfo{title}{{Experimental Comparison of Correlation Techniques}},
\newblock in: \bibinfo{booktitle}{IAS-4, International Conference on
  Intelligent Autonomous Systems}, \bibinfo{year}{1995}.
%Type = Inproceedings
\bibitem[{Muja and Lowe(2009)}]{muja_2009_CVTA}
\bibinfo{author}{M.~Muja}, \bibinfo{author}{D.~G. Lowe},
\newblock \bibinfo{title}{Fast approximate nearest neighbors with automatic
  algorithm configuration},
\newblock in: \bibinfo{booktitle}{In VISAPP International Conference on
  Computer Vision Theory and Applications}, \bibinfo{year}{2009}, pp.
  \bibinfo{pages}{331--340}.
%Type = Article
\bibitem[{Galvez-L{\'o}pez and Tardos(2012)}]{lopez_2012_TRO}
\bibinfo{author}{D.~Galvez-L{\'o}pez}, \bibinfo{author}{J.~D. Tardos},
\newblock \bibinfo{title}{{Bags of Binary Words for Fast Place Recognition in
  Image Sequences}},
\newblock \bibinfo{journal}{Robotics, IEEE Transactions on}
  \bibinfo{volume}{28} (\bibinfo{year}{2012}) \bibinfo{pages}{1188--1197}.
%Type = Article
\bibitem[{Umeyama(1991)}]{umeyama_1991_PAMI}
\bibinfo{author}{S.~Umeyama},
\newblock \bibinfo{title}{Least-squares estimation of transformation parameters
  between two point patterns},
\newblock \bibinfo{journal}{IEEE Trans. Pattern Anal. Mach. Intell.}
  \bibinfo{volume}{13} (\bibinfo{year}{1991}) \bibinfo{pages}{376--380}.
%Type = Article
\bibitem[{Davison et~al.(2007)Davison, Reid, Molton, and
  Stasse}]{davison_2007_PAMI}
\bibinfo{author}{A.~J. Davison}, \bibinfo{author}{I.~D. Reid},
  \bibinfo{author}{N.~D. Molton}, \bibinfo{author}{O.~Stasse},
\newblock \bibinfo{title}{{MonoSLAM: real-time single camera SLAM}},
\newblock \bibinfo{journal}{Pattern Analysis and Machine Intelligence (PAMI),
  IEEE Transactions on} \bibinfo{volume}{29} (\bibinfo{year}{2007})
  \bibinfo{pages}{1052--67}.
%Type = Article
\bibitem[{Longuet-Higgins(1981)}]{higgins_1981_LN}
\bibinfo{author}{H.~C. Longuet-Higgins},
\newblock \bibinfo{title}{A computer algorithm for reconstructing a scene from
  two projections},
\newblock \bibinfo{journal}{Letters to Nature} \bibinfo{volume}{293}
  (\bibinfo{year}{1981}) \bibinfo{pages}{133--135}.
%Type = Article
\bibitem[{Torr and Zisserman(2000)}]{torr_2000_CVIU}
\bibinfo{author}{P.~H.~S. Torr}, \bibinfo{author}{A.~Zisserman},
\newblock \bibinfo{title}{{MLESAC}},
\newblock \bibinfo{journal}{Computer Vision and Image Understanding}
  \bibinfo{volume}{78} (\bibinfo{year}{2000}) \bibinfo{pages}{138--156}.
%Type = Book
\bibitem[{Hartley and Zisserman(2003)}]{hartley_2003_Cambridge}
\bibinfo{author}{R.~Hartley}, \bibinfo{author}{A.~Zisserman},
  \bibinfo{title}{{Multiple View Geometry in Computer Vision}},
  \bibinfo{publisher}{Cambridge University Press}, \bibinfo{year}{2003}.
%Type = Article
\bibitem[{Boal et~al.(2014)Boal, S{\'a}nchez-Miralles, and
  Arranz}]{boal_2014_ROBOTICA}
\bibinfo{author}{J.~Boal}, \bibinfo{author}{{\'A}.~S{\'a}nchez-Miralles},
  \bibinfo{author}{{\'A}.~Arranz},
\newblock \bibinfo{title}{Topological simultaneous localization and mapping: a
  survey},
\newblock \bibinfo{journal}{Robotica} \bibinfo{volume}{32}
  (\bibinfo{year}{2014}) \bibinfo{pages}{803--821}.
%Type = Article
\bibitem[{Mur-Artal et~al.(2015)Mur-Artal, Montiel, and
  Tardos}]{mur-artal_2015_TRO}
\bibinfo{author}{R.~Mur-Artal}, \bibinfo{author}{J.~M.~M. Montiel},
  \bibinfo{author}{J.~D. Tardos},
\newblock \bibinfo{title}{{ORB-SLAM: A Versatile and Accurate Monocular SLAM
  System}},
\newblock \bibinfo{journal}{IEEE Transactions on Robotics} \bibinfo{volume}{PP}
  (\bibinfo{year}{2015}) \bibinfo{pages}{1--17}.
%Type = Incollection
\bibitem[{Engel et~al.(2014)Engel, Sch{\"{o}}ps, and Cremers}]{engel_2014_ECCV}
\bibinfo{author}{J.~Engel}, \bibinfo{author}{T.~Sch{\"{o}}ps},
  \bibinfo{author}{D.~Cremers},
\newblock \bibinfo{title}{{LSD-SLAM: Large-Scale Direct Monocular SLAM}},
\newblock in: \bibinfo{editor}{D.~Fleet}, \bibinfo{editor}{T.~Pajdla},
  \bibinfo{editor}{B.~Schiele}, \bibinfo{editor}{T.~Tuytelaars} (Eds.),
  \bibinfo{booktitle}{Computer Vision -- ECCV 2014 SE - 54}, volume
  \bibinfo{volume}{8690} of \textit{\bibinfo{series}{Lecture Notes in Computer
  Science}}, \bibinfo{publisher}{Springer International Publishing},
  \bibinfo{year}{2014}, pp. \bibinfo{pages}{834--849}.
%Type = Inproceedings
\bibitem[{Lim et~al.(2011)Lim, Frahm, and Pollefeys}]{lim_2011_CVPR}
\bibinfo{author}{J.~Lim}, \bibinfo{author}{J.~M. Frahm},
  \bibinfo{author}{M.~Pollefeys},
\newblock \bibinfo{title}{Online environment mapping},
\newblock in: \bibinfo{booktitle}{Computer Vision and Pattern Recognition
  (CVPR), 2011 IEEE Conference on}, \bibinfo{year}{2011}, pp.
  \bibinfo{pages}{3489--3496}. \DOIprefix\doi{10.1109/CVPR.2011.5995511}.
%Type = Inproceedings
\bibitem[{Lim et~al.(2014)Lim, Lim, and Kim}]{lim_2014_ICRA}
\bibinfo{author}{H.~Lim}, \bibinfo{author}{J.~Lim}, \bibinfo{author}{H.~J.
  Kim},
\newblock \bibinfo{title}{{Real-time 6-DOF monocular visual SLAM in a
  large-scale environment}},
\newblock in: \bibinfo{booktitle}{Robotics and Automation (ICRA), IEEE
  International Conference on}, \bibinfo{year}{2014}, pp.
  \bibinfo{pages}{1532--1539}.
%Type = Incollection
\bibitem[{Fern{\'{a}}ndez-Moral et~al.(2015)Fern{\'{a}}ndez-Moral,
  Ar{\'{e}}valo, and Gonz{\'{a}}lez-Jim{\'{e}}nez}]{fernandez_2015_Springer}
\bibinfo{author}{E.~Fern{\'{a}}ndez-Moral}, \bibinfo{author}{V.~Ar{\'{e}}valo},
  \bibinfo{author}{J.~Gonz{\'{a}}lez-Jim{\'{e}}nez},
\newblock \bibinfo{title}{{Hybrid Metric-topological Mapping for Large Scale
  Monocular SLAM }},
\newblock \bibinfo{publisher}{Springer International Publishing},
  \bibinfo{year}{2015}, pp. \bibinfo{pages}{217--232}.
%Type = Inproceedings
\bibitem[{Konolige(2010)}]{konolige_2010_BMVC}
\bibinfo{author}{K.~Konolige},
\newblock \bibinfo{title}{Sparse sparse bundle adjustment},
\newblock in: \bibinfo{booktitle}{Proceedings of the British Machine Vision
  Conference}, \bibinfo{publisher}{BMVA Press}, \bibinfo{year}{2010}, pp.
  \bibinfo{pages}{102.1--102.11}. \bibinfo{note}{Doi:10.5244/C.24.102}.
%Type = Article
\bibitem[{Hartley and Sturm(1997)}]{hartley_1997_CVIU}
\bibinfo{author}{R.~I. Hartley}, \bibinfo{author}{P.~Sturm},
\newblock \bibinfo{title}{Triangulation},
\newblock \bibinfo{journal}{Comput. Vis. Image Underst.} \bibinfo{volume}{68}
  (\bibinfo{year}{1997}) \bibinfo{pages}{146--157}.
%Type = Inproceedings
\bibitem[{Hochdorfer and Schlegel(2009)}]{hochdorfer_2009_ICAR}
\bibinfo{author}{S.~Hochdorfer}, \bibinfo{author}{C.~Schlegel},
\newblock \bibinfo{title}{Towards a robust visual slam approach: Addressing the
  challenge of life-long operation},
\newblock in: \bibinfo{booktitle}{Advanced Robotics, 2009. ICAR 2009.
  International Conference on}, \bibinfo{year}{2009}, pp.
  \bibinfo{pages}{1--6}.
%Type = Article
\bibitem[{Civera et~al.(2008)Civera, Davison, and Montiel}]{civera_2008_TRO}
\bibinfo{author}{J.~Civera}, \bibinfo{author}{A.~Davison},
  \bibinfo{author}{J.~Montiel},
\newblock \bibinfo{title}{{Inverse Depth Parametrization for Monocular SLAM}},
\newblock \bibinfo{journal}{IEEE Transactions on Robotics} \bibinfo{volume}{24}
  (\bibinfo{year}{2008}) \bibinfo{pages}{932--945}.
%Type = Inproceedings
\bibitem[{Kummerle et~al.(2011)Kummerle, Grisetti, Strasdat, Konolige, and
  Burgard}]{kummerle_2011_ICRA}
\bibinfo{author}{R.~Kummerle}, \bibinfo{author}{G.~Grisetti},
  \bibinfo{author}{H.~Strasdat}, \bibinfo{author}{K.~Konolige},
  \bibinfo{author}{W.~Burgard},
\newblock \bibinfo{title}{{G2o: A general framework for graph optimization}},
\newblock in: \bibinfo{booktitle}{Robotics and Automation (ICRA), IEEE
  International Conference on}, \bibinfo{publisher}{IEEE},
  \bibinfo{year}{2011}, pp. \bibinfo{pages}{3607--3613}.
%Type = Inbook
\bibitem[{Triggs et~al.(2000)Triggs, McLauchlan, Hartley, and
  Fitzgibbon}]{triggs_2000_va}
\bibinfo{author}{B.~Triggs}, \bibinfo{author}{P.~F. McLauchlan},
  \bibinfo{author}{R.~I. Hartley}, \bibinfo{author}{A.~W. Fitzgibbon},
  \bibinfo{title}{Bundle Adjustment --- A Modern Synthesis},
  \bibinfo{publisher}{Springer Berlin Heidelberg}, \bibinfo{address}{Berlin,
  Heidelberg}, \bibinfo{year}{2000}, pp. \bibinfo{pages}{298--372}. \URLprefix
  \url{http://dx.doi.org/10.1007/3-540-44480-7_21}.
  \DOIprefix\doi{10.1007/3-540-44480-7_21}.
%Type = Inproceedings
\bibitem[{Strasdat et~al.(2011)Strasdat, Davison, Montiel, and
  Konolige}]{strasdat_2011_ICCV}
\bibinfo{author}{H.~Strasdat}, \bibinfo{author}{A.~J. Davison},
  \bibinfo{author}{J.~M.~M. Montiel}, \bibinfo{author}{K.~Konolige},
\newblock \bibinfo{title}{{Double Window Optimisation for Constant Time Visual
  SLAM}},
\newblock in: \bibinfo{booktitle}{International Conference on Computer Vision,
  Proceedings of the}, ICCV '11, \bibinfo{publisher}{IEEE Computer Society},
  \bibinfo{address}{Washington, DC, USA}, \bibinfo{year}{2011}, pp.
  \bibinfo{pages}{2352--2359}.
%Type = Article
\bibitem[{Garcia-Fidalgo and Ortiz(2015)}]{garcia_2015_ras}
\bibinfo{author}{E.~Garcia-Fidalgo}, \bibinfo{author}{A.~Ortiz},
\newblock \bibinfo{title}{Vision-based topological mapping and localization
  methods: A survey},
\newblock \bibinfo{journal}{Robotics and Autonomous Systems}
  \bibinfo{volume}{64} (\bibinfo{year}{2015}) \bibinfo{pages}{1 -- 20}.
%Type = Misc
\bibitem[{Agarwal et~al.(2013)Agarwal, Mierle et~al.}]{agarwal_2013_ceres}
\bibinfo{author}{S.~Agarwal}, \bibinfo{author}{K.~Mierle}, et~al.,
  \bibinfo{title}{Ceres solver}, \bibinfo{year}{2013}.
%Type = Inproceedings
\bibitem[{Mouragnon et~al.(2006)Mouragnon, Lhuillier, Dhome, Dekeyser, and
  Sayd}]{mouragnon_2006_CVPR}
\bibinfo{author}{E.~Mouragnon}, \bibinfo{author}{M.~Lhuillier},
  \bibinfo{author}{M.~Dhome}, \bibinfo{author}{F.~Dekeyser},
  \bibinfo{author}{P.~Sayd},
\newblock \bibinfo{title}{Real time localization and 3d reconstruction},
\newblock in: \bibinfo{booktitle}{Computer Vision and Pattern Recognition, 2006
  IEEE Computer Society Conference on}, volume~\bibinfo{volume}{1},
  \bibinfo{year}{2006}, pp. \bibinfo{pages}{363--370}.
  \DOIprefix\doi{10.1109/CVPR.2006.236}.
%Type = Article
\bibitem[{Klein and Murray(2007)}]{klein_2007_ISMAR}
\bibinfo{author}{G.~Klein}, \bibinfo{author}{D.~Murray},
\newblock \bibinfo{title}{{Parallel Tracking and Mapping for Small AR
  Workspaces}},
\newblock \bibinfo{journal}{6th IEEE and ACM International Symposium on Mixed
  and Augmented Reality}  (\bibinfo{year}{2007}) \bibinfo{pages}{1--10}.
%Type = Article
\bibitem[{Silveira et~al.(2008)Silveira, Malis, and Rives}]{silveira_2008_TRO}
\bibinfo{author}{G.~Silveira}, \bibinfo{author}{E.~Malis},
  \bibinfo{author}{P.~Rives},
\newblock \bibinfo{title}{An efficient direct approach to visual slam},
\newblock \bibinfo{journal}{Robotics, IEEE Transactions on}
  \bibinfo{volume}{24} (\bibinfo{year}{2008}) \bibinfo{pages}{969--979}.
%Type = Inproceedings
\bibitem[{Strasdat et~al.(2010)Strasdat, Montiel, and
  Davison}]{strasdat_2010_MIT}
\bibinfo{author}{H.~Strasdat}, \bibinfo{author}{J.~Montiel},
  \bibinfo{author}{A.~Davison},
\newblock \bibinfo{title}{Scale drift-aware large scale monocular slam.},
\newblock \bibinfo{publisher}{The MIT Press}, \bibinfo{year}{2010}. \URLprefix
  \url{http://www.roboticsproceedings.org/rss06/}.
%Type = Inproceedings
\bibitem[{Newcombe and Davison(2010)}]{newcombe_2010_CVPR}
\bibinfo{author}{R.~A. Newcombe}, \bibinfo{author}{A.~J. Davison},
\newblock \bibinfo{title}{Live dense reconstruction with a single moving
  camera},
\newblock in: \bibinfo{booktitle}{Computer Vision and Pattern Recognition
  (CVPR), 2010 IEEE Conference on}, \bibinfo{organization}{IEEE},
  \bibinfo{year}{2010}, pp. \bibinfo{pages}{1498--1505}.
%Type = Inproceedings
\bibitem[{Newcombe et~al.(2011)Newcombe, Lovegrove, and
  Davison}]{newcombe_2011_ICCV}
\bibinfo{author}{R.~A. Newcombe}, \bibinfo{author}{S.~J. Lovegrove},
  \bibinfo{author}{A.~J. Davison},
\newblock \bibinfo{title}{Dtam: Dense tracking and mapping in real-time},
\newblock in: \bibinfo{booktitle}{Proceedings of the 2011 International
  Conference on Computer Vision}, ICCV '11, \bibinfo{publisher}{IEEE Computer
  Society}, \bibinfo{address}{Washington, DC, USA}, \bibinfo{year}{2011}, pp.
  \bibinfo{pages}{2320--2327}. \URLprefix
  \url{http://dx.doi.org/10.1109/ICCV.2011.6126513}.
  \DOIprefix\doi{10.1109/ICCV.2011.6126513}.
%Type = Inproceedings
\bibitem[{Pretto et~al.(2011)Pretto, Menegatti, and Pagello}]{pretto_2011_ICRA}
\bibinfo{author}{A.~Pretto}, \bibinfo{author}{E.~Menegatti},
  \bibinfo{author}{E.~Pagello},
\newblock \bibinfo{title}{Omnidirectional dense large-scale mapping and
  navigation based on meaningful triangulation},
\newblock in: \bibinfo{booktitle}{Robotics and Automation (ICRA), 2011 IEEE
  International Conference on}, \bibinfo{year}{2011}, pp.
  \bibinfo{pages}{3289--3296}. \DOIprefix\doi{10.1109/ICRA.2011.5980206}.
%Type = Inproceedings
\bibitem[{Pirker et~al.(2011)Pirker, R{\"{u}}ther, and
  Bischof}]{pirker_2011_IROS}
\bibinfo{author}{K.~Pirker}, \bibinfo{author}{M.~R{\"{u}}ther},
  \bibinfo{author}{H.~Bischof},
\newblock \bibinfo{title}{{CD SLAM - Continuous localization and mapping in a
  dynamic world.}},
\newblock in: \bibinfo{booktitle}{IEEE International Conference on Intelligent
  Robots Systems (IROS)}, \bibinfo{publisher}{IEEE}, \bibinfo{year}{2011}, pp.
  \bibinfo{pages}{3990--3997}.
%Type = Inproceedings
\bibitem[{Pirchheim and Reitmayr(2011)}]{pirchheim_2011_ISMAR}
\bibinfo{author}{C.~Pirchheim}, \bibinfo{author}{G.~Reitmayr},
\newblock \bibinfo{title}{Homography-based planar mapping and tracking for
  mobile phones},
\newblock in: \bibinfo{booktitle}{Mixed and Augmented Reality (ISMAR), 2011
  10th IEEE International Symposium on}, \bibinfo{year}{2011}, pp.
  \bibinfo{pages}{27--36}. \DOIprefix\doi{10.1109/ISMAR.2011.6092367}.
%Type = Article
\bibitem[{Tan et~al.(2013)Tan, Liu, Dong, Zhang, and Bao}]{tan_2013_ISMAR}
\bibinfo{author}{W.~Tan}, \bibinfo{author}{H.~Liu}, \bibinfo{author}{Z.~Dong},
  \bibinfo{author}{G.~Zhang}, \bibinfo{author}{H.~Bao},
\newblock \bibinfo{title}{{Robust monocular SLAM in dynamic environments}},
\newblock \bibinfo{journal}{2013 IEEE International Symposium on Mixed and
  Augmented Reality (ISMAR)}  (\bibinfo{year}{2013}) \bibinfo{pages}{209--218}.
%Type = Inproceedings
\bibitem[{Pirchheim et~al.(2013)Pirchheim, Schmalstieg, and
  Reitmayr}]{pirchheim_2013_ISMAR}
\bibinfo{author}{C.~Pirchheim}, \bibinfo{author}{D.~Schmalstieg},
  \bibinfo{author}{G.~Reitmayr},
\newblock \bibinfo{title}{Handling pure camera rotation in keyframe-based
  slam},
\newblock in: \bibinfo{booktitle}{Mixed and Augmented Reality (ISMAR), 2013
  IEEE International Symposium on}, \bibinfo{year}{2013}, pp.
  \bibinfo{pages}{229--238}. \DOIprefix\doi{10.1109/ISMAR.2013.6671783}.
%Type = Article
\bibitem[{Dong et~al.(2014)Dong, Zhang, Jia, and Bao}]{dong_2014_IJCV}
\bibinfo{author}{Z.~Dong}, \bibinfo{author}{G.~Zhang},
  \bibinfo{author}{J.~Jia}, \bibinfo{author}{H.~Bao},
\newblock \bibinfo{title}{Efficient keyframe-based real-time camera tracking},
\newblock \bibinfo{journal}{Computer Vision and Image Understanding}
  \bibinfo{volume}{118} (\bibinfo{year}{2014}) \bibinfo{pages}{97 -- 110}.
%Type = Inproceedings
\bibitem[{Forster et~al.(2014)Forster, Pizzoli, and
  Scaramuzza}]{forster_2014_ICRA}
\bibinfo{author}{C.~Forster}, \bibinfo{author}{M.~Pizzoli},
  \bibinfo{author}{D.~Scaramuzza},
\newblock \bibinfo{title}{Svo : Fast semi-direct monocular visual odometry},
\newblock in: \bibinfo{booktitle}{Robotics and Automation (ICRA), IEEE
  International Conference on}, \bibinfo{year}{2014}.
%Type = Inproceedings
\bibitem[{Herrera et~al.(2014)Herrera, Kannala, Pulli, and
  Heikkila}]{herrera_2014_IC3DV}
\bibinfo{author}{D.~Herrera}, \bibinfo{author}{J.~Kannala},
  \bibinfo{author}{K.~Pulli}, \bibinfo{author}{J.~Heikkila},
\newblock \bibinfo{title}{{DT-SLAM: Deferred Triangulation for Robust SLAM}},
\newblock in: \bibinfo{booktitle}{3D Vision, 2nd International Conference on},
  volume~\bibinfo{volume}{1}, \bibinfo{publisher}{IEEE}, \bibinfo{year}{2014},
  pp. \bibinfo{pages}{609--616}.
%Type = Inproceedings
\bibitem[{Bourmaud and Megret(2015)}]{bourmaud_2015_CVPR}
\bibinfo{author}{G.~Bourmaud}, \bibinfo{author}{R.~Megret},
\newblock \bibinfo{title}{Robust large scale monocular visual slam},
\newblock in: \bibinfo{booktitle}{Computer Vision and Pattern Recognition
  (CVPR), 2015 IEEE Conference on}, \bibinfo{year}{2015}, pp.
  \bibinfo{pages}{1638--1647}. \DOIprefix\doi{10.1109/CVPR.2015.7298772}.
%Type = Inproceedings
\bibitem[{Concha and Civera(2015)}]{concha_2015_IROS}
\bibinfo{author}{A.~Concha}, \bibinfo{author}{J.~Civera},
\newblock \bibinfo{title}{Dpptam: Dense piecewise planar tracking and mapping
  from a monocular sequence},
\newblock in: \bibinfo{booktitle}{Intelligent Robots and Systems (IROS), 2015
  IEEE/RSJ International Conference on}, \bibinfo{year}{2015}, pp.
  \bibinfo{pages}{5686--5693}. \DOIprefix\doi{10.1109/IROS.2015.7354184}.
%Type = Inproceedings
\bibitem[{Greene et~al.(2016)Greene, Ok, Lommel, and Roy}]{greene_2016_ICRA}
\bibinfo{author}{W.~N. Greene}, \bibinfo{author}{K.~Ok},
  \bibinfo{author}{P.~Lommel}, \bibinfo{author}{N.~Roy},
\newblock \bibinfo{title}{Multi-level mapping: Real-time dense monocular slam},
\newblock in: \bibinfo{booktitle}{2016 IEEE International Conference on
  Robotics and Automation (ICRA)}, \bibinfo{year}{2016}, pp.
  \bibinfo{pages}{833--840}. \DOIprefix\doi{10.1109/ICRA.2016.7487213}.
%Type = Article
\bibitem[{Liu et~al.(2016)Liu, Zhang, and Bao}]{liu_2016_ISMAR}
\bibinfo{author}{H.~Liu}, \bibinfo{author}{G.~Zhang}, \bibinfo{author}{H.~Bao},
\newblock \bibinfo{title}{Robust keyframe-based monocular slam for augmented
  reality},
\newblock \bibinfo{journal}{International Symposium on Mixed and Augmented
  Reality (ISMAR)}  (\bibinfo{year}{2016}).
%Type = Article
\bibitem[{Engel et~al.(2016)Engel, Koltun, and Cremers}]{engel_2016_ARXIV}
\bibinfo{author}{J.~Engel}, \bibinfo{author}{V.~Koltun},
  \bibinfo{author}{D.~Cremers},
\newblock \bibinfo{title}{Direct sparse odometry},
\newblock \bibinfo{journal}{CoRR} \bibinfo{volume}{abs/1607.02565}
  (\bibinfo{year}{2016}).
%Type = Inproceedings
\bibitem[{Rosten and Drummond(2006)}]{rosten_2006_ECCV}
\bibinfo{author}{E.~Rosten}, \bibinfo{author}{T.~Drummond},
\newblock \bibinfo{title}{{Machine Learning for High-speed Corner Detection}},
\newblock in: \bibinfo{booktitle}{9th European Conference on Computer Vision -
  Volume Part I, Proceedings of the}, ECCV'06,
  \bibinfo{publisher}{Springer-Verlag}, \bibinfo{address}{Berlin, Heidelberg},
  \bibinfo{year}{2006}, pp. \bibinfo{pages}{430--443}.
%Type = Article
\bibitem[{Nist{\'{e}}r(2004)}]{nister_2004_PAMI}
\bibinfo{author}{D.~Nist{\'{e}}r},
\newblock \bibinfo{title}{{An efficient solution to the five-point relative
  pose problem.}},
\newblock \bibinfo{journal}{Pattern Analysis and Machine Intelligence (PAMI),
  IEEE Transactions on} \bibinfo{volume}{26} (\bibinfo{year}{2004})
  \bibinfo{pages}{756--77}.
%Type = Article
\bibitem[{Faugeras and Lustman(1988)}]{faugeras_1988_IJPRAI}
\bibinfo{author}{O.~Faugeras}, \bibinfo{author}{F.~Lustman},
\newblock \bibinfo{title}{{Motion and structure from motion in a piecewise
  planar environment}},
\newblock \bibinfo{journal}{International Journal of Pattern Recognition and
  Artificial Intelligence} \bibinfo{volume}{02} (\bibinfo{year}{1988})
  \bibinfo{pages}{485--508}.
%Type = Techreport
\bibitem[{Tomasi and Kanade(1991)}]{tomasi_1991_IJCV}
\bibinfo{author}{C.~Tomasi}, \bibinfo{author}{T.~Kanade},
  \bibinfo{title}{{Detection and Tracking of Point Features}},
  \bibinfo{type}{Technical Report}, International Journal of Computer Vision,
  \bibinfo{year}{1991}.
%Type = Book
\bibitem[{Hall(2015)}]{hall_2015_Springer}
\bibinfo{author}{B.~C. Hall}, \bibinfo{title}{{Lie Groups, Lie Algebras, and
  Representations}}, volume \bibinfo{volume}{222}, \bibinfo{edition}{number
  102} ed., \bibinfo{publisher}{Springer- Verlag}, \bibinfo{year}{2015}.
%Type = Article
\bibitem[{Benhimane and Malis(2007)}]{Benhimane_2007_IJRR}
\bibinfo{author}{S.~Benhimane}, \bibinfo{author}{E.~Malis},
\newblock \bibinfo{title}{{Homography-based 2D Visual Tracking and Servoing}},
\newblock \bibinfo{journal}{International Journal of Robotics Research}
  \bibinfo{volume}{26} (\bibinfo{year}{2007}) \bibinfo{pages}{661--676}.
%Type = Book
\bibitem[{Moranna et~al.(2006)Moranna, Martin, and Yohai}]{maronna_2006_Wiley}
\bibinfo{author}{R.~a. Moranna}, \bibinfo{author}{R.~D. Martin},
  \bibinfo{author}{V.~J. Yohai}, \bibinfo{title}{{Robust Statistics}},
  \bibinfo{publisher}{Wiley}, \bibinfo{year}{2006}.
%Type = Inbook
\bibitem[{Kneip et~al.(2012)Kneip, Siegwart, and Pollefeys}]{kneip_2012_eccv}
\bibinfo{author}{L.~Kneip}, \bibinfo{author}{R.~Siegwart},
  \bibinfo{author}{M.~Pollefeys}, \bibinfo{title}{{Finding the Exact Rotation
  between Two Images Independently of the Translation}},
  \bibinfo{publisher}{Springer Berlin Heidelberg}, \bibinfo{address}{Berlin,
  Heidelberg}, \bibinfo{year}{2012}, pp. \bibinfo{pages}{696--709}. \URLprefix
  \url{http://dx.doi.org/10.1007/978-3-642-33783-3{\_}50}.
  \DOIprefix\doi{10.1007/978-3-642-33783-3{\_}50}.
%Type = Inproceedings
\bibitem[{Engel et~al.(2013)Engel, Sturm, and Cremers}]{engel_2013_ICCV}
\bibinfo{author}{J.~Engel}, \bibinfo{author}{J.~Sturm},
  \bibinfo{author}{D.~Cremers},
\newblock \bibinfo{title}{{Semi-dense Visual Odometry for a Monocular Camera}},
\newblock in: \bibinfo{booktitle}{Computer Vision (ICCV), IEEE International
  Conference on}, \bibinfo{publisher}{IEEE}, \bibinfo{year}{2013}, pp.
  \bibinfo{pages}{1449--1456}.
%Type = Article
\bibitem[{Vogiatzis and Hern{\'a}ndez(2011)}]{vogiatzis_2011_IJCV}
\bibinfo{author}{G.~Vogiatzis}, \bibinfo{author}{C.~Hern{\'a}ndez},
\newblock \bibinfo{title}{Video-based, real-time multi-view stereo},
\newblock \bibinfo{journal}{Image and Vision Computing} \bibinfo{volume}{29}
  (\bibinfo{year}{2011}) \bibinfo{pages}{434 -- 441}.
%Type = Inproceedings
\bibitem[{Pizzoli et~al.(2014)Pizzoli, Forster, and
  Scaramuzza}]{pizzoli_2014_ICRA}
\bibinfo{author}{M.~Pizzoli}, \bibinfo{author}{C.~Forster},
  \bibinfo{author}{D.~Scaramuzza},
\newblock \bibinfo{title}{{REMODE}: Probabilistic, monocular dense
  reconstruction in real time},
\newblock in: \bibinfo{booktitle}{IEEE International Conference on Robotics and
  Automation (ICRA)}, \bibinfo{year}{2014}.
%Type = Article
\bibitem[{Lepetit et~al.(2009)Lepetit, Moreno-Noguer, and
  Fua}]{lepetit_2009_IJCV}
\bibinfo{author}{V.~Lepetit}, \bibinfo{author}{F.~Moreno-Noguer},
  \bibinfo{author}{P.~Fua},
\newblock \bibinfo{title}{{EPnP: An Accurate O(n) Solution to the PnP
  Problem}},
\newblock \bibinfo{journal}{International Journal of Computer Vision}
  \bibinfo{volume}{81} (\bibinfo{year}{2009}) \bibinfo{pages}{155--166}.
%Type = Inproceedings
\bibitem[{Glover et~al.(2012)Glover, Maddern, Warren, Reid, Milford, and
  Wyeth}]{glover_2012_ICRA}
\bibinfo{author}{A.~Glover}, \bibinfo{author}{W.~Maddern},
  \bibinfo{author}{M.~Warren}, \bibinfo{author}{S.~Reid},
  \bibinfo{author}{M.~Milford}, \bibinfo{author}{G.~Wyeth},
\newblock \bibinfo{title}{{OpenFABMAP: An open source toolbox for
  appearance-based loop closure detection}},
\newblock in: \bibinfo{booktitle}{2012 IEEE International Conference on
  Robotics and Automation (ICRA)}, \bibinfo{publisher}{IEEE},
  \bibinfo{year}{2012}, pp. \bibinfo{pages}{4730--4735}.
%Type = Inproceedings
\bibitem[{Dong et~al.(2009)Dong, Zhang, Jia, and Bao}]{dong_2009_IJCV}
\bibinfo{author}{Z.~Dong}, \bibinfo{author}{G.~Zhang},
  \bibinfo{author}{J.~Jia}, \bibinfo{author}{H.~Bao},
\newblock \bibinfo{title}{Keyframe-based real-time camera tracking},
\newblock in: \bibinfo{booktitle}{2009 IEEE 12th International Conference on
  Computer Vision}, \bibinfo{year}{2009}, pp. \bibinfo{pages}{1538--1545}.
  \DOIprefix\doi{10.1109/ICCV.2009.5459273}.
%Type = Inproceedings
\bibitem[{Pirker(2010)}]{pirker_2010_BMVC}
\bibinfo{author}{K.~Pirker},
\newblock \bibinfo{title}{Histogram of oriented cameras - a new descriptor for
  visual slam in dynamic environments},
\newblock in: \bibinfo{booktitle}{Proceedings of the British Machine Vision
  Conference}, \bibinfo{publisher}{BMVA Press}, \bibinfo{year}{2010}, pp.
  \bibinfo{pages}{76.1--76.12}. \bibinfo{note}{Doi:10.5244/C.24.76}.
%Type = Article
\bibitem[{Bentley(1975)}]{bentley_1975_ACM}
\bibinfo{author}{J.~L. Bentley},
\newblock \bibinfo{title}{{Multidimensional Binary Search Trees Used for
  Associative Searching}},
\newblock \bibinfo{journal}{Communications ACM} \bibinfo{volume}{18}
  (\bibinfo{year}{1975}) \bibinfo{pages}{509--517}.
%Type = Article
\bibitem[{Wagner et~al.(2010)Wagner, Reitmayr, Mulloni, Drummond, and
  Schmalstieg}]{wagner_2010_TOVCC}
\bibinfo{author}{D.~Wagner}, \bibinfo{author}{G.~Reitmayr},
  \bibinfo{author}{A.~Mulloni}, \bibinfo{author}{T.~Drummond},
  \bibinfo{author}{D.~Schmalstieg},
\newblock \bibinfo{title}{Real-time detection and tracking for augmented
  reality on mobile phones},
\newblock \bibinfo{journal}{Visualization and Computer Graphics, IEEE
  Transactions on} \bibinfo{volume}{16} (\bibinfo{year}{2010})
  \bibinfo{pages}{355--368}.
%Type = Inproceedings
\bibitem[{Pradeep et~al.(2013)Pradeep, Rhemann, Izadi, Zach, Bleyer, and
  Bathiche}]{pradeep_2013_ISMAR}
\bibinfo{author}{V.~Pradeep}, \bibinfo{author}{C.~Rhemann},
  \bibinfo{author}{S.~Izadi}, \bibinfo{author}{C.~Zach},
  \bibinfo{author}{M.~Bleyer}, \bibinfo{author}{S.~Bathiche},
\newblock \bibinfo{title}{Monofusion: Real-time 3d reconstruction of small
  scenes with a single web camera},
\newblock in: \bibinfo{booktitle}{Mixed and Augmented Reality (ISMAR), 2013
  IEEE International Symposium on}, \bibinfo{organization}{IEEE},
  \bibinfo{year}{2013}, pp. \bibinfo{pages}{83--88}.
%Type = Inproceedings
\bibitem[{Curless and Levoy(1996)}]{curless_1996_CGIT}
\bibinfo{author}{B.~Curless}, \bibinfo{author}{M.~Levoy},
\newblock \bibinfo{title}{A volumetric method for building complex models from
  range images},
\newblock in: \bibinfo{booktitle}{Proceedings of the 23rd Annual Conference on
  Computer Graphics and Interactive Techniques}, SIGGRAPH '96,
  \bibinfo{publisher}{ACM}, \bibinfo{address}{New York, NY, USA},
  \bibinfo{year}{1996}, pp. \bibinfo{pages}{303--312}. \URLprefix
  \url{http://doi.acm.org/10.1145/237170.237269}.
  \DOIprefix\doi{10.1145/237170.237269}.
%Type = Article
\bibitem[{Engel et~al.(2016)Engel, Usenko, and Cremers}]{engel2_2016_arxiv}
\bibinfo{author}{J.~Engel}, \bibinfo{author}{V.~C. Usenko},
  \bibinfo{author}{D.~Cremers},
\newblock \bibinfo{title}{A photometrically calibrated benchmark for monocular
  visual odometry},
\newblock \bibinfo{journal}{CoRR} \bibinfo{volume}{abs/1607.02555}
  (\bibinfo{year}{2016}).
%Type = Inproceedings
\bibitem[{Engel et~al.(2015)Engel, Stuckler, and Cremers}]{engel_2015_IROS}
\bibinfo{author}{J.~Engel}, \bibinfo{author}{J.~Stuckler},
  \bibinfo{author}{D.~Cremers},
\newblock \bibinfo{title}{Large-scale direct slam with stereo cameras},
\newblock in: \bibinfo{booktitle}{Intelligent Robots and Systems (IROS), 2015
  IEEE/RSJ International Conference on}, \bibinfo{year}{2015}, pp.
  \bibinfo{pages}{1935--1942}. \DOIprefix\doi{10.1109/IROS.2015.7353631}.
%Type = Article
\bibitem[{Bista et~al.(2016)Bista, Giordano, and Chaumette}]{bista_2016_RAL}
\bibinfo{author}{S.~R. Bista}, \bibinfo{author}{P.~R. Giordano},
  \bibinfo{author}{F.~Chaumette},
\newblock \bibinfo{title}{Appearance-based indoor navigation by ibvs using line
  segments},
\newblock \bibinfo{journal}{IEEE Robotics and Automation Letters}
  \bibinfo{volume}{1} (\bibinfo{year}{2016}) \bibinfo{pages}{423--430}.
%Type = Inproceedings
\bibitem[{Zhang and Koch(2011)}]{zhang_2011_IMVIP}
\bibinfo{author}{L.~Zhang}, \bibinfo{author}{R.~Koch},
\newblock \bibinfo{title}{Hand-held monocular slam based on line segments},
\newblock in: \bibinfo{booktitle}{Proceedings of the 2011 Irish Machine Vision
  and Image Processing Conference}, IMVIP '11, \bibinfo{publisher}{IEEE
  Computer Society}, \bibinfo{address}{Washington, DC, USA},
  \bibinfo{year}{2011}, pp. \bibinfo{pages}{7--14}. \URLprefix
  \url{http://dx.doi.org/10.1109/IMVIP.2011.11}.
  \DOIprefix\doi{10.1109/IMVIP.2011.11}.
%Type = Article
\bibitem[{Dub{\'{e}} et~al.(2016)Dub{\'{e}}, Dugas, Stumm, Nieto, Siegwart, and
  Cadena}]{dube_2016_ARXIV}
\bibinfo{author}{R.~Dub{\'{e}}}, \bibinfo{author}{D.~Dugas},
  \bibinfo{author}{E.~Stumm}, \bibinfo{author}{J.~Nieto},
  \bibinfo{author}{R.~Siegwart}, \bibinfo{author}{C.~Cadena},
\newblock \bibinfo{title}{Segmatch: Segment based loop-closure for 3d point
  clouds},
\newblock \bibinfo{journal}{CoRR} \bibinfo{volume}{abs/1609.07720}
  (\bibinfo{year}{2016}).
%Type = Inproceedings
\bibitem[{Klein and Murray(2008)}]{klein_2008_ECCV}
\bibinfo{author}{G.~Klein}, \bibinfo{author}{D.~Murray},
\newblock \bibinfo{title}{{Improving the Agility of Keyframe-Based
  {\{}SLAM{\}}}},
\newblock in: \bibinfo{booktitle}{Proc. 10th European Conference on Computer
  Vision (ECCV)}, \bibinfo{address}{Marseille}, \bibinfo{year}{2008}, pp.
  \bibinfo{pages}{802--815}.
%Type = Inproceedings
\bibitem[{Micusik and Wildenauer(2015)}]{micusik_2015_CVPR}
\bibinfo{author}{B.~Micusik}, \bibinfo{author}{H.~Wildenauer},
\newblock \bibinfo{title}{Descriptor free visual indoor localization with line
  segments},
\newblock in: \bibinfo{booktitle}{2015 IEEE Conference on Computer Vision and
  Pattern Recognition (CVPR)}, \bibinfo{year}{2015}, pp.
  \bibinfo{pages}{3165--3173}. \DOIprefix\doi{10.1109/CVPR.2015.7298936}.
%Type = Inbook
\bibitem[{Vakhitov et~al.(2016)Vakhitov, Funke, and
  Moreno-Noguer}]{vakhitov_2016_ECCV}
\bibinfo{author}{A.~Vakhitov}, \bibinfo{author}{J.~Funke},
  \bibinfo{author}{F.~Moreno-Noguer}, \bibinfo{title}{Accurate and Linear Time
  Pose Estimation from Points and Lines}, \bibinfo{publisher}{Springer
  International Publishing}, \bibinfo{address}{Cham}, \bibinfo{year}{2016}, pp.
  \bibinfo{pages}{583--599}. \URLprefix
  \url{http://dx.doi.org/10.1007/978-3-319-46478-7_36}.
  \DOIprefix\doi{10.1007/978-3-319-46478-7_36}.
%Type = Inproceedings
\bibitem[{Verhagen et~al.(2014)Verhagen, Timofte, and
  Gool}]{verhagen_2014_WACV}
\bibinfo{author}{B.~Verhagen}, \bibinfo{author}{R.~Timofte},
  \bibinfo{author}{L.~V. Gool},
\newblock \bibinfo{title}{Scale-invariant line descriptors for wide baseline
  matching},
\newblock in: \bibinfo{booktitle}{IEEE Winter Conference on Applications of
  Computer Vision (WACV)}, \bibinfo{publisher}{IEEE}, \bibinfo{year}{2014}, pp.
  \bibinfo{pages}{493--500}.
%Type = Article
\bibitem[{Yammine et~al.(2014)Yammine, Wige, Simmet, Niederkorn, and
  Kaup}]{yammine_2014_CSVT}
\bibinfo{author}{G.~Yammine}, \bibinfo{author}{E.~Wige},
  \bibinfo{author}{F.~Simmet}, \bibinfo{author}{D.~Niederkorn},
  \bibinfo{author}{A.~Kaup},
\newblock \bibinfo{title}{Novel similarity-invariant line descriptor and
  matching algorithm for global motion estimation},
\newblock \bibinfo{journal}{IEEE Transactions on Circuits and Systems for Video
  Technology} \bibinfo{volume}{24} (\bibinfo{year}{2014})
  \bibinfo{pages}{1323--1335}.
%Type = Inproceedings
\bibitem[{Concha and Civera(2014)}]{concha_2014_ICRA}
\bibinfo{author}{A.~Concha}, \bibinfo{author}{J.~Civera},
\newblock \bibinfo{title}{Using superpixels in monocular slam},
\newblock in: \bibinfo{booktitle}{2014 IEEE International Conference on
  Robotics and Automation (ICRA)}, \bibinfo{year}{2014}, pp.
  \bibinfo{pages}{365--372}. \DOIprefix\doi{10.1109/ICRA.2014.6906883}.
%Type = Inproceedings
\bibitem[{Martinez-Carranza and Calway(2010)}]{martinez_2010_BMVA}
\bibinfo{author}{J.~Martinez-Carranza}, \bibinfo{author}{A.~Calway},
\newblock \bibinfo{title}{Unifying planar and point mapping in monocular slam},
\newblock in: \bibinfo{booktitle}{Proceedings of the British Machine Vision
  Conference}, \bibinfo{publisher}{BMVA Press}, \bibinfo{year}{2010}, pp.
  \bibinfo{pages}{43.1--43.11}. \bibinfo{note}{Doi:10.5244/C.24.43}.
%Type = Article
\bibitem[{G{\'{a}}lvez{-}L{\'{o}}pez et~al.(2015)G{\'{a}}lvez{-}L{\'{o}}pez,
  Salas, Tard{\'{o}}s, and Montiel}]{galvez_2015_ARXIV}
\bibinfo{author}{D.~G{\'{a}}lvez{-}L{\'{o}}pez}, \bibinfo{author}{M.~Salas},
  \bibinfo{author}{J.~D. Tard{\'{o}}s}, \bibinfo{author}{J.~M.~M. Montiel},
\newblock \bibinfo{title}{Real-time monocular object {SLAM}},
\newblock \bibinfo{journal}{CoRR} \bibinfo{volume}{abs/1504.02398}
  (\bibinfo{year}{2015}).
%Type = Inproceedings
\bibitem[{Simo-Serra et~al.(2015)Simo-Serra, Trulls, Ferraz, Kokkinos, Fua, and
  Moreno-Noguer}]{serra_2015_ICCV}
\bibinfo{author}{E.~Simo-Serra}, \bibinfo{author}{E.~Trulls},
  \bibinfo{author}{L.~Ferraz}, \bibinfo{author}{I.~Kokkinos},
  \bibinfo{author}{P.~Fua}, \bibinfo{author}{F.~Moreno-Noguer},
\newblock \bibinfo{title}{{Discriminative Learning of Deep Convolutional
  Feature Point Descriptors}},
\newblock in: \bibinfo{booktitle}{Proceedings of the International Conference
  on Computer Vision (ICCV)}, \bibinfo{year}{2015}.
%Type = Inproceedings
\bibitem[{Verdie et~al.(2015)Verdie, Yi, Fua, and Lepetit}]{verdie_2015_CVPR}
\bibinfo{author}{Y.~Verdie}, \bibinfo{author}{K.~M. Yi},
  \bibinfo{author}{P.~Fua}, \bibinfo{author}{V.~Lepetit},
\newblock \bibinfo{title}{Tilde: A temporally invariant learned detector},
\newblock in: \bibinfo{booktitle}{2015 IEEE Conference on Computer Vision and
  Pattern Recognition (CVPR)}, \bibinfo{year}{2015}, pp.
  \bibinfo{pages}{5279--5288}. \DOIprefix\doi{10.1109/CVPR.2015.7299165}.
%Type = Inproceedings
\bibitem[{Han et~al.(2015)Han, Leung, Jia, Sukthankar, and
  Berg}]{han_2015_CVPR}
\bibinfo{author}{X.~Han}, \bibinfo{author}{T.~Leung}, \bibinfo{author}{Y.~Jia},
  \bibinfo{author}{R.~Sukthankar}, \bibinfo{author}{A.~C. Berg},
\newblock \bibinfo{title}{Matchnet: Unifying feature and metric learning for
  patch-based matching},
\newblock in: \bibinfo{booktitle}{2015 IEEE Conference on Computer Vision and
  Pattern Recognition (CVPR)}, \bibinfo{year}{2015}, pp.
  \bibinfo{pages}{3279--3286}. \DOIprefix\doi{10.1109/CVPR.2015.7298948}.
%Type = Inproceedings
\bibitem[{Zagoruyko and Komodakis(2015)}]{zagoruyko_2015_CVPR}
\bibinfo{author}{S.~Zagoruyko}, \bibinfo{author}{N.~Komodakis},
\newblock \bibinfo{title}{Learning to compare image patches via convolutional
  neural networks},
\newblock in: \bibinfo{booktitle}{2015 IEEE Conference on Computer Vision and
  Pattern Recognition (CVPR)}, \bibinfo{year}{2015}, pp.
  \bibinfo{pages}{4353--4361}. \DOIprefix\doi{10.1109/CVPR.2015.7299064}.
%Type = Inproceedings
\bibitem[{Yi et~al.(2016)Yi, Verdie, Fua, and Lepetit}]{yi_2016_CVPR}
\bibinfo{author}{K.~M. Yi}, \bibinfo{author}{Y.~Verdie},
  \bibinfo{author}{P.~Fua}, \bibinfo{author}{V.~Lepetit},
\newblock \bibinfo{title}{{L}earning to {A}ssign {O}rientations to {F}eature
  {P}oints},
\newblock in: \bibinfo{booktitle}{Proceedings of the {C}omputer {V}ision and
  {P}attern {R}ecognition}, \bibinfo{year}{2016}.
%Type = Article
\bibitem[{Pillai and Leonard(2015)}]{sudeep_2015_ARXIV}
\bibinfo{author}{S.~Pillai}, \bibinfo{author}{J.~J. Leonard},
\newblock \bibinfo{title}{Monocular {SLAM} supported object recognition},
\newblock \bibinfo{journal}{CoRR} \bibinfo{volume}{abs/1506.01732}
  (\bibinfo{year}{2015}).
%Type = Incollection
\bibitem[{Kundu et~al.(2014)Kundu, Li, Dellaert, Li, and
  Rehg}]{kundu_2014_ECCV}
\bibinfo{author}{A.~Kundu}, \bibinfo{author}{Y.~Li},
  \bibinfo{author}{F.~Dellaert}, \bibinfo{author}{F.~Li},
  \bibinfo{author}{J.~Rehg},
\newblock \bibinfo{title}{Joint semantic segmentation and 3d reconstruction
  from monocular video},
\newblock in: \bibinfo{editor}{D.~Fleet}, \bibinfo{editor}{T.~Pajdla},
  \bibinfo{editor}{B.~Schiele}, \bibinfo{editor}{T.~Tuytelaars} (Eds.),
  \bibinfo{booktitle}{Computer Vision -- ECCV 2014}, volume
  \bibinfo{volume}{8694} of \textit{\bibinfo{series}{Lecture Notes in Computer
  Science}}, \bibinfo{publisher}{Springer International Publishing},
  \bibinfo{year}{2014}, pp. \bibinfo{pages}{703--718}. \URLprefix
  \url{http://dx.doi.org/10.1007/978-3-319-10599-4_45}.
  \DOIprefix\doi{10.1007/978-3-319-10599-4_45}.
%Type = Inproceedings
\bibitem[{Fioraio and Stefano(2013)}]{fioraio_2013_CVPR}
\bibinfo{author}{N.~Fioraio}, \bibinfo{author}{L.~D. Stefano},
\newblock \bibinfo{title}{Joint detection, tracking and mapping by semantic
  bundle adjustment},
\newblock in: \bibinfo{booktitle}{Computer Vision and Pattern Recognition
  (CVPR), 2013 IEEE Conference on}, \bibinfo{year}{2013}, pp.
  \bibinfo{pages}{1538--1545}. \DOIprefix\doi{10.1109/CVPR.2013.202}.
%Type = Article
\bibitem[{Savarese et~al.(2012)Savarese, Chao, Bagra, and
  Bao}]{savarese_2012_CVPR}
\bibinfo{author}{S.~Savarese}, \bibinfo{author}{Y.-W. Chao},
  \bibinfo{author}{M.~Bagra}, \bibinfo{author}{S.~Y. Bao},
\newblock \bibinfo{title}{Semantic structure from motion with points, regions,
  and objects},
\newblock \bibinfo{journal}{2012 IEEE Conference on Computer Vision and Pattern
  Recognition (CVPR)} \bibinfo{volume}{00} (\bibinfo{year}{2012})
  \bibinfo{pages}{2703--2710}.
%Type = Inproceedings
\bibitem[{Yang et~al.(2016)Yang, Song, Kaess, and Scherer}]{yang_2016_IROS}
\bibinfo{author}{S.~Yang}, \bibinfo{author}{Y.~Song},
  \bibinfo{author}{M.~Kaess}, \bibinfo{author}{S.~Scherer},
\newblock \bibinfo{title}{Pop-up slam: Semantic monocular plane slam for
  low-texture environments},
\newblock in: \bibinfo{booktitle}{IEEE/RSJ International Conference on
  Intelligent Robots and Systems (IROS)}, \bibinfo{publisher}{IEEE},
  \bibinfo{year}{2016}.

\end{thebibliography}
}	
\end{singlespace}

\end{document}